%% file: main.tex
\documentclass{article}

\usepackage[svgnames]{xcolor}         
\usepackage{hyperref}       
\hypersetup{colorlinks,linkcolor={blue},citecolor={blue},urlcolor={RoyalBlue}}
\PassOptionsToPackage{numbers, compress}{natbib}

\usepackage{graphicx} 
\usepackage{caption}  
\usepackage{subcaption} 

\usepackage[skins]{tcolorbox}
\usepackage{tikz}
\usepackage[preprint]{neurips_2024}

\usepackage[utf8]{inputenc} 
\usepackage[T1]{fontenc}    
\usepackage{url}            
\usepackage{booktabs}       
\usepackage{amsfonts}       
\usepackage{nicefrac}       
\usepackage{microtype}      
\usepackage{amsmath}

\usepackage[capitalize]{cleveref}
\crefformat{section}{\S#2#1#3} 
\crefformat{subsection}{\S#2#1#3}
\crefformat{subsubsection}{\S#2#1#3}

\usepackage{listings}
\usepackage{enumitem}
\usepackage{tabularx}
\usepackage{float}
\usepackage{ltxtable}
\usepackage{cprotect}
\usepackage{fancyvrb,fvextra}
\usepackage{placeins}
\usepackage[framemethod=tikz]{mdframed}
\definecolor{light-grey}{gray}{0.95}
\newmdenv[
  innerlinewidth=0.5pt,
  roundcorner=5pt,
  linecolor=light-grey,
  backgroundcolor=light-grey,
  innerleftmargin=6pt,
  innerrightmargin=6pt,
  innertopmargin=6pt,
  innerbottommargin=6pt
]{roundedbox}
\usepackage{titletoc}

\newcommand{\modelname}{Gemini 1.5 Pro}

\newcommand{\autoaisGFC}{AutoAIS$_{G15-FC}$}

\title{Long-Span Question-Answering: Automatic Question Generation and QA-System Ranking via Side-by-Side Evaluation}

\author{%
  Bernd Bohnet \And
  Kevin Swersky \And
  Rosanne Liu \And 
  Pranjal Awasthi \And
  Azade Nova \And
  Javier Snaider \And 
  Hanie Sedghi \And
  Aaron T Parisi \AND 
  Michael Collins \And
  Angeliki Lazaridou \And
  Orhan Firat \And
  Noah Fiedel \AND
  Google DeepMind 
}

\begin{document}

\maketitle

\begin{abstract}
  We explore the use of long-context capabilities in large language models to create synthetic reading comprehension data from entire books. Previous efforts to construct such datasets relied on crowd-sourcing~\cite{narrativeqa}, but the emergence of transformers with a context size of 1 million or more tokens~\cite{gemini1-5} now enables entirely automatic approaches. Our objective is to test the capabilities of LLMs to analyze, understand, and reason over problems that require a detailed comprehension of long spans of text, such as questions involving character arcs, broader themes, or the consequences of early actions later in the story. We propose a holistic pipeline for automatic data generation including question generation, answering, and model scoring using an ``Evaluator''. We find that a relative approach, comparing answers between models in a pairwise fashion and ranking with a Bradley-Terry model, provides a more consistent and differentiating scoring mechanism than an absolute scorer that rates answers individually. We also show that LLMs from different model families produce moderate agreement in their ratings. We ground our approach using the manually curated NarrativeQA dataset, where our evaluator shows excellent agreement with human judgement and even finds errors in the dataset. Using our automatic evaluation approach, we show that using an entire book as context produces superior reading comprehension performance compared to baseline no-context (parametric knowledge only) and retrieval-based approaches.
\end{abstract}

\addtocontents{toc}{\protect\setcounter{tocdepth}{0}}
\section{Introduction}
\label{sec:intro}

\input{section_1_introduction}

\section{Question Generation}
\label{sec:qa_gen}

\input{section_2_question_generation_full_recipe}

\section{Evaluation Methods}
\label{sec:eval}
\input{section_3_automatic_eval}

\section{Evaluation of Automatically Generated QA Datasets}
\label{sec:exps}

\input{section_4_experimental_results}

\section{Analysis of Automatic Raters}
\label{sec:grounding}
\input{section_5_grounding_auto_raters}

\section{Related Work}
\label{sec:related}
\input{section_6_related_work}

\section{Conclusion}
\label{sec:concl}
\input{section_7_conclusion}

\bibliographystyle{unsrtnat}
\bibliography{references}

\addtocontents{toc}{\protect\setcounter{tocdepth}{2}}

\appendix
\include{appendix}

\end{document}

%% file: section_1_introduction.tex
The advent of long-context large language models (LLMs), capable of processing millions of tokens at once~\citep{gemini1-5}, has recently become available, unlocking new potential to rapidly process large amounts of new data, without the need for re-training or fine-tuning. These models hold the potential to revolutionize fields like document analysis, historical research, and scientific discovery by enabling nuanced reasoning over extensive amounts of data.

However, this potential remains largely untapped due to the scarcity of datasets specifically designed to benchmark and train these advanced reasoning capabilities over long context lengths. Existing datasets often focus on shorter context lengths with short-form, factual answers and are ill-suited for evaluating the complex reasoning required to understand and synthesize information from large amounts of data. This lack of suitable benchmarks hinders both the evaluation and improvement of long-context LLMs.

To address this critical gap, we propose a novel framework for automatically constructing and evaluating complex question-answering (QA) benchmarks tailored for long-context LLMs. Our approach specifically focuses on book-based QA, a domain that presents a unique opportunity to test the limits of long-context reasoning. Books, with their rich narratives and complex character relationships, demand a deep understanding of both the explicit text and the implicit context. Manually creating such benchmarks, however, is an arduous task, requiring significant human effort and expertise, ultimately limiting the scale and complexity of the resulting datasets.

\begin{figure}[t!]
    \centering
    \includegraphics[width=\textwidth]{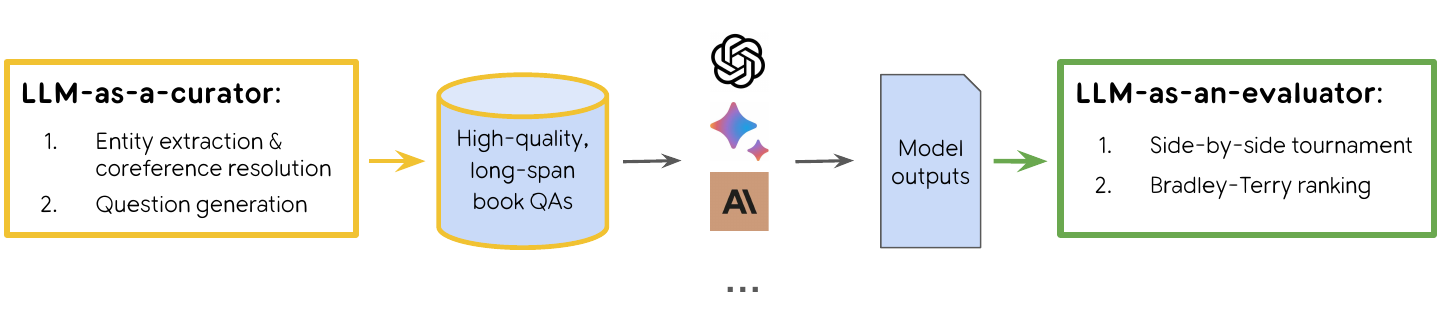}
    \caption{Overview of our framework. We use LLM-as-a-curator to generate a high-quality dataset, and then LLM-as-an-evaluator to rank the performances of a range of models on this dataset. The whole process incurs very little manual labor from humans, and instead leverages the creation and judgement power of LLMs.}
    \label{fig:overview}
    \vspace{-4pt}
\end{figure}

Previous work on long-context QA has developed such benchmarks manually through crowd-sourcing~\citep{narrativeqa}, but this is not easily scalable. We develop a framework using a long-context LLM~\cite{gemini1-5} to automatically create challenging QA pairs from books, and crucially, to automatically evaluate performance; \cref{fig:overview} outlines the framework.
We validate this framework using a suite of commercial frontier models, including \modelname{}~\cite{geminiteam2024gemini, gemini1-5}, GPT-4 Turbo~\cite{openai2024gpt4} and Claude 3 Opus~\cite{claude} to answer these questions in no-context (using only parametric knowledge) and retrieval-based settings.

Evaluating long-form answers involving many portions of a long text via human raters is a time-intensive task that requires expertise in the subject matter and reading comprehension assessment. This prohibits the manual evaluation of models at scale. Instead, we explore automatic methods for comparing model performance. We propose both absolute and relative metrics based on using a model as an attributable-to-identifiable-sources (AIS) system. The absolute approach prompts an LLM to rate whether a proposed answer is correct, given the question and the book as context. We find that it gives a good sense of factual accuracy, but does not produce an informative ranking between models. The relative system prompts a long-context LLM to state which of two proposed answers is better (or that neither is correct), given the source text. We find that this produces a much more informative and discriminative ranking across models.

We focus our evaluation on two distinct tasks.
First, we focus on two specific books from the PG-19 corpus~\cite{pg-19} --- Les Misérables (732k tokens) and The Wild Huntress (230k tokens). Our analysis shows that providing the full book as context provides significantly improved results, both in absolute and relative terms.

Second, we apply LLMs to the NarrativeQA dataset~\citep{narrativeqa}, a manually curated dataset of long-context QA pairs from books and movie scripts. This dataset, created through crowd-sourcing, serves as a grounded validation of our approach. Using Gemini 1.5 Pro on the full text, we find strong agreement between model-based and human answers. Surprisingly, we also find that the Gemini 1.5 Pro-based rater detects a number of incorrect ground-truth answers in the dataset (see Appendix~\ref{sec:narrativeqa_noise} for examples).

%% file: section_2_question_generation_full_recipe.tex
In this section we outline our approach to question generation. We aim to generate questions that challenge the capabilities of current retrieval-augmented and long-context generative AI models.
We prompt the model to generate questions that require reasoning and synthesis over large spans of input text (i.e. content of a book), and to answer in a factually accurate and comprehensive way. The questions should  \textbf{not} ask for localized facts, and instead require the model to incorporate information from the entire input. Examples of generated questions can be found in Appendix~\ref{sec:examples_of_generated_questions}.

The overall idea is to use a prompt-based method alongside an LLM --- in this case, the Gemini 1.5 Pro~\citep{gemini1-5} --- to generate questions with selected entities. These entities are typically proper nouns, such as characters in a fictional book, important locations, or significant events. The entities are extracted via a coreference resolution system, which outputs the complete coreference chains for each entity. This method allows us to identify the importance of an entity through its frequency of mention, as well as to gather all text passages where the entity is not only named but also referred to by a pronoun.

\paragraph{Entity extraction and co-reference resolution.}

To extract entities and their reference chains, we apply coreference resolution to books to identify the most frequent entities and occurrences in the text\footnote{The entity chains allow us to identify relevant passages for an entity through coreferences, even if the entity is not directly named, enabling the use of shorter inputs that fit within the size constraints of the employed language model while still providing relevant content.}. We use the top-ranked coreference resolution system \citep{bohnet-etal-2023-coreference}  according to Papers with Code\footnote{\url{https://paperswithcode.com/sota/coreference-resolution-on-ontonotes}} (May 2024). The system has demonstrated high accuracy across various languages and for languages unseen during training \citep{bohnet-etal-2023-coreference}, which 
is advantageous when dealing with older books due to language changes over time. An example annotation is formatted as follows: {\em \textcolor{blue}{[15 Sire]}, said \textcolor{red}{[6 M. Myriel]}, \textcolor{blue}{[15 you]} are looking at a good man, and \textcolor{red}{[6 I]} at a great man.''}

\paragraph{Prompt-based question generation.}

After extracting the entities and their reference chains, we sort the entities by frequency and go down the list to generate questions involving each entity. Our goal is to create challenging questions that require the model to reason across large spans of text. Consequently, we instruct the model to avoid explicit mention of the entities, thus requiring it to resolve the entities involved. Although datasets like Quoref \citep{dasigi-etal-2019-quoref} emphasize resolving referring expressions, our approach goes further by generating questions that require deeper understanding and reasoning. Our question generating prompt can be found in~Appendix \ref{sec:appendix_question_generating_prompt}. We generated question datasets for two books: {\em Les Misérables by Victor Hugo} and {\em The Wild Huntress by Mayne Reid}.
The texts for these books were sourced from the PG19 dataset \cite{pg-19}. 
{\em Les Misérables} was selected due to its extensive length, while {\em The Wild Huntress} was chosen for its lower popularity and reduced online presence, making it less likely to be represented in language model training data.
This process yielded 1,117 questions for {\em Les Misérables} and 1,000 questions for {\em The Wild Huntress}. 
\paragraph{Limitations}
At the time of writing, \modelname{} is the only publicly available frontier model with long-context capabilities. Claude 3 allows for 1M tokens only for specific use cases an upon inquiry~\footnote{``*1M tokens available for specific use cases, please inquire.'', \url{https://www.anthropic.com/news/claude-3-family}}. We therefore use \modelname{} in our full context experiments (question generator, auto-evaluation model), but note that the methodology presented here can just as easily be applied to any long context model.

Our questions are designed to target specific criteria: difficulty for retrieval-based models, and requirement of long-span reasoning, but there are other criteria that may be of interest that we do not focus on here. We note that the methodology can be easily adapted to suit different needs by changing the question generating prompt in Appendix~\ref{sec:appendix_question_generating_prompt} as needed.

%% file: section_3_automatic_eval.tex
Evaluating answers generated by generative AI models often involves expert human raters, but this requires the raters to be familiar with the text and involves significant time and costs. Ideally, the raters would also have expertise in reading comprehension assessment. This makes such evaluation at scale too costly in practice. Instead, we explore the potential of long-context models to automatically evaluate answers from different systems. In the literature, automatic evaluation has recently become more widely adopted and accepted~\citep{rashkin_ais_2023,gao-etal-2023-rarr,bohnet2023attributed}. With long-context models, we can create an automatic evaluator by providing the entire book as context, followed by a question and candidate answer.

We not only seek technical correctness, but also a high level of quality in the model answers. Answers to a question can be factually correct but may lack sufficient detail or contain unnecessary content. Factual correctness is typically evaluated on a binary  scale, classifying answers as either correct or incorrect based on a given source. Current frontier models can achieve high factual accuracy already in no-context settings, particularly for well-known works like {\em Les Misérables} (see Section~\ref{sec:exps}). This high accuracy makes it challenging to differentiate between models using absolute performance measures, as scores often cluster closely.

We therefore introduce a side-by-side comparison method to assess the quality of answers between different models and models using different context lengths. Side-by-side comparisons are widely used in human studies to assess difficult-to-quantify elements in systems. 
This approach has been adopted in a number of areas for evaluation, such as ranking conversational LLM performance~\citep{lmsys}, preference tuning language models~\cite{directpreferenceoptimization}, and rating text-to-image models~\cite{imagen, stablediffusion3}.

With this method, we get comparisons between two systems. We convert this into a total ordering using the well-established Bradley-Terry model~\cite{BradleyTerry} to compute a ranking. The strength score from this can easily be used to compute a probability that an answer from system A is better than the answer from system B.
In the following, we will review absolute ratings for QA and present the side-by-side evaluation as a complementary approach. 

\subsection{Ratings with AutoAIS}
\label{sec:autoais}

Attributable to Identified Sources (AIS), a human evaluation method proposed by \citet{rashkin_ais_2023} assigns a binary result to a pair $(s, c)$, where $s$ is a sentence and $c = (c_\ell, t)$ is a tuple consisting of a linguistic context $c_\ell$ and an optional time $t$ (some statements are only entailed by the context when conditioned on a specific time). We chose {\em AutoAIS} for its ability to assess the factual grounding of answers within a large context, specifically the entire book used as the source material.

Given some trusted source text $P$, AIS is {\em True} when $s$ in the context of $c$ at a time $t$ is Attributable to Identified Sources $P$ otherwise {\em False}.
This definition is extended by \cite{rashkin_ais_2023} to {\em Attribution of Entire Utterances} or even a multi-sentence utterance $U$. Then, the utterance is evaluated in a ``single shot''. The latter procedure is simpler and less costly to apply. Following this procedure, {\em AutoAIS} has been used on fine-grained sentence-level data by \citet{gao-etal-2023-rarr} using Natural Language Inference (NLI) \citep{honovich-etal-2021-q2} as a rater, which correlates well with AIS scores \citep{gao-etal-2023-rarr}.


In the context of Question-Answering, the ``single shot'' attribution method was applied by \cite{bohnet2023attributed} to evaluate the performance of a number of QA-systems using the questions of the Natural Question corpus \citep{kwiatkowski2019natural}. For the AutoAIS-score, they use the output of the NLI classifier (1 for attributable vs 0 for non-attributable) if $P$ entails a question-answer pair. The total AutoAIS score is simply the average of the individual AutoAIS scores in the dataset.


We adapt this method using an entire book as context $(c_\ell = P)$ and the full answers as a multi-sentence utterance, testing if an answer is attributable to a book. See Appendix~\ref{sec:appendix_autoais_prompt} for the prompt.

While AutoAIS provides an absolute measure of factual accuracy, we primarily employ a relative rating approach using the Bradley-Terry model to compare systems. Side-by-side comparisons provide a more nuanced assessment as they take into account the answer strengths and weaknesses, even when factual absolute scores are similar. 

\subsection{Side-by-Side Evaluation and Ranking with Bradley-Terry Model}
\label{sec:bradley-terry}

We employ \modelname{} with up to 1M token context window as an auto-rater for side-by-side evaluations. For a given question, this auto-rater compares a pair of answers. Its responses are either {\em system-A is  better}, {\em system-B is better} or {\em None}, if both answers are deemed non-factual (see prompt in Appendix~\ref{sec:relative_comparison_prompt}). Non-factual ratings are excluded from further analysis.
%
%
To produce a ranking, we utilize the Bradley-Terry model, which is commonly used in domains like chess and Go to assess player strength. Here, $\gamma$ denotes the playing strength or skill of players.
\begin{equation}
P(i~\mathrm{beats}~j) = \frac{\gamma_i}{\gamma_i + \gamma_j}.
\label{eq:bradley-terry}
\end{equation}
By fitting the Bradley-Terry \cite{BradleyTerry, hunter_2004_mm, Zermelo1929} model to our pairwise comparisons, we obtain learned scores that enable us to rank the models. $W$ captures the results of our side-by-side evaluation process. $W$ is a matrix where each cell ($w_{ij}$) reflects how often model $i$ outperformed model $j$.

 When we generalize to $m$ players, we need to estimate the strength for ${\gamma_1 \ldots \gamma_m}$. The log-likelihood can be written as,
\begin{equation}
\ell(\gamma) = \sum_{i=1}^{m} \sum_{j=1}^{m} \left[ w_{ij} \ln \gamma_i - w_{ij} \ln (\gamma_i + \gamma_j) \right].
\label{eq:bradley-terry-m}
\end{equation}
We can then apply maximum likelihood estimation to learn the $\gamma$ parameters.


A limitation of the Bradley-Terry model is that it necessitates a number of pairwise evaluations to yield statistical significance. 
Given our limited number of systems $n$, we conduct pairwise comparisons between all systems, sampling $c$ questions from our datasets, resulting in a total of $c \cdot n \cdot (n - 1) / 2$ LLM calls. In this study we set $c$ to 200. These questions are randomly sampled, and we ensure the same set of questions is used for all system comparisons. We further randomize the ordering of the presented answers in the pairwise comparisons to mitigate presentation order as a potential source of bias. The confidence intervals throughout the paper are estimated via bootstrapping.

%% file: section_4_experimental_results.tex
As outlined, our goal is to enable generative AIs to create and evaluate datasets. To this end, we explore two question-answer datasets from the book \textit{Les Misérables} and \textit{The Wild Huntress} (see \cref{sec:qa_gen}).  We use \modelname{}, GPT-4 Turbo and Claude 3 to answer the questions of both datasets. These state-of-the-art language models are commonly referred to as \emph{frontier models}. Due to the limited context window of some models, we explore Retrieval-Augmented Generation (RAG) to retrieve useful passages from the books. This method indexes passages using BM25, a TF-IDF-based retrieval algorithm \cite{robertson2009probabilistic}, and stores the results in an index. For each question, we query BM25, retrieving the most relevant passages up to a maximum of 4k tokens. To ensure coherent context for the models, the retrieved passages are arranged chronologically to reflect the book's timeline. In contrast, \modelname{} can accommodate an entire book, eliminating the need for data pre-processing, indexing, and retrieval pipelines.

\subsection{Evaluating Factual Correctness with AutoAIS}

To evaluate the factual accuracy of the generated answers, we use the AutoAIS method described in Section \ref{sec:autoais}. When leveraging the long-context capabilities of Gemini (providing an entire book as context), we refer to the method as \autoaisGFC{}. Using \autoaisGFC{}, we prompt \modelname{} to determine whether the answer, given the book and the question, is factually correct. We show prompts for both question answering and \autoaisGFC{} evaluation in~\cref{sec:appendix_prompts}. 
\cref{tab:absolute-ratings} presents the accuracy and 95\% Confidence Intervals (CI)\footnote{CI calculated using the standard formula for a Bernoulli distribution: $\hat{p} \pm z \cdot \sqrt{\hat{p}(1 - \hat{p})/n}$, where $n$ = number of QA-pairs, $\hat{p}$ = accuracy, and $z=1.96$ (95\% CI).} for each frontier model and context size. %

\begin{table}[h]
\vspace{-2pt}
  \centering
  \caption{\autoaisGFC{} accuracy and CI, using different LLMs and context sizes. }
  \renewcommand{\arraystretch}{1.1}
  \begin{tabular}{llcc} 
    \toprule
    Context & System &  \em{Les Misérables} & \em{The Wild Huntress}\\
            &        & Accuracy \& CI       & Accuracy \& CI\\
    \midrule
    \midrule
    No Context & Gemini 1.5 Pro     & 87.7 $\pm$ 1.8 & 27.3 $\pm$ 2.8  \\ \hline
    4k RAG & Claude 3               & 85.6 $\pm$ 2.1 &  72.2 $\pm$ 2.8\\
    4k RAG & GPT-4 Turbo            & 84.6 $\pm$ 2.1 &  72.1 $\pm$ 2.8 \\ \hline
    Full Context & Gemini 1.5 Pro   & 92.2 $\pm$ 1.6 & 90.0 $\pm$ 1.9 \\
    \bottomrule
  \end{tabular}
  \label{tab:absolute-ratings}
\end{table}

As shown in Table \ref{tab:absolute-ratings}, all LLMs and settings achieve high factual accuracy on the {\em Les Misérables} question set. This is expected, given the book's widespread popularity, extensive online presence, and numerous adaptations across various media. The systems likely possess a high level of pre-trained knowledge about this book. This is in contrast to {\em The Wild Huntress}, which is less well-known, as reflected in lower accuracy scores for smaller context sizes.  
Despite the dataset for {\em Les Misérables} comprising 1,117 questions,  statistically significant accuracy differences (p$<$0.001) consistently emerge only when using the entire book as context. For {\em The Wild Huntress}, only No context and Full Context show statistically significant accuracy differences compared to all other settings. This finding suggests that {\em factual accuracy alone may not be a sufficiently discerning metric} for evaluating LLM performance on widely known texts. 

Our hypothesis is that the answer quality should also be considered: specifically, whether the answer is sufficient and provides the right amount of detail.

\subsection{Side-by-Side Evaluation and Ranking with the Bradley-Terry Model}

We employ the Bradley-Terry Model \citep{BradleyTerry}, as outlined in \cref{sec:bradley-terry}, to rank frontier models based on their relative answer quality strengths using no context, RAG 4k and entire book as context.

\begin{figure}[h!]
    \centering
    \begin{minipage}{0.495\textwidth}
        \centering
        \includegraphics[width=1.0\textwidth]{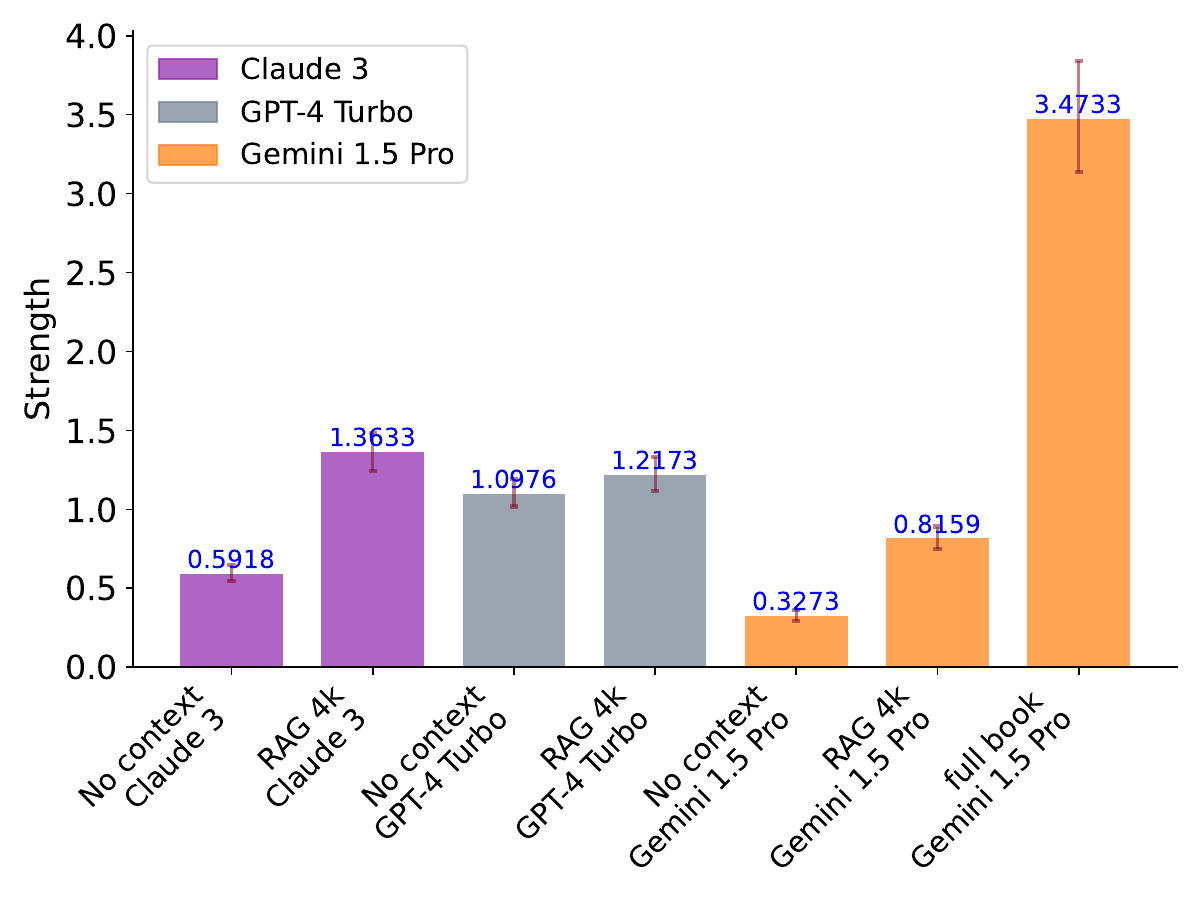}
        \caption{{\em Les Mi\'serables} QA-quality ranking. \label{fig:long-context-les_mis-qa}  }
    \end{minipage}\hfill
    \begin{minipage}{0.495\textwidth}
        \centering
        \includegraphics[width=1.0\textwidth]{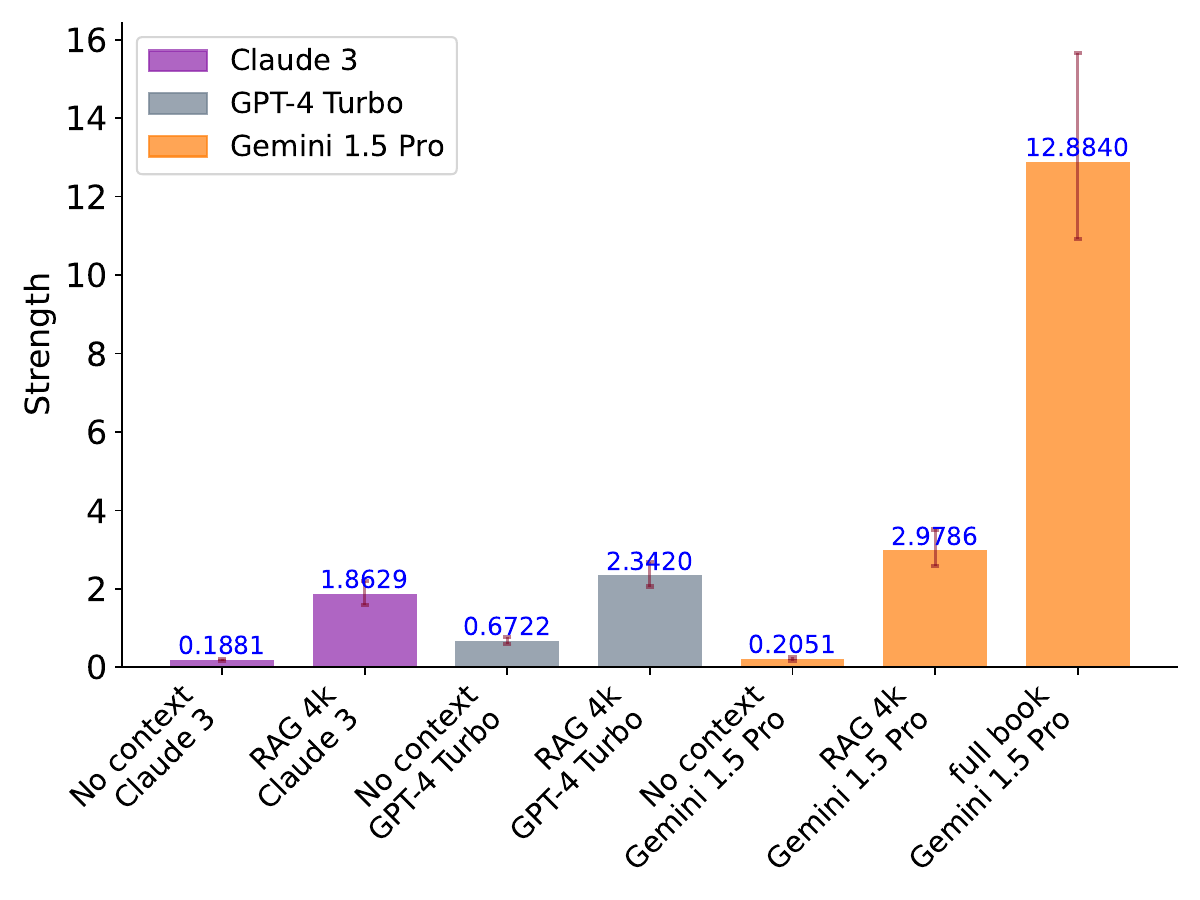} 
        \caption{{\em The Wild Huntress} QA-quality ranking.
        \label{fig:long-context_twh-qa} }
    \end{minipage}
\end{figure}


The log-odds of model $i$ outperforming model $j$ is represented by the difference of respective scores. The model strength has a direct mapping to the probability that an answer from Model $M_A$ is better than an answer from $M_B$: $P(M_A \text{ answers better than } M_B) = \frac{\gamma_A}{\gamma_A + \gamma_B} = \frac{e^{\beta_A}}{e^{\beta_A} + e^{\beta_B}}$.

Figure \ref{fig:long-context-les_mis-qa} and \ref{fig:long-context_twh-qa}  summarize the results of this evaluation. As expected, providing relevant context through retrieval improves the answers of the models. When using the entire book {\em Les Misérables} as context, Gemini 1.5 Pro outperforms all other systems by a large margin. 
For example, given $e^{\beta}$ values shown on top of bars in the Figures, full context \modelname{} provides better answers than retrieval-augmented generation with 4k tokens using \modelname{} with probability $P=\frac{3.4733}{3.4733 + 0.8159} = 0.8097$, or in $81\%$ of cases. Using the full book as context with \modelname{} provides a better answer compared to retrieval-augmented GPT4-Turbo with 4k tokens in $74\%$ of cases. Similar conclusions can be drawn from the relative scores for {\em The Wild Huntress}. Overall, more context consistently improves performance in comparative evaluation, regardless of whether there is significant prior parametric knowledge about the book.

%% file: section_5_grounding_auto_raters.tex
In this section, we analyze the reliability of using LLMs as auto-raters in long-context QA tasks.
We first investigate the variability in using different models as auto-raters. We test whether models may favor outputs from their own family when presented with side-by-side answers.
Next, we validate our approach of using an LLM as an auto-rater by comparing model outputs to gold-standard answers from the NarrativeQA dataset~\cite{narrativeqa}, using the evaluation methods introduced in Section~\ref{sec:eval}.

\subsection{Do Auto-Raters prefer their own answers?}

To test whether automatic raters exhibit a bias (i.e. prefer their own answers), we designed a 2x2 factorial experiment in which both Gemini 1.5 Pro and GPT-4 Turbo generate answers to a shared set of questions. Each model then evaluates the full set of answers, including its own. 
We investigate this under two conditions: (1) without additional context, and (2) with context retrieved from the source text.  The retrieval-based rater uses prompts which include up to 4k sentence piece tokens retrieved with BM25~\cite{robertson2009probabilistic} from the book {\em Les Misérables}. The query passed to the retriever is the concatenated text of the question and the two answers under comparison. The answer sets are from Gemini 1.5 Pro and GPT-4 Turbo in the no-context setting, which are labeled as system A and system B, respectively.

\begin{figure}[th!]
    \begin{subfigure}[b]{0.25\textwidth}
        \centering
        \includegraphics[width=\textwidth]{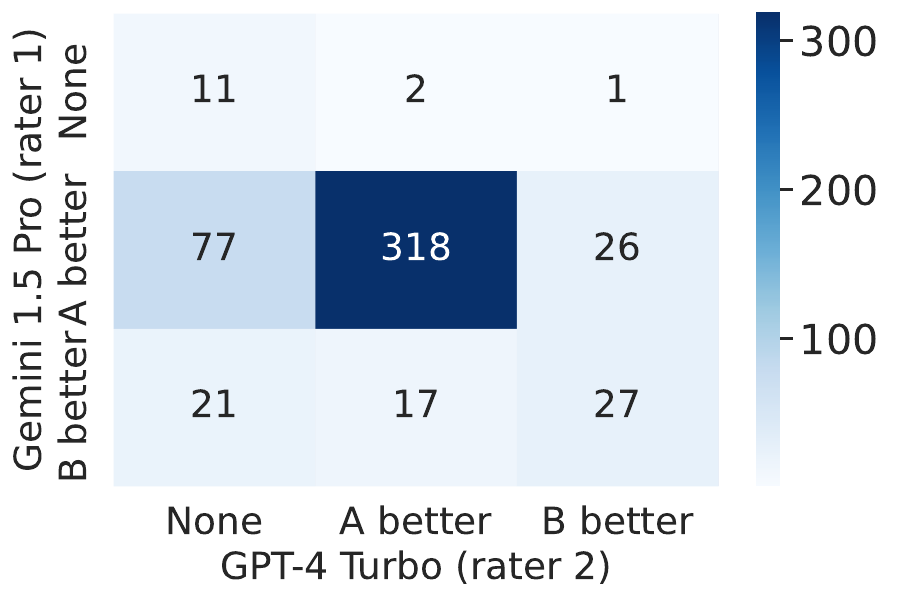}
      
        \caption{Auto-Rater using a prompt without context.}  
        \label{fig:zero-shot-raters-bias}  
    \end{subfigure}
    \hfill
    \begin{subfigure}[b]{0.24\textwidth}
        \centering
        \includegraphics[width=\textwidth]{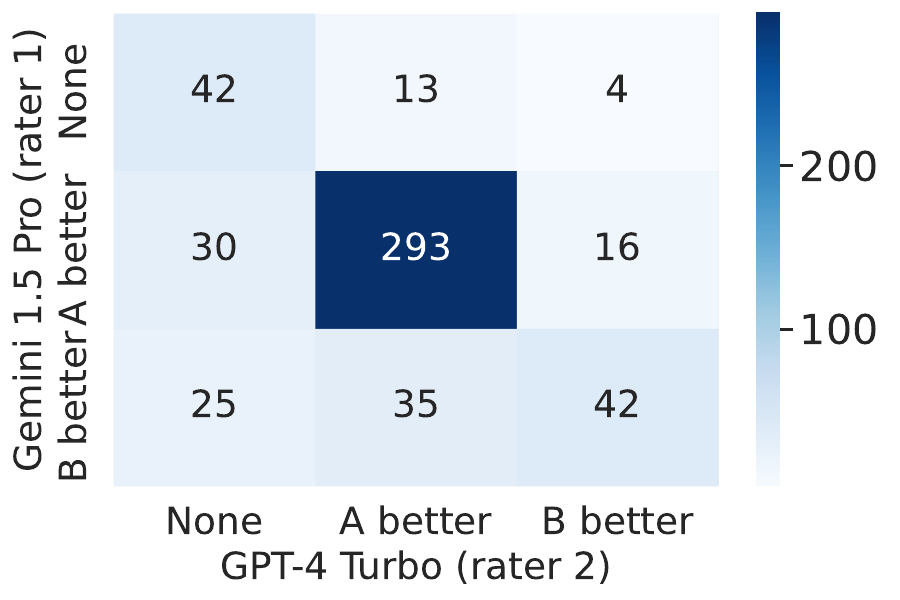}

        \caption{Auto-Rater with 4k retrieved context.}  
        \label{fig:rag4k-bias} 
    \end{subfigure}
    \hfill
    \begin{subfigure}[b]{0.25\textwidth}
        \centering
        \includegraphics[width=\textwidth]{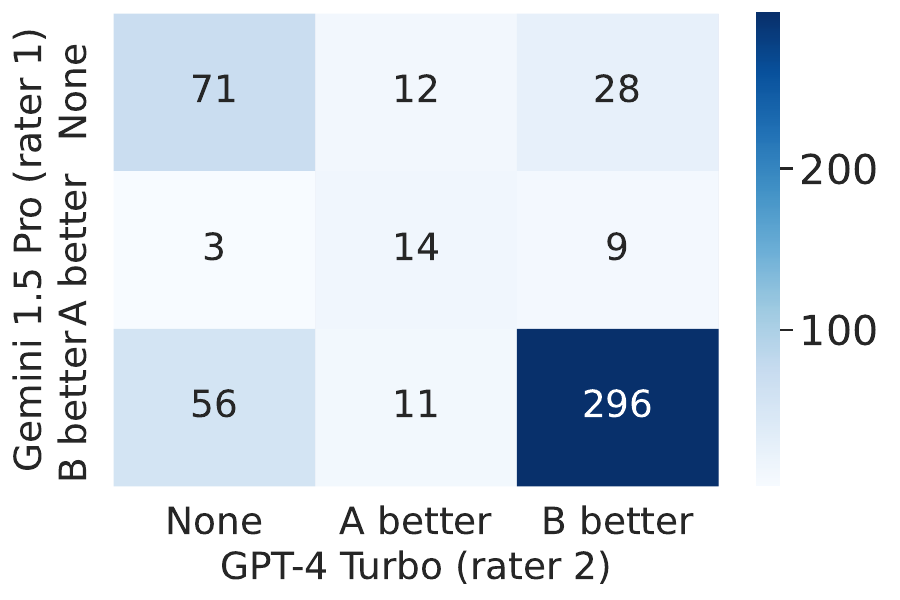}
        \caption{Auto-Rater 4k context and swapped systems.}  
        \label{fig:swap_system_a_b} 
    \end{subfigure}
    \hfill
    \begin{subfigure}[b]{0.24\textwidth}
        \centering
        \includegraphics[width=\textwidth]{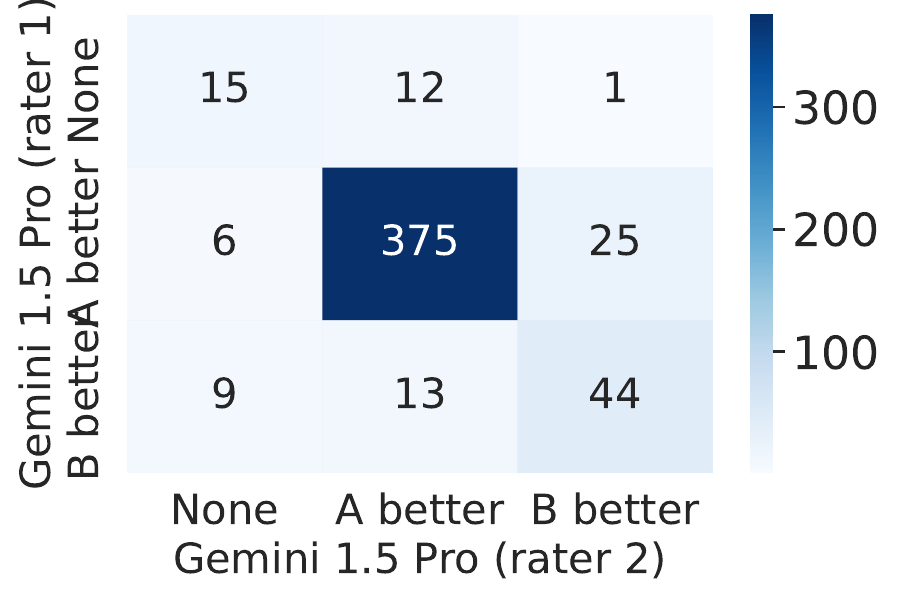}
        \caption{Self-consistency using the entire book as context.}  
        \label{fig:full-context-self-bias} 
    \end{subfigure}
    \caption{Auto-Rater bias analysis. In all matrices System A$=$Gemini 1.5 Pro, and B$=$GPT-4 Turbo.} 
    \label{fig:bias-analysis}
\end{figure}

The most difficult scenario for raters is when they must rely solely on their prior knowledge, as occurs when the model is prompted without additional context in a zero-shot fashion. Figure~\ref{fig:zero-shot-raters-bias}  shows the heat-map for the \textbf{no-context raters}. The matrix trace indicates 356 agreements out of 500 total trials, resulting in a 71.2\% agreement rate.  Regarding inter-rater agreement, Cohen's Kappa is calculated as  $\kappa = 0.302$ which is considered fair agreement. 
In contrast, for the retrieval-augmented \textbf{4k-context raters}, we observe a higher agreement rate of 75.4\% and  Cohen's Kappa  is $\kappa = 0.497$ indicating moderate agreement.
We also evaluated performance swapping the response labels using System A for GPT-4 Turbo, and System B for Gemini 1.5 Pro (Figure~\ref{fig:swap_system_a_b}).  The results showed similar agreement rate (76.2\%), and  Cohen's Kappa ($\kappa = 0.477$). Figure \ref{fig:full-context-self-bias} shows self-consistency when using \modelname{} with the book as context which has 86\% agreement and $\kappa = 0.598$. This analysis indicates moderate agreement between Gemini 1.5 Pro and GPT-4, suggesting that both models could serve as suitable auto-rater, provided they have sufficient context.

\subsection{Grounding LLM-as-an-evaluator performance with NarrativeQA}

Our goal is to determine if an auto-rater as used in \cref{sec:exps}, aware only of the context (excluding the ground-truth answer), produces a similar ranking compared to a rater who has access to correct answers which call in the {\em ground-truth rater}. To this end, we use NarrativeQA~\cite{narrativeqa} which consists of $46,765$ question-answer pairs created from Wikipedia summaries of source texts by human annotators via crowd-sourcing. These pairs span $1,567$ stories where each story corresponds to either a book or a movie script. We utilize a randomly sampled set of 500 question-answer pairs from the dataset's test split. See Appendix~\ref{sec:narrativeqa_examples} for examples from the dataset. We use \modelname{} with the entire book or script as context to answer the questions using the prompt in~\cref{app:answering_prompts_narrative_qa}.

\paragraph{Ground-truth Raters.} To evaluate the correctness of a model ratings of the answers from NarrativeQA, we employ again an LLM-based rater. The rater receives the original question, the ground-truth answer(s), and a generated response, and is tasked with judging whether the response is correct. The dataset associates each question with two ground-truth answers. As it is common practice for this dataset, we rate a model response as correct if the rater evaluates it to match either of the ground-truth answers. 
We utilize two ground-truth auto-raters, namely the {\em GPT-4 Rater}, AutoAIS$_{GPT-4}$, and the {\em \modelname{} Rater}, AutoAIS$_{G15}$, where the underlying LLM is either the GPT-4 Turbo model~\cite{openai2024gpt4} or the Gemini 1.5 Pro model~\cite{gemini1-5}, respectively. The prompt is given in~\cref{app:gt_rater_prompts_narrative_qa}
A manual inspection of 300 examples shows an agreement of 95\% between the ratings provided by the raters and the ground-truth answers.

\begin{figure}[!htb]
 \centering
 \subfloat[]{%
      \includegraphics[width=0.4\textwidth]{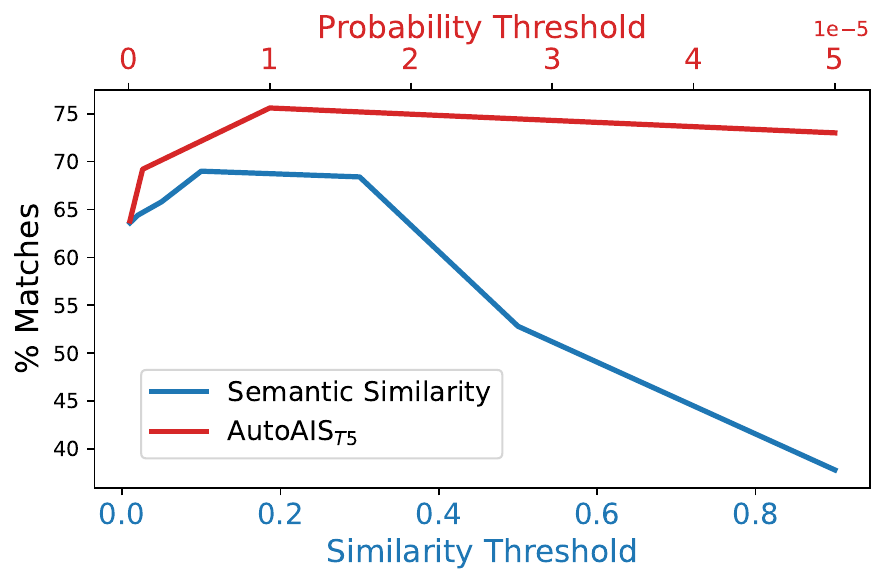}}
 \qquad
 \subfloat[]{%
      \includegraphics[width=0.4\textwidth]{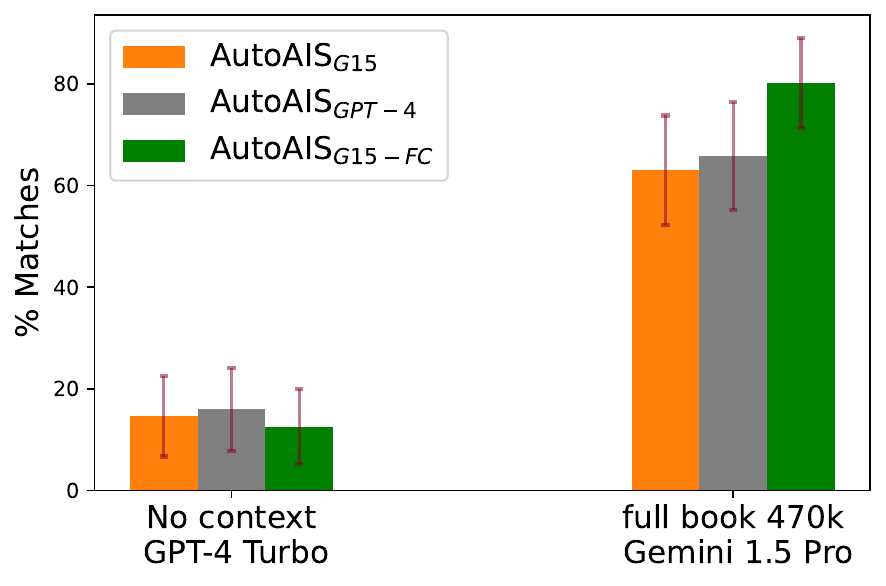}}
      \caption{Figure (a) shows the \% of times the semantic similarity rater and the AutoAIS$_{T5}$ rater agree with AutoAIS$_{GPT-4}$. Figure (b) shows the \% of times the responses of two models (No context GPT-4 Turbo and Full context Gemini 1.5 Pro) are correct as rated by the three raters.}
       \label{fig:narrative_qa_1}
\end{figure}

In comparison, two additional ground-truth raters are used as baselines: (1) an {\em AutoAIS$_{T5}$}\footnote{The model for AutoAIS$_{T5}$ is \url{https://huggingface.co/google/t5_xxl_true_nli_mixture}} rater~\cite{rashkin_ais_2023} trained specifically for rating model responses~\cite{honovich-etal-2022-true}, and (2) a simple {\em semantic similarity} rater that uses the cosine similarity metric in an embedding space (obtained via a universal sentence encoder model)~\footnote{\url{https://www.tensorflow.org/hub/tutorials/semantic_similarity_with_tf_hub_universal_encoder}} to measure the similarity of the model response to the ground-truth answers. We find that these baseline methods are less effective (see Figure~\ref{fig:narrative_qa_1} (a)), hence we use LLM-based raters as ground-truth raters for the rest of our analysis.

\paragraph{Grounding Factual Correctness.} To ground the performance of our proposed LLM-based auto-rater (as used in \cref{sec:exps}), we employ Gemini 1.5 Pro with full context, AutoAIS$_{G15-FC}$. As defined in \cref{sec:eval}, this rater takes as input the entire text, a question, and a model response, and rates the correctness of the response. We first evaluate how well this entailment rater performs compared to the ground-truth raters. The prompt is given in~\cref{app:autoAIS_rater_prompts_narrative_qa}. 

The results are shown in Figure~\ref{fig:narrative_qa_1} (b). We find strong agreement between the two ground-truth LLM raters. However, AutoAIS$_{G15-FC}$ is more optimistic in rating the responses of Gemini 1.5 Pro. We conduct a manual evaluation of the model responses and discover that many questions in the dataset contain incorrect ground-truth answers, causing the raters relying exclusively on the ground-truth answers of the Gemini 1.5 Pro model as close to $65\%$. The source of the ground-truth errors stem from the use of Wikipedia summaries to derive the answers.

In many of these erroneous cases, the model response is in indeed entailed by the full context. The AutoAIS$_{G15-FC}$ rater correctly identifies this, and hence we observe the difference in the absolute numbers as seen in Figure~\ref{fig:narrative_qa_1} (b) (see Appendix~\ref{sec:narrativeqa_noise} for several randomly picked examples).
This provides an intriguing prospect that long-context models are indeed able to perform nuanced self-evaluation.

\paragraph{Grounding Side-by-Side Evaluation and Ranking.} We next perform a side-by-side comparison and ranking with the method detailed in \cref{sec:bradley-terry}. 
We use again GPT-4 Turbo model~\cite{openai2024gpt4}, Claude 3 Opus ~\cite{claude} and \modelname{} \cite{gemini1-5}. In each case we use a setup with no context and RAG 4k.
We employ BM25~\cite{robertson2009probabilistic} as before to extract the context for each question, and the same context is used for all the models. Finally, we use \modelname{} with long context, using the entire book/movie script as input  to answer the question.

\begin{figure}[!htbp]
    \centering
    \begin{minipage}{0.49\textwidth}
        \centering
        \includegraphics[width=1.0\textwidth]{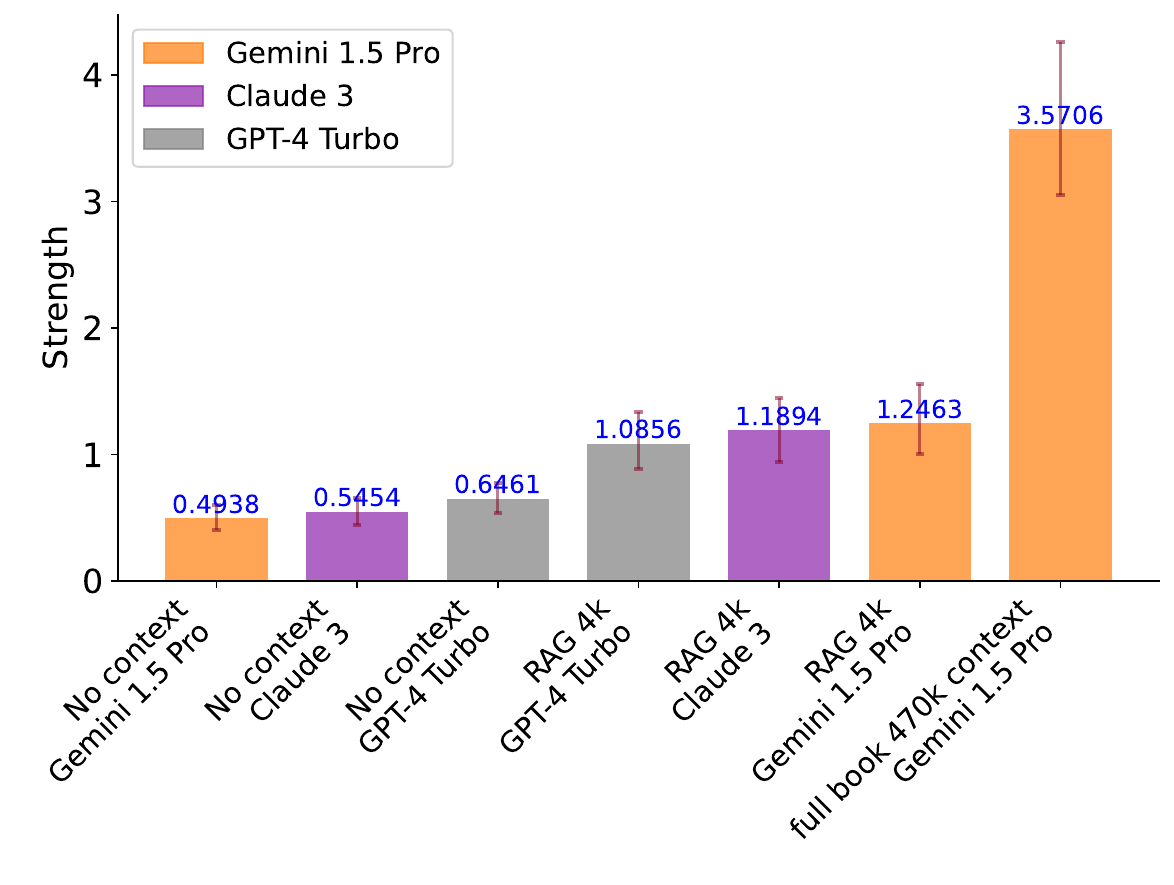} 
        \caption{Ranking based on AutoAIS$_{GPT-4}$.\label{fig:gpt_4_side_by_side}}
    \end{minipage}
    \hfill
    \begin{minipage}{0.49\textwidth}
        \centering
        \vspace{-3pt}
        \includegraphics[width=1.0\textwidth]{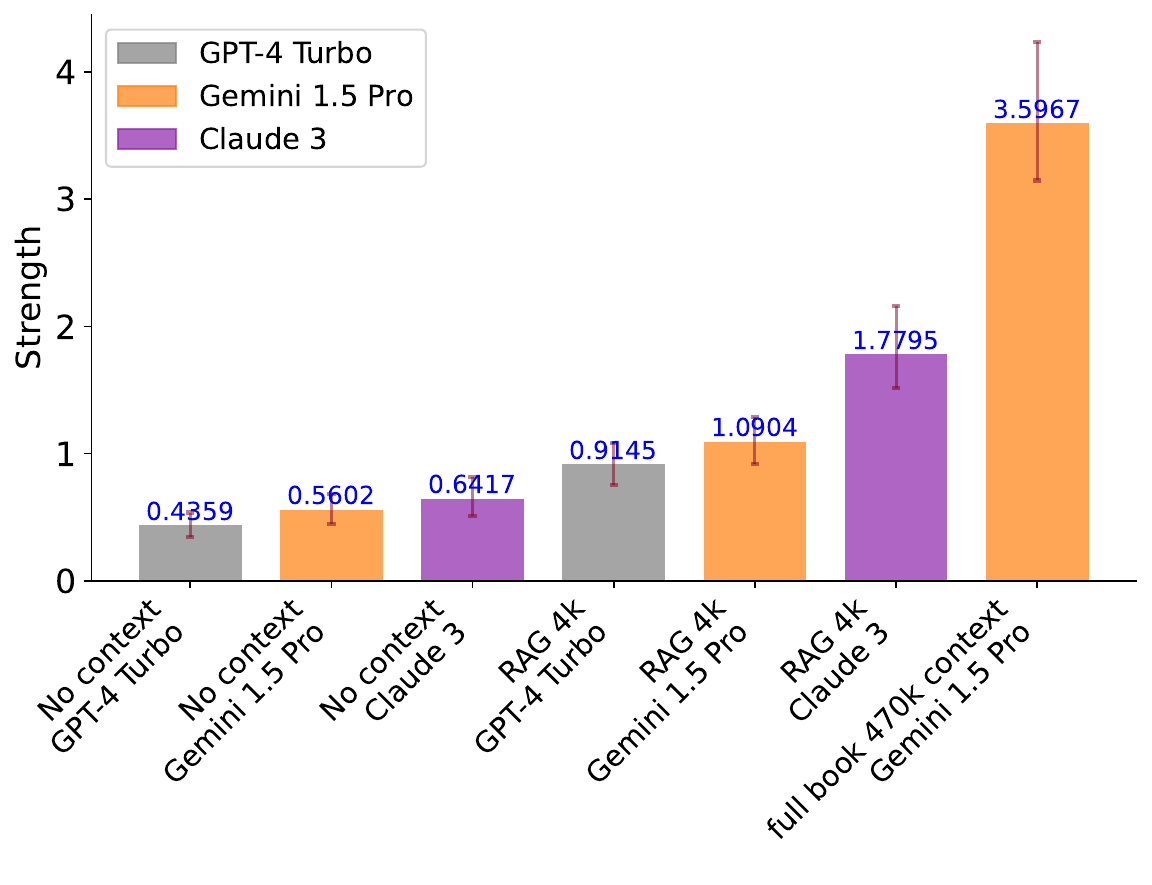} 
        \caption{Ranking based on AutoAIS$_{G15-FC}$.\label{fig:gemini_entailment_side_by_side}}
    \end{minipage}
\end{figure} 

The results are shown in Figure~\ref{fig:gpt_4_side_by_side} and Figure~\ref{fig:gemini_entailment_side_by_side}.  We see broadly consistent behavior between the ground-truth rater (Figure~\ref{fig:gpt_4_side_by_side}) and AutoAIS$_{G15-FC}$ (Figure~\ref{fig:gemini_entailment_side_by_side}). In both cases, the no context models fall towards the lower end of the ranking (with overlapping CIs), the RAG 4k models fall in the middle, and full context Gemini 1.5 Pro performs significantly better.

A notable exception is the high rating of RAG 4k Claude 3 according to the AutoAIS$_{G15-FC}$ rater. Claude 3 consistently tends to 
include additional details in its response, thereby making them more preferable (even if the other RAG 4k models are also factually correct). 
See Appendix~\ref{sec:narraative_qa_claude_4k_analysis} for examples of this behavior. Taken together, these results further validate our hypothesis that long context models are capable of generating complex questions and can self-evaluate themselves faithfully. 

%% file: section_6_related_work.tex
We introduce a new method for automatically creating long-form reading comprehension datasets and--crucially--evaluation using large language models. While numerous 
QA datasets exist to assess reading comprehension, they typically rely on human annotation and localized, factual content, limiting their applicability in long-span understanding.

\paragraph{QA datasets.} Question answering datasets have long been used in the evaluation of natural language processing, information retrieval, and other systems~\citep{yang-etal-2015-wikiqa, yang2018hotpotqa, rajpurkar2016squad, reddy2019coqa}. These datasets often involve laborious human annotation \citep{kwiatkowski2019natural,narrativeqa}, and answering them usually does not require a long span of knowledge. For example, factoid question answering \citep{chen2017reading, joshi2017triviaqa,talmor2018commonsenseqa, sakaguchi2021winogrande} only requires locating a text span in an article that contains the verbatim (or simply paraphrased) answer. Temporal QA datasets~\citep{zhang2021situatedqa, liska2022streamingqa, kasai2024realtime} contain more challenging, time-dependent answers, however still does not require long-context reasoning ability. Moreover, existing public QA datasets are almost certainly contained in the training data of modern LLMs, and hence no longer suitable for ongoing evaluation. As more capable LLMs are released, more challenging datasets are needed to properly assess their capabilities. 
    

\paragraph{LLM evaluation benchmarks.} The development of machine learning models cannot advance without proper evaluation. This is especially true for Large Language Models (LLMs), whose increasing complexity and broad range of applications demand rigorous assessment. Early work focused on task-specific benchmarks like GLUE~\cite{wang2018glue} and SuperGLUE~\cite{wang2019superglue}. However, the increasing generality of LLMs has led to the development of more comprehensive benchmarks like MMLU~\cite{hendrycks2020measuring} and BIGbench~\cite{srivastava2022beyond}, which assess performance across a wide range of tasks. Nevertheless, these benchmarks often rely on relatively short input sequences, potentially overlooking the unique challenges and capabilities associated with processing long-context inputs.

In recent years, there has been a growing interest in evaluating LLMs in the context of long documents and extended conversations. This has led to 
new benchmarks 
such as long-form question answering~\cite{fan2019eli5,narrativeqa}, long document summarization~\cite{cohan-etal-2018-discourse}, and multi-turn dialogue~\cite{budzianowski2018multiwoz}. However, the existing datasets are still not challenging enough for the state-of-the-art LLMs with 1M token context lengths. Moreover, the construction of these dataests 
involve intense 
manual labor. Instead we present, for the first time, a fully automated, LLM-assisted long-span benchmark generation framework.

%% file: section_7_conclusion.tex
This work addresses the crucial need for benchmarks 
to evaluate long-form reading comprehension of LLMs. 
We present a novel approach for automatically constructing and evaluating such benchmarks, tackling the unique challenges posed by assessing comprehension using large context sizes.
Our framework generates
challenging questions from a source text, whose answers require comprehending long spans of text and outputting multiple sentences in response. We propose both absolute and relative metrics for evaluating these responses using long-context LLMs as auto-raters.

While absolute evaluations are good for assessing factuality and general correctness, we find that relative comparisons allow the auto-rater to further emphasize answer quality, providing a more robust differentiation between models. Long-context LLMs perform extremely well on these evaluations, even against competing models with a high amount of parametric knowledge of the source text.

We analyze our approach for bias, finding moderate agreement between raters from different model families, and good performance on the NarrativeQA dataset. In fact, the long-context model was adept enough to find errors in the dataset that originated in its construction methodology (i.e., use of Wikipedia summaries).

Researchers can now build extremely ambitious and challenging long-context benchmarks that can be used to both evaluate and fine-tune future models, leading to more highly capable and useful systems that can reason over extremely long documents and media.

%% file: appendix.tex
\newcommand{\appendixhead}%
{\centering\textbf{\huge Appendix}
\vspace{0.25in}}

\appendixhead

\tableofcontents
\section{Prompts}
\label{sec:appendix_prompts}
\input{appendix/prompts}

\FloatBarrier

\section{Examples of generated questions}
\label{sec:examples_of_generated_questions}
\FloatBarrier
\input{appendix/question_examples}

\section{Examples of side-by-side comparisons}
\input{appendix/side_by_side_examples}
\FloatBarrier

\section{Example question answer pairs from the Narrative QA dataset}
\label{sec:narrativeqa_examples}
\FloatBarrier
\begin{table}[h]
    \centering
    \begin{tabular}{|p{4cm}|p{4cm}|p{4cm}|}
    \hline
         Question & Ground Truth Answer 1 & Ground Truth Answer 2 \\
         \hline
         How many people eventually commit suicide on Zuleika's behalf? & All of the undergraduates & All the Oxford undergraduates \\
         \hline
         Who does William Ashe propose to? & Lady Kitty Bristol & Lady Kitty Bristol \\
         \hline
         What was the legend of the Great Stone Face? & A man would be born that resembled the image portrayed and he would be the ``greatest and noblest personage of his time'' & That someone would be born to look like the Great Stone face and be the greatest and noblest person of his time.\\
         \hline
    \end{tabular}
    \vspace{2mm}
    \caption{Examples of question answer pairs from the NarrativeQA dataset~\cite{narrativeqa}. Each question has two ground truth answers.}
    \label{tab:narrativeqa_examples}
\end{table}

\clearpage

\section{Analysis of noise in the Narrative QA dataset}
\label{sec:narrativeqa_noise}
\begin{flushleft}
Below we analyze randomly chosen 10 examples where the ground truth LLM judges rate the response of the Gemini 1.5 Pro model as incorrect whereas AutoAIS$_{G15-FC}$ rates the response as correct. We find that in 9/10 cases the ground truth answers are either full or partially incorrect. On the other hand the responses of the Gemini 1.5 Pro model are correct and are rightly judged by AutoAIS$_{G15-FC}$.
\end{flushleft}
\cprotect\fbox{
  \begin{minipage}{0.8\textwidth}
\textbf{Q:} Where do Crockett and Tubbs face off in a firefight against Yero and the Aryan Brotherhood?\\

\textbf{Ground Truth Answers:} a) Port of Miami. b) Miami Port\\

\textbf{Model Answer:} Crockett and Tubbs face off in a firefight against Yero and the Aryan Brotherhood on the docks and bridge of the Indian River. 
\\
 
\textbf{Analysis:} This question is from the script of the movie Miami Vice. Based on the ground truth answer the model answer is deemed incorrect by a human. However based on the text the model answer is spot on and AutoAIS$_{G15-FC}$ marks it as correct.
\\

\textbf{Relevant passage from Wikipedia:}
    \medskip
    \begin{tcolorbox}[enhanced, size=fbox, fontupper=\bfseries, drop shadow southeast, 
    sharp corners, halign=left]
        \begin{Verbatim}[breaklines=true, breakanywhere=true, breaksymbol=, breakanywheresymbolpre=]
Soon afterward, Yero reveals Isabella's betrayal to Montoya and captures her. In the showdown, Crockett and Tubbs face off against Yero, his men, and the Aryan Brotherhood in a shipyard at the port of Miami.
        \end{Verbatim}
    \end{tcolorbox}

\medskip

\textbf{Relevant passages from the full text:}
    \medskip
    \begin{tcolorbox}[enhanced, size=fbox, fontupper=\bfseries, drop shadow southeast, 
    sharp corners, halign=left]
        \begin{verbatim}
EXT. ABANDONED METAL SHED, INDIAN RIVER - WIDE - NIGHT

  Nothing.  Rusting marine parts and corrugated metal fall into
  weeds next on a channel of the Indian River.  It cuts through
  inner-city Miami.  Distant noise.  But...

  INT. METAL SHED - ARYAN BROTHERS

  in low light are strapping up.  Kevlar vests go over tattooed
  arms and chests.  Glimpses of swastikas.  Handguns in
  waistbands.  SMGs (Small Machine Guns) slung from shoulders
  over white shirts, business-suit jackets.  Everybody wears
  surgical gloves.  Shoes are industrial.

  REAR SHOT:  TWO OF THE BROTHERS

  are leaving.  One carries a very long rifle with a flash
  suppressor.  It is a bolt action .50 caliber Burrows.

  NEARBY ON THE RIVER

  is a second Aryan Brother sniper team with an identical
  weapon aimed at an empty lot a thousand yards away across the
  Indian River...

  INT. METAL SHED

  Two Suburbans and an Escalade pull out...
        \end{verbatim}
    \end{tcolorbox}
  \end{minipage}
}















\cprotect\fbox{
  \begin{minipage}{0.8\textwidth}

\textbf{Q:} How does Grommet die?\\

\textbf{Ground Truth Answers:} a) He cramps and falls while rock climbing. b) Falls to his death.\\

\textbf{Model Answer:} Grommet is shot in the throat during the bank robbery and dies from blood loss. \\

\textbf{Analysis:} This question is from the movie Point Break (1991). The QA creation process incorrectly used the Wikipedia summary from the 2015 remake.
Here the ground truth answers are incorrect and the model answer is spot on. Hence human marks the model answer as incorrect. However AutoAIS$_{G15-FC}$ correctly judges.\\

\textbf{Relevant passage from Wikipedia:}
    \medskip
    \begin{tcolorbox}[enhanced, size=fbox, fontupper=\bfseries, drop shadow southeast, 
    sharp corners, halign=left]
        \begin{Verbatim}[breaklines=true, breakanywhere=true, breaksymbol=, breakanywheresymbolpre=]
Point Break (1991)
Utah is blackmailed into participating in the Ex-Presidents' last bank robbery of the summer. Grommet is killed, along with an off-duty police officer, and a bank security guard who attempt to foil the robbery. Outraged by Grommet's death, Bodhi knocks out Utah and leaves the scene.

Point Break (2015)
Utah finds the location of the next ordeal: free solo climbing with no safety beside Angel Falls in Venezuela. He finds Bodhi and Grommet and chases them on the climb, but Grommet falters, falling to his death.
        \end{Verbatim}
    \end{tcolorbox}

\medskip

\textbf{Relevant passages from the full text:}
\medskip
\begin{tcolorbox}[enhanced,size=fbox,fontupper=\bfseries,drop shadow southwest,
    sharp corners, halign=left]

\begin{verbatim} 

The cop FIRES TWICE.  Two in the chest.  Utah is hurled
back.  Slams into the tellers' counter.

NATHANIEL/LBJ lets go with the 12 gauge.  BOOM!
The cop spins, hit by some 00 buck pellets.
The guard caps-off like a maniac.  BLAM, BLAM, BLAM,
BLAM!!!
All of them wild, but serving to get Nathaniel pinned down
behind a check-writing island.

ROACH/NIXON LEAPS THE COUNTER and lands behind the guard.
The guard's shoulder explodes as Roach's 12 gauge
detonates.  He pitches onto his face, screaming and
holding the wreckage of his upper arm.
Grommet/LBJ is coming over the counter behind Roach as the
off-duty cop spins and fires.  BLAM!  BLAM!  Roach takes
it in the chest.  Staggers back, firing into the floor.
Trips and drops his shotgun.
Grommet clutches his throat as blood streams over his
fingers.

IN SLOW MOTION Reagan's face rises above the counter.
The mask is askew.  In order to see, Bodhi pulls it off.
The Casull comes up in a slow, deliberate arc as--

THE COP turns, terrified now, whipping his gun around,
nightmarishly slow as--
  \end{verbatim}
  \end{tcolorbox}
\end{minipage}
}

\cprotect\fbox{
  \begin{minipage}{0.8\textwidth}

\textbf{Q:} What does Tempest do when he actually sees Lucio for who he really is?\\

\textbf{Ground Truth Answers:} a) He renounces evil and his friendship with Lucio to go back to his life just as broke as he was to begin with or even worse. b) Gives up his fortune to repent.
\\

\textbf{Model Answer:} Tempest faints in terror and awe when he realizes Lucio is actually Lucifer.
\\
 
\textbf{Analysis:} This question is from the book sorrows of satan. Again here the ground truth answers are only partially correct and hence model response is marked incorrectly. However AutoAIS$_{G15-FC}$ correctly marks the response as correct.
 \\
 
\textbf{Relevant passage from Wikipedia:}
    \medskip
    \begin{tcolorbox}[enhanced, size=fbox, fontupper=\bfseries, drop shadow southeast, 
    sharp corners, halign=left]
        \begin{Verbatim}[breaklines=true, breakanywhere=true, breaksymbol=, breakanywheresymbolpre=]
Eventually, when confronted with the true nature of his companion, he renounces evil and returns to society penniless but content with the chance to purify his soul
        \end{Verbatim}
    \end{tcolorbox}

\medskip

\textbf{Relevant passage from the full text:}

\medskip

\begin{tcolorbox}[enhanced,size=fbox,fontupper=\bfseries,drop shadow southwest,
    sharp corners, halign=left]
\begin{verbatim}
In Chapter XLI, when Lucio reveals his true identity as Lucifer,
Tempest is overwhelmed with terror and awe.
The text describes him as 'frozen and mute'.
\end{verbatim}

\end{tcolorbox}
\end{minipage}
}

\cprotect\fbox{
  \begin{minipage}{0.8\textwidth}

\textbf{Q:} How old is Anne's youngest daughter?\\

\textbf{Ground Truth Answers:} a) 14 but almost 15. b) 15.
\\

\textbf{Model Answer:} Anne's youngest daughter, Rilla, is 19 years old.
\\
 
\textbf{Analysis:} This is a question from book rilla of ingleside.
Here again human evaluation marks the model response as incorrect. However the response is actually correct. Rilla starts off as 15 in the earlier part of the book, but towards the end is nineteen. AutoAIS$_{G15-FC}$ correctly identifies this.
 \\
 
\textbf{Relevant passage from Wikipedia:}
    \medskip
    \begin{tcolorbox}[enhanced, size=fbox, fontupper=\bfseries, drop shadow southeast, 
    sharp corners, halign=left]
        \begin{Verbatim}[breaklines=true, breakanywhere=true, breaksymbol=, breakanywheresymbolpre=]
Set almost a decade after Rainbow Valley, Europe is on the brink of the First World War, and Anne's youngest daughter Rilla is an irrepressible almost-15-year-old, excited about her first adult party and blissfully unaware of the chaos that the Western world is about to enter. Her parents worry because Rilla seems not to have any ambition, is not interested in attending college, and is more concerned with having fun.
        \end{Verbatim}
    \end{tcolorbox}

\medskip

\textbf{Relevant passages from the full text:}
\medskip
\begin{tcolorbox}[enhanced,size=fbox,fontupper=\bfseries,drop shadow southwest,
    sharp corners, halign=left]

\begin{Verbatim}[breaklines=true, breakanywhere=true, breaksymbol=, breakanywheresymbolpre=]

"It is four years tonight since the dance at the lighthouse—four years of war. It seems like three times four. I was fifteen then. I am nineteen now. I expected that these past four years would be the most delightful years of my life and they have been years of war—years of fear and grief and worry—but I humbly hope, of a little growth in strength and character as well.
"Today I was going through the hall and I heard mother saying something to father about me. I didn't mean to listen—I couldn't help hearing her as I went along the hall and upstairs—so perhaps that is why I heard what listeners are said never to hear—something good of myself. And because it was mother who said it I'm going to write it here in my journal, for my comforting when days of discouragement come upon me, in which I feel that I am vain and selfish and weak and that there is no good thing in me.
"'Rilla has developed in a wonderful fashion these past four years. She used to be such an irresponsible young creature. She has changed into a capable, womanly girl and she is such a comfort to me. Nan and Di have grown a little away from me—they have been so little at home—but Rilla has grown closer and closer to me. We are chums. I don't see how I could have got through these terrible years without her, Gilbert.'
\end{Verbatim}
\end{tcolorbox}
\end{minipage}
}

\cprotect\fbox{
  \begin{minipage}{0.8\textwidth}

\textbf{Q:} What did Blanco's brother try to discuss with him?\\

\textbf{Ground Truth Answers:} a) He tried to lecture him about morality and judgement. b) Lectured on morality and judgement
\\

\textbf{Model Answer:} Blanco's brother, Elder Daniels, tried to discuss the whereabouts of the stolen horse with him. \\
 
\textbf{Analysis:} This question is from the play the shewing-up of blanco posnet.

Again the ground truth answers are incorrect and hence model response is marked as incorrect by human judge. However AutoAIS$_{G15-FC}$ correctly judges the response.
 \\
 
\textbf{Relevant passage from Wikipedia:}
    \medskip
    \begin{tcolorbox}[enhanced, size=fbox, fontupper=\bfseries, drop shadow southeast, 
    sharp corners, halign=left]
        \begin{Verbatim}[breaklines=true, breakanywhere=true, breaksymbol=, breakanywheresymbolpre=]
His brother, a reformed drunkard who is now a church Deacon, lectures Blanco on morality and judgement, but Blanco ridicules his brother's view of God.
        \end{Verbatim}
    \end{tcolorbox}

\medskip

\textbf{Relevant passages from the full text:}
\medskip
\begin{tcolorbox}[enhanced,size=fbox,fontupper=\bfseries,drop shadow southwest,
    sharp corners, halign=left]

\begin{Verbatim}[breaklines=true, breakanywhere=true, breaksymbol=, breakanywheresymbolpre=]

BLANCO. Not a bit of it. Neither God nor Devil tempted me to take the horse: I took it on my own. He had a cleverer trick than that ready for me. [He takes his hands out of his pockets and clenches his fists]. Gosh! When I think that I might have been safe and fifty miles away by now with that horse; and here I am waiting to be hung up and filled with lead! What came to me? What made me such a fool? Thats what I want to know. Thats the great secret.
ELDER DANIELS [at the opposite side of the table] Blanco: the great secret now is, what did you do with the horse?
BLANCO [striking the table with his fist] May my lips be blighted like my soul if ever I tell that to you or any mortal men! They may roast me alive or cut me to ribbons; but Strapper Kemp shall never have the laugh on me over that job. Let them hang me. Let them shoot. So long as they are shooting a man and not a sniveling skunk and softy, I can stand up to them and take all they can give me—game.
ELDER DANIELS. Dont be headstrong, Blanco.
\end{Verbatim}
\end{tcolorbox}
\end{minipage}
}

\cprotect\fbox{
  \begin{minipage}{0.8\textwidth}

\textbf{Q:} What did Peter believe a pram to be?\\

\textbf{Ground Truth Answers:} a) animal. b) an animal.
\\

\textbf{Model Answer:} Peter believed a pram was something to sit in.  \\
 
\textbf{Analysis:} This question is from the book peter pan in kensington gardens. Here the ground truth answer is correct and model answer if wrong. But AutoAIS$_{G15-FC}$ marks it as correct and thereby makes a judgement error.
 \\
 
\textbf{Relevant passage from Wikipedia:}
    \medskip
    \begin{tcolorbox}[enhanced, size=fbox, fontupper=\bfseries, drop shadow southeast, 
    sharp corners, halign=left]
        \begin{Verbatim}[breaklines=true, breakanywhere=true, breaksymbol=, breakanywheresymbolpre=]
Throughout the novel, Peter misunderstands simple things like children's games. He does not know what a pram is, mistaking it for an animal, and he becomes extremely attached to a boy's lost kite.
        \end{Verbatim}
    \end{tcolorbox}

\medskip

\textbf{Relevant passages from the full text:}
\medskip
\begin{tcolorbox}[enhanced,size=fbox,fontupper=\bfseries,drop shadow southwest,
    sharp corners, halign=left]

\begin{Verbatim}[breaklines=true, breakanywhere=true, breaksymbol=, breakanywheresymbolpre=]
Perhaps the most surprising thing he found was a perambulator. It was under a lime-tree, near the entrance to the Fairy Queen’s Winter Palace (which is within the circle of the seven Spanish chestnuts), and Peter approached it warily, for the birds had never mentioned such things to him. Lest it was alive, he addressed it politely, and then, as it gave no answer, he went nearer and felt it cautiously. He gave it a little push, and it ran from him, which made him think it must be alive after all; but, as it had run from him, he was not afraid. So he stretched out his hand to pull it to him, but this time it ran at him, and he was so alarmed that he leapt the railing and scudded away to his boat. You must not think, however, that he was a coward, for he came back next night with a crust in one hand and a stick in the other, but the perambulator had gone, and he never saw another one. I have promised to tell you also about his paddle. It was a child’s spade which he had found near St. Govor’s Well, and he thought it was a paddle.
\end{Verbatim}
\end{tcolorbox}
\end{minipage}
}

\cprotect\fbox{
  \begin{minipage}{0.8\textwidth}

\textbf{Q:} What did the "Caller" shoot at while Stu was in the phone booth?\\

\textbf{Ground Truth Answers:} a) a toy robot. b) a toy robot.
\\

\textbf{Model Answer:} The Caller shot at the yellow pages hanging in the phone booth. \\
 
\textbf{Analysis:} This question is from the movie script phone booth.

Again the ground truth answer is incorrect and the model response is correct. AutoAIS$_{G15-FC}$ correctly judges this.
 \\
 
\textbf{Relevant passage from Wikipedia:}
    \medskip
    \begin{tcolorbox}[enhanced, size=fbox, fontupper=\bfseries, drop shadow southeast, 
    sharp corners, halign=left]
        \begin{Verbatim}[breaklines=true, breakanywhere=true, breaksymbol=, breakanywheresymbolpre=]
To demonstrate the threat, the caller fires a suppressed sniper rifle at a toy robot sold by a nearby vendor; the damage is unseen by anyone but Stu, the caller, and the vendor. 
        \end{Verbatim}
    \end{tcolorbox}

\medskip

\textbf{Relevant passages from the full text:}
\medskip
\begin{tcolorbox}[enhanced,size=fbox,fontupper=\bfseries,drop shadow southwest,
    sharp corners, halign=left]

\begin{verbatim}
Nothing to shatter when the sniper squeezes off his shot.

				VOICE
		May I call attention to the yellow
		pages?

	The frayed yellow phonebook dangling from a chain under the
	telephone shudders under the impact of a direct hit.

	There's been no sound of a gunshot, but the damage is there
	to behold.

	Stu reaches for the phonebook.

	There's a bullet hole straight through it.  Pieces of the .30
	calibre slug have shattered into many tiny fragments and are
	imbedded between the pages, half-way through the thick
	volume.

	Stu pries pieces out of the pages of the directory.  He looks
	at them in the palm of his hand.
\end{verbatim}
\end{tcolorbox}
\end{minipage}
}

\cprotect\fbox{
  \begin{minipage}{0.8\textwidth}

\textbf{Q:} How many children does Sethe have?\\

\textbf{Ground Truth Answers:} a) Three b) Three.
\\

\textbf{Model Answer:} Sethe has four children: Howard, Bulgar, Denver, and Beloved.  \\
 
\textbf{Analysis:} This is a question from the movie beloved. Sethe has another daughter who dies and hence the correct answer is indeed four. The model gets it correct and AutoAIS$_{G15-FC}$ marks it as correct as well, thereby correcting the ground truth responses.
 \\
 
\textbf{Relevant passage from Wikipedia:}
    \medskip
    \begin{tcolorbox}[enhanced, size=fbox, fontupper=\bfseries, drop shadow southeast, 
    sharp corners, halign=left]
        \begin{Verbatim}[breaklines=true, breakanywhere=true, breaksymbol=, breakanywheresymbolpre=]
Sethe is a former slave living on the outskirts of Cincinnati, Ohio shortly after the Civil War. An angry poltergeist residing in the family home terrorizes her and her three children, causing two of them to run away forever.
        \end{Verbatim}
    \end{tcolorbox}

\medskip

\textbf{Relevant passages from the full text:}
\medskip
\begin{tcolorbox}[enhanced,size=fbox,fontupper=\bfseries,drop shadow southwest,
    sharp corners, halign=left]

\begin{verbatim}
Paul turns back to find the red light in the hall is gone.

                         PAUL
              I thought you said she died soft as cream.

                         SETHE
                    (busying herself in kitchen)
              Oh that's not Baby Suggs. That's my
              daughter. The one I sent ahead with the
              boys before I run off.

                         PAUL
              She didn't live?

                         SETHE
              No.

                         PAUL
              The boys too?

                         SETHE
              No, they alive - they run off before Baby
              Suggs died. The one I was carrying when I
              left Sweet Home is all I got left.
\end{verbatim}
\end{tcolorbox}
\end{minipage}
}

\cprotect\fbox{
  \begin{minipage}{0.8\textwidth}

\textbf{Q:} Who tries unsuccessfully to get Sonny to give himself up?\\

\textbf{Ground Truth Answers:} a) mother. b) his mother.
\\

\textbf{Model Answer:} Detective Sergeant Moretti tries unsuccessfully to get Sonny to give himself up. \\
 
\textbf{Analysis:} This is a question from the movie dog day afternoon. Here both the ground truth answers and the model answers are correct. Both the mother and the detective try to get Sonny to give himself up.
 \\
 
\textbf{Relevant passage from Wikipedia:}
    \medskip
    \begin{tcolorbox}[enhanced, size=fbox, fontupper=\bfseries, drop shadow southeast, 
    sharp corners, halign=left]
        \begin{Verbatim}[breaklines=true, breakanywhere=true, breaksymbol=, breakanywheresymbolpre=]
Sonny agrees to let Mulvaney leave, but the manager refuses to leave his employees. The FBI calls Sonny out of the bank to talk to his mother, who fails to persuade him to give himself up. Back inside, Sonny dictates his will to one of the hostages, acting as a secretary, leaving money from his life insurance for Angie, and for Leon to have the surgery.
        \end{Verbatim}
    \end{tcolorbox}

\medskip

\textbf{Relevant passages from the full text:}
\medskip
\begin{tcolorbox}[enhanced,size=fbox,fontupper=\bfseries,drop shadow southwest,
    sharp corners, halign=left]

\begin{verbatim}
                                MORETTI
                     Let Sal come out, take a look.  What 
                     hope you got?  Quit while you're 
                     ahead.  All you got is attempted 
                     robbery.

                                 SONNY
                     ...armed robbery...

                                 MORETTI
                     Well, armed, then.  Nobody's been 
                     hurt.  Release the hostages, nobody 
                     is gonna worry over kidnapping 
                     charges, the worst you're gonna get 
                     is five years -- you can be out in a 
                     year.

                     Sonny stares at him, his face utterly blank.

                                 SONNY
                    Kiss me.

                    Moretti stops, stares back.
\end{verbatim}
\end{tcolorbox}
\end{minipage}
}

\cprotect\fbox{
  \begin{minipage}{0.8\textwidth}

\textbf{Q:} Why is the man dreaming of Venus told to read, Memoirs of a Suprasensual Man?\\

\textbf{Ground Truth Answers:} a) To break his infatuation with cruel women. b) To break himself fascinating cruel women.
\\

\textbf{Model Answer:} The man is told to read Memoirs of a Suprasensual Man because it details Severin's own experiences with a woman who embodies the "Venus in Furs" ideal, offering potential insight into the man's own desires and the possible consequences of pursuing them.  \\
 
\textbf{Analysis:} This question is from the book venus in furs. Here again the model answer is more correct and detailed but is judged as incorrect based solely on ground truth answers. AutoAIS$_{G15-FC}$ corrects it.
 \\
 
\textbf{Relevant passage from Wikipedia:}
    \medskip
    \begin{tcolorbox}[enhanced, size=fbox, fontupper=\bfseries, drop shadow southeast, 
    sharp corners, halign=left]
        \begin{Verbatim}[breaklines=true, breakanywhere=true, breaksymbol=, breakanywheresymbolpre=]
The framing story concerns a man who dreams of speaking to Venus about love while she wears furs. The unnamed narrator tells his dreams to a friend, Severin, who tells him how to break himself of his fascination with cruel women by reading a manuscript, Memoirs of a Suprasensual Man.
        \end{Verbatim}
    \end{tcolorbox}

\medskip

\textbf{Relevant passages from the full text:}
\medskip
\begin{tcolorbox}[enhanced,size=fbox,fontupper=\bfseries,drop shadow southwest,
    sharp corners, halign=left]

\begin{Verbatim}[breaklines=true, breakanywhere=true, breaksymbol=, breakanywheresymbolpre=]
Venus in Furs,” I cried, pointing to the picture. “That is the way I saw her in my dream.”
“I, too,” said Severin, “only I dreamed my dream with open eyes.”
“Indeed?”
“It is a tiresome story.”
“Your picture apparently suggested my dream,” I continued. “But do tell me what it means. I can imagine that it played a role in your life, and perhaps a very decisive one. But the details I can only get from you.”
“Look at its counterpart,” replied my strange friend, without heeding my question.
The counterpart was an excellent copy of Titian’s well-known “Venus with the Mirror” in the Dresden Gallery.
“And what is the significance?”
Severin rose and pointed with his finger at the fur with which Titian garbed his goddess of love.
“It, too, is a ‘Venus in Furs,’” he said with a slight smile. “I don’t believe that the old Venetian had any secondary intention. He simply painted the portrait of some aristocratic Mesalina, and was tactful enough to let Cupid hold the mirror in which she tests her majestic allure with cold satisfaction. He looks as though his task were becoming burdensome enough. The picture is painted flattery. Later an ‘expert’ in the Rococo period baptized the lady with the name of Venus. The furs of the despot in which Titian’s fair model wrapped herself, probably more for fear of a cold than out of modesty, have become a symbol of the tyranny and cruelty that constitute woman’s essence and her beauty.
\end{Verbatim}
\end{tcolorbox}
\end{minipage}
}

\section{Analysis of 4k context Claude 3 Opus on Narrative QA}
\label{sec:narraative_qa_claude_4k_analysis}

\begin{flushleft}In this section we present examples to explain the higher rating of the 4k context Claude 3 Opus model as judged by AutoAIS$_{G15-FC}$. We observe that in many cases while all the RAG 4k context models response correctly, Claude 3 consistently tends to add more detail to the answer thereby making the judge prefer its answers over the others.\end{flushleft}

\noindent\fbox{%
    \parbox{\textwidth}{%

\textbf{Q:} What happen  when the trade  ship got to water of Kush?\\

\textbf{Ground Truth Answers:} a) The trade ship was attacked  by the infamous raevers. b) IT IS ATTACKED BY THE REAVERS LED BY BELIT.
\\

\textbf{RAG 4k GPT-4 Turbo response:} When the trade ship reached the waters of Kush, it encountered pirates who investigated and damaged the ship, leading to conflict and challenges for the crew.  \\

\textbf{RAG 4k Gemini 1.5 Pro response:} The trade ship was attacked by pirates in the waters of Kush.  \\

\textbf{RAG 4k Claude 3 Opus response:} When the trade ship reached the waters of Kush, they were attacked by pirates led by Belit, the Queen of the Black Coast.  \\

}%
}

\noindent\fbox{%
    \parbox{\textwidth}{%

\textbf{Q:} What happens when Spinner gets mistaken for Mopes?\\

\textbf{Ground Truth Answers:} a) He is murdered. b) He is murdered by Green's men.\\

\textbf{RAG 4k GPT-4 Turbo response:} Spinner is dragged into the garage and knocked to the ground when mistaken for Mopes.  \\

\textbf{RAG 4k Gemini 1.5 Pro response:} Spinner gets attacked by two thugs who mistake him for Mopes.  \\

\textbf{RAG 4k Claude 3 Opus response:} According to the script, Spinner gets dragged into a garage and knocked to the ground by thugs who mistook him for Sheldon Mopes while he was wearing a Smoochy costume.\\

}%
}

\noindent\fbox{%
    \parbox{\textwidth}{%

\textbf{Q:} How is Gus killed?\\

\textbf{Ground Truth Answers:} a) Stabbed with an ice pick. b) He is Stabbed with an ice pick.\\

\textbf{RAG 4k GPT-4 Turbo response:} Gus is killed when an icepick goes into his neck.  \\

\textbf{RAG 4k Gemini 1.5 Pro response:} Gus is killed when Catherine Tramell stabs him in the neck with an ice pick.  \\

\textbf{RAG 4k Claude 3 Opus response:} Catherine Tramell stabs Gus with an icepick in the neck while he is in an elevator.\\

}%
}

\noindent\fbox{%
    \parbox{\textwidth}{%

\textbf{Q:} How do Ian, Sarah, and Kelly get off the island with the dinosaurs?\\

\textbf{Ground Truth Answers:} a) A helicopter comes to rescue them. b) A helicopter rescues them.\\

\textbf{RAG 4k GPT-4 Turbo response:} Ian, Sarah, and Kelly escape the island by scrambling toward a helicopter, which then takes off.  \\

\textbf{RAG 4k Gemini 1.5 Pro response:} Ian, Sarah, and Kelly escape the island with the dinosaurs by helicopter.\\

\textbf{RAG 4k Claude 3 Opus response:} According to the text, Ian, Sarah, and Kelly get off the island in a helicopter while being pursued by the dinosaurs.\\

}%
}

\noindent\fbox{%
    \parbox{\textwidth}{%

    
\textbf{Q:} What is Buffy destined to do?\\


\textbf{Ground Truth Answers:} a) She is destined to slay vampires. b) To kill vampires.\\

\textbf{RAG 4k GPT-4 Turbo response:} Buffy is destined to fight vampires.  \\

\textbf{RAG 4k Gemini 1.5 Pro response:} Buffy is destined to fight vampires and other evil forces.  \\


\textbf{RAG 4k Claude 3 Opus response:} Based on the context, Buffy is destined to fight vampires and other supernatural threats as the chosen Slayer.\\

}%
}

%% file: appendix/prompts.tex
\subsection{Question generating prompt}
\label{sec:appendix_question_generating_prompt}
\begin{figure}[H]
\begin{roundedbox}
\begin{lstlisting}[breaklines=true, basicstyle=\ttfamily\footnotesize, breakautoindent=false, breakindent=0pt]
{context_text}

I require {num_questions} thought-provoking questions designed to assess a comprehensive understanding of a fictional text. These questions should be crafted in a way that encourages indirect references to characters, settings, and key events, focusing on their roles rather than their explicit names. The aim is to test the reader's ability to identify and interpret these elements through their contextual importance in the narrative.

Each question should:

1. Use indirect references to characters but they should still uniquely identifiable.
2. Address the impact of specific events or decisions on the story's progression, without directly naming these events.
3. Explore themes and motifs through their representation in the narrative, rather than explicitly stating them (e.g., 'how the concept of betrayal is portrayed through the actions of key characters').
4. Analyze the narrative structure, such as the effect of the story's timeline on its unfolding, without directly citing chapter numbers or specific plot points.
5. Require drawing inferences or understanding symbolism and imagery, focusing on their effects rather than their direct descriptions.
6. Focus on your question on '{entity}' without naming it directly, please paraphrase it in still uniquely identifiable way.
7. Questions should be meticulously designed to challenge even the most attentive readers, requiring not just a superficial recall of the text but a deep and nuanced understanding of its themes, intricacies, and subtleties.
8. Format the output in the form of 'Question: <the question>' and in the next line 'Answer: <the answer>'.

Please list each question along with an answer that demonstrates deep and contextual understanding of the text and the entities who you are referring too. Go:
\end{lstlisting}
\end{roundedbox}
\caption{Question generating prompt.}
\label{fig:question_generating_prompt}
\end{figure}

\subsection{Question answering prompts}
\label{sec:question_answering_prompts}

\begin{figure}[H]
\begin{roundedbox}
\begin{lstlisting}[breaklines=true, basicstyle=\ttfamily\footnotesize, breakautoindent=false, breakindent=0pt]
The following is an Question on the book {title_and_author}.

Please provide a concise and accurate answer to the question. Your answer should be no longer than 5 sentences, but it can be shorter if the question can be fully addressed in fewer sentences. Aim for brevity and relevance in your response.
Question: {question}

Answer:
\end{lstlisting}
\end{roundedbox}
\caption{No-Context question answering prompt.}
\label{fig:question_generating_prompt}
\end{figure}

\begin{figure}[H]
\begin{roundedbox}
\begin{lstlisting}[breaklines=true, basicstyle=\ttfamily\footnotesize, breakautoindent=false, breakindent=0pt]
The book as context for the question:

{context}

Given the context of the book provided above, please provide a concise and accurate answer to the question. Your answer should be no longer than 5 sentences, but it should be shorter if the question can be fully addressed in fewer sentences. Aim for brevity and relevance in your response. Answer in full sentences, not a list.
Question: {question}

Answer:
\end{lstlisting}
\end{roundedbox}
\caption{Full-context question answering prompt.}
\label{fig:question_generating_prompt}
\end{figure}

\begin{figure}[H]
\begin{roundedbox}
\begin{lstlisting}[breaklines=true, basicstyle=\ttfamily\footnotesize, breakautoindent=false, breakindent=0pt]
Context for the question:

{context}

Given the passages of the book provided above, please provide a concise and accurate answer to the question. Your answer should be no longer than 5 sentences, but it should be shorter if the question can be fully addressed in fewer sentences. Aim for brevity and relevance in your response. Answer in full sentences, not a list.
Question: {question}

Answer:
\end{lstlisting}
\end{roundedbox}
\caption{Retrieved-context question answering prompt.}
\label{fig:question_generating_prompt}
\end{figure}

\subsection{AutoAIS absolute evaluation prompt}
\label{sec:appendix_autoais_prompt}

\begin{figure}[H]
\begin{roundedbox}
\begin{lstlisting}[breaklines=true, basicstyle=\ttfamily\footnotesize, breakautoindent=false, breakindent=0pt]
Given the context from the text (e.g., book) provided:

{context}

Evaluate if the answer to the question below is supported by the context. Your judgment should specify whether the answer is correct: 'yes' or 'no',
indicating if the answer is directly supported by the text. Also, extract literal passages from the context as evidence to support your judgment.

Format the result as JSON:
{{
'answer_is_entailed_by_context': 'yes or no',
'evidence': [
  'extracted passage(s)',
  ...
]
}}

Question: "{question}"

Answer: "{answer}"

Is this answer correct according to the context? Provide your judgment (yes or no) and the necessary evidence:
{{'answer_is_entailed_by_context':
\end{lstlisting}
\end{roundedbox}
\caption{AutoAIS absolute evaluation prompt.}
\label{fig:question_generating_prompt}
\end{figure}

\subsection{Relative comparison prompt}
\label{sec:relative_comparison_prompt}
\vspace{-9em}
\begin{figure}[H]
\begin{roundedbox}
\begin{lstlisting}[breaklines=true, basicstyle=\ttfamily\footnotesize, breakautoindent=false, breakindent=0pt]
Book as context for the following evaluation.
# CONTEXT

{context}

# CONTEXT END

# TASK

You will conduct a side-by-side evaluation. You will receive two answers for each question.

## Evaluation criteria

*Accuracy*: The primary consideration is that a system should provide only correct statements based on the above context, which need to be factually correct (not hallucinated).
If one system provides correct statements and the other does not, the system with correct statements is considered better. If both systems A and B provide statements that are not correct,
then the rating is 'None' when one or more statements in A and B are incorrect according to the context.

To evaluate the systems side by side, rank the system that fulfills the following criteria better:

*Relevance*: Do the answers directly address the questions without providing unnecessary information?
*Detail*: Are the answers detailed enough to provide a full understanding of the topic?
*Clarity*: How clear and understandable are the answers?

## Evaluation

Question: "{question}"

## System outputs:

Answer A: "{system_answer_A}"

Answer B: "{system_answer_B}"

## System rating

Please rate the systems either with 'A is better', 'B is better' or 'None'. Provide explanations and evidence for your rating. Support your explanations and evidence with excerpts from the context.

Format the result as JSON:
{{
'system is better': 'A is better|B is better|None',
'evidence': [
  'explanations and evidence supporting the rating with excerpts from the context',
  ...
]
}}
\end{lstlisting}
\end{roundedbox}
\caption{Relative comparison prompt for running side-by-side evaluations.}
\label{fig:question_generating_prompt}
\end{figure}

\subsection{Question answering prompts used for the Narrative QA dataset}
\label{app:answering_prompts_narrative_qa}

\begin{figure}[H]
\begin{roundedbox}
\begin{lstlisting}[breaklines=true, basicstyle=\ttfamily\footnotesize, breakautoindent=false, breakindent=0pt]
Here is a question related to the {movie/book} {title}.

Question: {q}

Please provide a short answer to the question in at most one sentence.

Answer:
\end{lstlisting}
\end{roundedbox}
\caption{No-context question answering prompt.}
\label{fig:zero_shot_question_answering_prompt_narrative_qa}
\end{figure}

\begin{figure}[H]
\begin{roundedbox}
\begin{lstlisting}[breaklines=true, basicstyle=\ttfamily\footnotesize, breakautoindent=false, breakindent=0pt]
Here are certain passages that are from either a movie script or a book.

{context}

Based on the above context here is a question.

Question: {q}

Please provide a short answer to the question in at most one sentence.

Answer:
\end{lstlisting}
\end{roundedbox}
\caption{Retrieved context question answering prompt.}
\label{fig:rag_question_answering_prompt_narrative_qa}
\end{figure}

\begin{figure}[H]
\begin{roundedbox}
\begin{lstlisting}[breaklines=true, basicstyle=\ttfamily\footnotesize, breakautoindent=false, breakindent=0pt]
Here is a piece of text that is either a movie script or a book.

{text}

Based on the above text here is a question.

Question: {q}

Please provide a short answer to the question in at most one sentence.

Answer:
\end{lstlisting}
\end{roundedbox}
\caption{Full context question answering prompt.}
\label{fig:full_context_question_answering_prompt_narrative_qa}
\end{figure}

\subsection{Ground truth rater prompts used for the Narrative QA dataset}
\label{app:gt_rater_prompts_narrative_qa}

\begin{figure}[H]
\begin{roundedbox}
\begin{lstlisting}[breaklines=true, basicstyle=\ttfamily\footnotesize, breakautoindent=false, breakindent=0pt]
Here is a question.

Question: {q}

Here are two answers to the question.

Answer 1: {a1}

Answer 2: {a2}

Answer 1 is the ground truth answer and Answer 2 is the proposed answer as
suggested by a student. Given that Answer 1 is the truth, judge whether Answer 2 is correct.
Answer 2 should be *very* similar to Answer 1, but may differ slightly in how it is worded.
However, Answer 2 should not directly contradict any facts or information from Answer 1.

Is Answer 2 correct? Respond with only yes or no.
\end{lstlisting}
\end{roundedbox}
\caption{Ground truth absolute rater prompt.}
\label{fig:gt_rater_prompt_narrative_qa}
\end{figure}

\begin{figure}[H]
\begin{roundedbox}
\begin{lstlisting}[breaklines=true, basicstyle=\ttfamily\footnotesize, breakautoindent=false, breakindent=0pt]
# TASK

You will conduct a side-by-side evaluation. You will receive two cadidate answers for each question.
You will also receive a groud truth answer for the question.

## Evaluation criteria

*Accuracy*: The primary consideration is that a system should provide only correct statements based on the ground truth answer.
If one system provides correct statements and the other does not, the system with correct statements is considered better. If both systems A and B provide statements that are not correct,
then the rating is 'None'. If both systems provide statements that are correct, then the rating is 'Equal'.

## Evaluation

Question: "{question}"

Ground truth answer: "{answer}"

## System outputs:

Answer A: "{system_answer_A}"

Answer B: "{system_answer_B}"

## System rating

Please rate the systems either with 'A is better', 'B is better' , 'None' or 'Equal'. Provide explanations and evidence for your rating.

Format the result as JSON:
{{
'system is better': 'A is better|B is better|None|Equal',
'evidence': [
  'explanations and evidence supporting the rating',
  ...
]
}}

Your evaluation:
{{
'system is better':
\end{lstlisting}
\end{roundedbox}
\caption{Ground truth side-by-side rater prompt.}
\label{fig:gt_side_by_side_rater_prompt_narrative_qa}
\end{figure}

\subsection{AutoAIS rater prompts used for the Narrative QA dataset}
\label{app:autoAIS_rater_prompts_narrative_qa}

\begin{figure}[H]
\begin{roundedbox}
\begin{lstlisting}[breaklines=true, basicstyle=\ttfamily\footnotesize, breakautoindent=false, breakindent=0pt]
Given the context from the text (e.g., book or movie script) provided:

{context}

Evaluate if the answer to the question below is supported by the context. Your judgment should specify whether the answer is correct: 'yes' or 'no',
indicating if the answer is directly supported by the text. Also, extract literal passages from the context as evidence to support your judgment.

Format the result as JSON:
{{
'answer_is_entailed_by_context': 'yes or no',
'evidence': [
  'extracted passage(s)',
  ...
]
}}

Question: "{question}"

Answer: "{answer}"

Is this answer correct according to the context? Provide your judgment (yes or no) and the necessary evidence:
{{'answer_is_entailed_by_context':
\end{lstlisting}
\end{roundedbox}
\caption{Auto-rater absolute rater prompt.}
\label{fig:autoAIS_rater_prompt_narrative_qa}
\end{figure}

\begin{figure}[H]
\begin{roundedbox}
\begin{lstlisting}[breaklines=true, basicstyle=\ttfamily\footnotesize, breakautoindent=false, breakindent=0pt]
Book or movie script as context for the following evaluation.
# CONTEXT

{context}

# CONTEXT END

# TASK

You will conduct a side-by-side evaluation. You will receive two candidate answers for each question.

## Evaluation criteria

*Accuracy*: The primary consideration is that a system should provide only correct statements based on the context above.
If one system provides correct statements and the other does not, the system with correct statements is considered better. If both systems A and B provide statements that are not correct,
then the rating is 'None'. If both systems provide statements that are correct, then the rating is 'Equal'.

## Evaluation

Question: "{question}"

## System outputs:

Answer A: "{system_answer_A}"

Answer B: "{system_answer_B}"

## System rating

Please rate the systems either with 'A is better', 'B is better' , 'None' or 'Equal'. Support your explanations and evidence with excerpts from the context.

Format the result as JSON:
{{
'system is better': 'A is better|B is better|None|Equal',
'evidence': [
  'explanations and evidence supporting the rating',
  ...
]
}}

Your evaluation:
{{
'system is better':
\end{lstlisting}
\end{roundedbox}
\caption{Auto-rater side-by-side evaluation prompt.}
\label{fig:autoAIS_side_by_side_rater_prompt_narrative_qa}
\end{figure}

%% file: appendix/question_examples.tex
\renewcommand{\arraystretch}{1.5} 
\newcolumntype{L}{>{\raggedright\arraybacks             lash}p{0.9\textwidth}} 

\subsection{Les Misérables Questions}
\begin{table}[th]
\centering
\begin{tabularx}{\textwidth}{@{} X @{}} 
\toprule
\textbf{Character \& Relationships} \\ 
\midrule
\begin{itemize}[leftmargin=*]
    \item How does the relationship between the "fallen woman" and the "man of God" shape the trajectory of the "redeemed soul"?
    \item How does the encounter with the "man who was not even a dog" influence the former master of the state's path towards personal redemption?
    \item How does the "child of the shadows"navigate the complexities of love and societal expectations after leaving the "austere and gloomy edifice"?
    \item How does the "man of the law" reconcile his unwavering belief in the "black and white" of justice with the "shades of gray" presented by the "man of the people"?
    \item How does the relationship between the "fallen woman" and the "man of God" shape the protagonist's journey towards redemption? 
\end{itemize}\\
\midrule
\textbf{Symbolism \& Imagery} \\
\midrule
\begin{itemize}[leftmargin=*]
    \item How does the "stolen loaf" incident impact the protagonist's perception of justice and shape his future actions?
    \item How does the "silverware" symbolize the protagonist's internal struggle between his past and his desire for a new life?
    \item How does the "underground labyrinth" function as both a refuge and a symbol of the protagonist's internal struggles?
    \item How does the recurring image of light and darkness function as a symbol throughout the narrative, reflecting the characters' internal struggles and the broader societal conflicts?
    \item How does the author employ the imagery of "light" and "darkness" to symbolize the struggle between good and evil, both within individuals and in society as a whole? 
\end{itemize}\\
\midrule
\textbf{Theme \& Social Commentary} \\
\midrule
\begin{itemize}[leftmargin=*]
    \item How is the theme of societal injustice explored through the contrasting experiences of the protagonist and the female characters in the story?
    \item How does the author utilize the setting of the Parisian underworld, with its unique language and customs, to explore the hidden realities of poverty and social exclusion?
    \item How is the theme of societal injustice explored through the contrasting experiences of two groups of characters: those who have transgressed the law and those who have not?
    \item How does the "innocent child who was hung up by the armpits" challenge the notion of divine right and the inherent goodness of those in power?
    \item How is the theme of societal injustice explored through the experiences of a young woman forced into a life of hardship and degradation? 
\end{itemize}\\
\bottomrule
\end{tabularx}
\end{table}

\begin{table}[th]
\centering
\begin{tabularx}{\textwidth}{@{} X @{}} 
\toprule
\textbf{Narrative \& Structure} \\
\midrule
\begin{itemize}[leftmargin=*]
    \item How does the narrative structure, with its frequent shifts in time and perspective, contribute to the reader's understanding of the characters' motivations and the complex web of interconnected destinies?
    \item How does the non-linear narrative structure, with its frequent flashbacks and shifts in perspective, contribute to the reader's understanding of the characters' motivations and the complex web of relationships that bind them together?
    \item How does the "revolution arrested midway" reflect the ongoing struggle between societal progress and the forces of the past?
\end{itemize}\\
\midrule
\textbf{Sacrifice \& Redemption} \\
\midrule
\begin{itemize}[leftmargin=*]
    \item How does the protagonist's ultimate sacrifice, driven by his love for a young woman, demonstrate the transformative power of love and its ability to redeem even the most broken souls?
    \item How does the "final sacrifice" of the protagonist illuminate the themes of love, redemption, and the enduring power of the human spirit?
\end{itemize}\\
\bottomrule
\end{tabularx}
\vspace{3pt}
\caption{20 examples of generated questions for Les Misérables, grouped by different question themes, analyzed post hoc.}
\end{table}

\FloatBarrier

\subsection{The Wild Huntress Questions}
\begin{table}[th]
\centering
\begin{tabularx}{\textwidth}{@{} X @{}}
\toprule
\textbf{Character Relationships \& Dynamics} \\
\midrule
\begin{itemize}[leftmargin=*]
    \item How does the narrative portray the complex relationship between the two sisters, highlighting their contrasting personalities and the impact of their shared past?
    
    \item How does the protagonist's initial encounter with the younger sibling transform his perspective on life and influence his subsequent actions?
    
    \item How does the narrative portray the complex relationship between the protagonist and the elder sibling of their beloved, highlighting the shift from initial mistrust to a strong alliance?
\end{itemize}\\
\midrule
\textbf{Betrayal \& Deception} \\
\midrule
\begin{itemize}[leftmargin=*]
    \item How does the concept of betrayal manifest through the actions of key characters, particularly the religious figure and the father, and how does it impact the lives of the two sisters?
    
    \item How does the story explore the concept of betrayal through the actions of key characters, particularly those who manipulate others for personal gain?
    
    \item How does the recurring motif of the "wolf" symbolize the predatory nature of certain characters and the dangers faced by the sisters, particularly the younger one?
\end{itemize}\\
\midrule
\textbf{Setting \& Atmosphere} \\
\midrule
\begin{itemize}[leftmargin=*]
    \item How does the setting of the "mountain parks," with its contrasting landscapes of beauty and danger, reflect the conflicting emotions and experiences of the characters?

    \item How does the story's setting, particularly the contrast between the wilderness and settlements, contribute to the development of key themes?

    \item How does the setting of the American West, with its vast landscapes and encounters with both natural and human dangers, contribute to the themes of isolation, resilience, and the pursuit of happiness?
\end{itemize}\\
\midrule
\textbf{Narrative Structure \& Symbolism} \\
\midrule
\begin{itemize}[leftmargin=*]
    \item How does the story's non-linear timeline, with its shifts between past and present, contribute to the unfolding of the narrative and the development of key characters?

    \item How does the narrative structure, with its shifting perspectives and interwoven storylines, contribute to the suspense and intrigue of the plot?

    \item How does the symbolism of the "bell-flower" contribute to the development of the romantic relationship between the protagonist and the younger sibling, and how does its fate foreshadow future challenges?
\end{itemize}\\
\bottomrule
\end{tabularx}
\end{table}

\begin{table}[th]
\centering
\begin{tabularx}{\textwidth}{@{} X @{}}
\toprule
\textbf{Individual Character Arcs \& Motivations} \\
\midrule
\begin{itemize}[leftmargin=*]
    \item How does the younger sibling's act of leaving a written message reveal her character and influence the course of events?

    \item How does the impulsive act of affection in the flowery glade set in motion a chain of events that alters the course of several lives?

    \item The military man who embarks on a journey driven by love rather than gold, how does his past experience shape his approach to the challenges he faces in the wilderness?
\end{itemize}\\
\midrule
\textbf{Broader Themes \& Messages} \\
\midrule
\begin{itemize}[leftmargin=*]
    \item How does the story explore the clash between traditional values and the allure of new ideologies, particularly through the portrayal of the Mormon community?

    \item How does the story explore the tension between civilized society and the wilderness, particularly through the experiences of the narrator and the contrasting lifestyles of the two families?

    \item How does the story explore the theme of redemption through the actions of characters who confront their past mistakes and seek to make amends?
\end{itemize}\\
\midrule
\textbf{Specific Details} \\
\midrule
\begin{itemize}[leftmargin=*]
    \item How does the protagonist's encounter with the dead horse near the squatter's abandoned cabin serve as a crucial turning point in the narrative, providing vital information and intensifying the urgency of their pursuit?

    \item How does the elder sibling's transformation into a "wild huntress" reflect her resilience and determination in the face of adversity?
\end{itemize}\\
\bottomrule
\end{tabularx}
\vspace{3pt}
\caption{20 examples of generated questions for The Wild Huntress, grouped by different question themes, analyzed post hoc.}
\end{table}
\FloatBarrier

%% file: appendix/side_by_side_examples.tex
\subsection{Les Misérables Example 1: Gemini 1.5 Pro Full-Context vs GPT 4 Turbo No-Context}
\begin{minipage}{0.9\linewidth}
\begin{tcolorbox}[enhanced, colback=white, colframe=black, title=Question, sharp corners]
How does the relationship between the "fallen woman" and the "man of God" shape the trajectory of the "redeemed soul"?
\end{tcolorbox}
\end{minipage}
\begin{minipage}{0.9\linewidth}
\begin{tcolorbox}[enhanced, colback=white, colframe=black, title=System A: Gemini 1.5 Pro (Full-Context) Answer, sharp corners]
The relationship between Fantine, the "fallen woman," and Bishop Myriel, the "man of God," profoundly shapes Jean Valjean's path to redemption. Bishop Myriel's act of mercy and forgiveness, exemplified by his gifting of the silver candlesticks to Jean Valjean despite the theft, awakens a dormant conscience in the ex-convict.  Fantine's suffering and tragic fate, a consequence of societal injustices, further fuels Jean Valjean's desire to live a virtuous life. He vows to become an honest man and dedicates himself to helping others, particularly Cosette, Fantine's daughter.  Thus, Fantine and Bishop Myriel, though absent for much of the narrative, act as guiding forces in Jean Valjean's journey toward moral and spiritual transformation.
\end{tcolorbox}
\end{minipage}

\begin{minipage}{0.9\linewidth}
\begin{tcolorbox}[enhanced, colback=white, colframe=black, title=System B: GPT-4 Turbo (No-Context) Answer, sharp corners]
In "Les Misérables," the relationship between the "fallen woman," Fantine, and the "man of God," Bishop Myriel, significantly influences the trajectory of the "redeemed soul," Jean Valjean. Although Bishop Myriel and Fantine never meet, the Bishop's act of mercy towards Valjean sets the foundation for Valjean's transformation and redemption. Inspired by the Bishop's compassion, Valjean resolves to live a life of kindness and integrity. This change in Valjean directly affects Fantine, as he promises to care for her daughter, Cosette, ultimately altering both their lives. Thus, the Bishop's influence on Valjean extends to Fantine and shapes Valjean's journey from a hardened convict to a compassionate guardian, illustrating the ripple effect of grace and redemption.
\end{tcolorbox}
\end{minipage}

\begin{minipage}{0.9\linewidth}
\begin{tcolorbox}[enhanced, colback=white, colframe=black, title=Model Analysis, sharp corners]
\begin{Verbatim}[breaklines=true, breakanywhere=true, breaksymbol=, breakanywheresymbolpre=]
"{
'system is better': 'A is better',
'evidence': [
  'Answer A is more accurate in its portrayal of the relationship between Fantine and Jean Valjean. It correctly states that Fantine\'s suffering and tragic fate, witnessed by Jean Valjean, contribute to his desire for a virtuous life.  Answer B incorrectly states that Bishop Myriel and Fantine never meet.  They do, in fact, meet when Jean Valjean, as Monsieur Madeleine, rescues Fantine from being arrested by Javert.',
  'Excerpt from the text supporting A: ""So the monk\'s widow was good for something. But M. Madeleine had heard nothing of all this. Life is full of just such combinations of events. M. Madeleine was in the habit of almost never entering the women\'s workroom.""',
  'Excerpt from the text supporting A: ""Jean Valjean left the town as though he were fleeing from it. He set out at a very hasty pace through the fields, taking whatever roads and paths presented themselves to him, without perceiving that he was incessantly retracing his steps. He wandered thus the whole morning, without having eaten anything and without feeling hungry. He was the prey of a throng of novel sensations. He was conscious of a sort of rage; he did not know against whom it was directed. He could not have told whether he was touched or humiliated. There came over him at moments a strange emotion which he resisted and to which he opposed the hardness acquired during the last twenty years of his life. This state of mind fatigued him. He perceived with dismay that the sort of frightful calm which the injustice of his misfortune had conferred upon him was giving way within him. He asked himself what would replace this. At times he would have actually preferred to be in prison with the gendarmes, and that things should not have happened in this way; it would have agitated him less. Although the season was tolerably far advanced, there were still a few late flowers in the hedge-rows here and there, whose odor as he passed through them in his march recalled to him memories of his childhood. These memories were almost intolerable to him, it was so long since they had recurred to him. Unutterable thoughts assembled within him in this manner all day long."" '
]
}"
\end{Verbatim}
\end{tcolorbox}
\end{minipage}

\subsection{Les Misérables Example 2: Gemini 1.5 Pro Full-Context vs Gemini 1.5 Pro 4k Retrieved-Context}
\begin{minipage}{0.9\linewidth}
\begin{tcolorbox}[enhanced, colback=white, colframe=black, title=Question, sharp corners]
How does the relationship between the "fallen woman" and the "man of God" shape the trajectory of the "redeemed soul"?
\end{tcolorbox}
\end{minipage}
\begin{minipage}{0.9\linewidth}
\begin{tcolorbox}[enhanced, colback=white, colframe=black, title=System A: Gemini 1.5 Pro (Full-Context) Answer, sharp corners]
The relationship between Fantine, the "fallen woman," and Bishop Myriel, the "man of God," profoundly shapes Jean Valjean's path to redemption. Bishop Myriel's act of mercy and forgiveness, exemplified by his gifting of the silver candlesticks to Jean Valjean despite the theft, awakens a dormant conscience in the ex-convict.  Fantine's suffering and tragic fate, a consequence of societal injustices, further fuels Jean Valjean's desire to live a virtuous life. He vows to become an honest man and dedicates himself to helping others, particularly Cosette, Fantine's daughter.  Thus, Fantine and Bishop Myriel, though absent for much of the narrative, act as guiding forces in Jean Valjean's journey toward moral and spiritual transformation.
\end{tcolorbox}
\end{minipage}

\begin{minipage}{0.9\linewidth}
\begin{tcolorbox}[enhanced, colback=white, colframe=black, title=System B: Gemini 1.5 Pro (4k Retrieved-Context) Answer, sharp corners]
The text provided does not mention a "fallen woman" or a "man of God" in the context of a "redeemed soul." Therefore, I cannot answer your question using the given text. Please provide the relevant passages for an accurate response.
\end{tcolorbox}
\end{minipage}

\begin{minipage}{0.9\linewidth}
\begin{tcolorbox}[enhanced, colback=white, colframe=black, title=Model Analysis, sharp corners]
\begin{Verbatim}[breaklines=true, breakanywhere=true, breaksymbol=, breakanywheresymbolpre=]
"{
 'system is better': 'A is better',
 'evidence': [
  'Answer A accurately identifies the ""fallen woman"" as Fantine and the ""man of God"" as Bishop Myriel, and correctly explains their impact on Jean Valjean\'s redemption.',
  'Answer B claims the text does not mention these figures in the context of a ""redeemed soul,"" which is incorrect. The entire first volume, ""Fantine,"" details this relationship and its impact on Jean Valjean.',
  'For example, the text states: ""What had taken place in that soul? ...He was, as we have said, an ignorant man, but he was not a fool. The light of nature was ignited in him. Unhappiness, which also possesses a clearness of vision of its own, augmented the small amount of daylight which existed in this mind. Beneath the cudgel, beneath the chain, in the cell, in hardship, beneath the burning sun of the galleys, upon the plank bed of the convict, he withdrew into his own consciousness and meditated."" This passage, and many others throughout the first volume, show how Fantine\'s suffering and the Bishop\'s mercy contribute to Jean Valjean\'s internal transformation.',
  'Answer A is also more relevant, detailed, and clear in its explanation.'
 ]
}"
\end{Verbatim}
\end{tcolorbox}
\end{minipage}

\subsection{Les Misérables Example 3: Claude 3 4k Retrieved-Context vs Gemini 1.5 Pro Full-Context}
\begin{minipage}{0.9\linewidth}
\begin{tcolorbox}[enhanced, colback=white, colframe=black, title=Question, sharp corners]
How does the relationship between the "fallen woman" and the "man of God" shape the trajectory of the "redeemed soul"?
\end{tcolorbox}
\end{minipage}
\begin{minipage}{0.9\linewidth}
\begin{tcolorbox}[enhanced, colback=white, colframe=black, title=System A: Claude 3 (4k Retrieved-Context) Answer, sharp corners]
Based on the passages provided, the relationship between Jean Valjean, a former criminal, and the Bishop shapes Jean Valjean's trajectory of redemption. The Bishop's kindness and forgiveness towards Jean Valjean, despite his past, sparks a transformation in Jean Valjean's soul. The Bishop sees the divine potential in Jean Valjean and encourages him to become an honest man, symbolically "buying" his soul for God. This act of grace sets Jean Valjean on a path of redemption, where he struggles with his conscience but ultimately chooses to live a life of virtue and selflessness.
\end{tcolorbox}
\end{minipage}

\begin{minipage}{0.9\linewidth}
\begin{tcolorbox}[enhanced, colback=white, colframe=black, title=System B: Gemini 1.5 Pro (Full-Context) Answer, sharp corners]
The relationship between Fantine, the "fallen woman," and Bishop Myriel, the "man of God," profoundly shapes Jean Valjean's path to redemption. Bishop Myriel's act of mercy and forgiveness, exemplified by his gifting of the silver candlesticks to Jean Valjean despite the theft, awakens a dormant conscience in the ex-convict.  Fantine's suffering and tragic fate, a consequence of societal injustices, further fuels Jean Valjean's desire to live a virtuous life. He vows to become an honest man and dedicates himself to helping others, particularly Cosette, Fantine's daughter.  Thus, Fantine and Bishop Myriel, though absent for much of the narrative, act as guiding forces in Jean Valjean's journey toward moral and spiritual transformation.
\end{tcolorbox}
\end{minipage}

\begin{minipage}{0.9\linewidth}
\begin{tcolorbox}[enhanced, colback=white, colframe=black, title=Model Analysis, sharp corners]
\begin{Verbatim}[breaklines=true, breakanywhere=true, breaksymbol=, breakanywheresymbolpre=]
"{
'system is better': 'B is better',
'evidence': [
  'Answer B is more relevant to the question, as it accurately identifies both Fantine and Bishop Myriel as the ""fallen woman"" and ""man of God"" respectively, and explains their influence on Jean Valjean\'s redemption.',
  'Answer A focuses solely on the Bishop\'s influence, neglecting Fantine\'s significant role. While the Bishop\'s act of grace is crucial, Fantine\'s suffering and Jean Valjean\'s promise to her are equally important motivators for his transformation.',
  'Answer B provides more detail and clarity by highlighting specific examples, such as the gifting of the candlesticks and Jean Valjean\'s dedication to Cosette. This provides a more comprehensive understanding of the complex dynamics at play.',
  'For example, the text states: ""The man was ruined. He was shortly to be tried at Aix with his accomplice. They were relating the matter, and each one was expressing enthusiasm over the cleverness of the magistrate. By bringing jealousy into play, he had caused the truth to burst forth in wrath, he had educed the justice of revenge. The Bishop listened to all this in silence. When they had finished, he inquired,-- \'Where are this man and woman to be tried?\'""',
  'This passage demonstrates the Bishop\'s unique perspective on justice and mercy, which deeply impacts Jean Valjean. Similarly, Fantine\'s plight and Jean Valjean\'s promise to her are central to his journey: ""Jean Valjean, my brother, you no longer belong to evil, but to good. It is your soul that I buy from you; I withdraw it from black thoughts and the spirit of perdition, and I give it to God.""',
  'Answer B captures these nuances more effectively, making it a better response.'
]
}"
\end{Verbatim}
\end{tcolorbox}
\end{minipage}

\subsection{The Wild Huntress Example 1: Gemini 1.5 Pro Full-Context vs GPT 4 Turbo No-Context}
\begin{minipage}{0.9\linewidth}
\begin{tcolorbox}[enhanced, colback=white, colframe=black, title=Question, sharp corners]
How does the protagonist's encounter with the huntress in the mountain ravine reveal a surprising connection to their own pursuit and introduce a new layer of complexity to the narrative?
\end{tcolorbox}
\end{minipage}
\begin{minipage}{0.9\linewidth}
\begin{tcolorbox}[enhanced, colback=white, colframe=black, title=System A: Gemini 1.5 Pro Full-Context Answer, sharp corners]
The protagonist's encounter with the huntress, Marian, in the mountain ravine reveals a surprising connection to his pursuit of Lilian as she is revealed to be Lilian's sister. Marian's story of being deceived and nearly abducted by a Mormon named Stebbins confirms the protagonist's suspicions about Lilian's situation. This connection not only deepens the protagonist's determination to rescue Lilian but also adds a layer of personal urgency and emotional complexity to the narrative. Marian's presence as a skilled huntress and knowledgeable ally further complicates the rescue mission, as her involvement carries significant risks due to her past relationship with Stebbins.
\end{tcolorbox}
\end{minipage}
\begin{minipage}{0.9\linewidth}
\begin{tcolorbox}[enhanced, colback=white, colframe=black, title=System B: GPT-4 Turbo No-Context Answer, sharp corners]
In "The Wild Huntress," the protagonist's encounter with the huntress in the mountain ravine reveals a surprising connection as they discover shared objectives in their respective pursuits. This encounter not only adds a personal dimension to their goals but also introduces romantic elements and moral dilemmas, thereby complicating the narrative. The protagonist, initially on a pursuit distinct from the huntress, finds their paths intertwined, which magnifies the stakes and challenges of their adventures. This intersection of goals and burgeoning relationships deepens the plot and enhances the complexity of the story, leading to new conflicts and alliances that drive the narrative forward.
\end{tcolorbox}
\end{minipage}
\begin{minipage}{0.9\linewidth}
\begin{tcolorbox}[enhanced, colback=white, colframe=black, title=Model Analysis, sharp corners]
\begin{Verbatim}[breaklines=true, breakanywhere=true, breaksymbol=, breakanywheresymbolpre=]
{
'system is better': 'A is better',
'evidence': [
'Answer A accurately identifies the huntress as Marian Holt and correctly states that she is Lilian's sister, as revealed in Chapter 71: ""I think it high time they looked to their own lives, without being so particular about that of their prisoner. Santissima Virgen! As I live, there's a woman among them!"" ""Yes--I see her--I know her. Her presence explains why they are taking him alive."" ""You know her?"" ""And him too. Poor fellow! I hope she will befriend him; but--""',
'Answer A correctly points out that Marian's story confirms the protagonist's suspicions about Lilian's situation, as detailed in Chapter 76: ""Virgen Santa! for the harem of the Mormon prophet!""',
'Answer A accurately mentions the risks associated with Marian's involvement due to her past relationship with Stebbins, as highlighted in Chapter 89: ""As for Marian Holt, hers would be a position of positive danger--even equalling that in which her sister was now placed. Stebbins could claim her--if not by a true husband's right, at least by the laws of Mormon matrimony; and of course by those laws would the case be judged in a Mormon camp--the apostle himself being their interpreter!""',
'Answer B provides a more generic and less accurate account of the encounter. It fails to identify the huntress as Marian and doesn't mention the crucial connection to Stebbins and the Mormon plot. While it touches upon the introduction of romantic elements and moral dilemmas, it lacks the specific details and accuracy present in Answer A.'
]
}
\end{Verbatim}
\end{tcolorbox}
\end{minipage}

\subsection{The Wild Huntress Example 2: Gemini 1.5 Pro Full-Context vs GPT 4 Turbo No-Context}
\begin{minipage}{0.9\linewidth}
\begin{tcolorbox}[enhanced, colback=white, colframe=black, title=Question, sharp corners]
How does the elder sister's transformation from a victim of betrayal to a skilled huntress symbolize resilience and the power of self-reliance in overcoming adversity?
\end{tcolorbox}
\end{minipage}
\begin{minipage}{0.9\linewidth}
\begin{tcolorbox}[enhanced, colback=white, colframe=black, title=System A: Gemini 1.5 Pro Full-Context, sharp corners]
Marian Holt's transformation into a skilled huntress symbolizes her resilience and strength in the face of betrayal and adversity. After being deceived by her lover and forced into a sham marriage, she escapes into the wilderness and embraces a life of self-reliance. Through her mastery of hunting and survival skills, she reclaims her agency and finds empowerment in the face of hardship. This transformation highlights the potential for personal growth and the ability to overcome challenges through self-determination. Marian's journey embodies the idea that even in the face of betrayal and loss, one can find strength and forge a new path.
\end{tcolorbox}
\end{minipage}
\begin{minipage}{0.9\linewidth}
\begin{tcolorbox}[enhanced, colback=white, colframe=black, title=System B: GPT-4 Turbo No-Context, sharp corners]
In "The Wild Huntress" by Mayne Reid, the elder sister's transformation from a victim of betrayal to a skilled huntress symbolizes resilience and self-reliance by showcasing her ability to adapt and thrive in a challenging environment. After being betrayed, she does not succumb to victimhood but instead learns hunting skills that enable her to survive and protect herself in the wilderness. This shift not only illustrates her personal growth and empowerment but also highlights the theme that adversity can be a catalyst for developing strength and self-sufficiency. Her mastery of hunting, traditionally seen as a male domain, further underscores her break from dependency towards independence and self-reliance. This transformation is emblematic of her reclaiming control over her life and destiny.
\end{tcolorbox}
\end{minipage}
\begin{minipage}{0.9\linewidth}
\begin{tcolorbox}[enhanced, colback=white, colframe=black, title=Model Analysis, sharp corners]
\begin{Verbatim}[breaklines=true, breakanywhere=true, breaksymbol=, breakanywheresymbolpre=]
B is better
Both answers accurately capture the essence of Marian's transformation and how it symbolizes resilience and self-reliance. However, Answer B provides a more nuanced and detailed analysis.
Answer B specifically highlights how Marian's mastery of hunting, "traditionally seen as a male domain," underscores her shift from dependency to independence. This detail, rooted in the context of the story, adds a layer of depth to the analysis that Answer A lacks.
For example, the text states: "Although there is nothing masculine in this young girl’s beauty, a single glance at her features impresses you with the idea of a character of no ordinary kind—a nature more resolute than tender—a heart endowed with courage equalling that of a man." This excerpt supports Answer B's point about Marian embracing a traditionally male domain.
\end{Verbatim}
\end{tcolorbox}
\end{minipage}

\subsection{The Wild Huntress Example 3:\modelname{} Full-Context vs Claude 3 No-Context}
\begin{minipage}{0.9\linewidth}
\begin{tcolorbox}[enhanced, colback=white, colframe=black, title=Question, sharp corners]
How does the protagonist's encounter with the huntress in the mountain ravine reveal a surprising connection to their own pursuit and introduce a new layer of complexity to the narrative?
\end{tcolorbox}
\end{minipage}
\begin{minipage}{0.9\linewidth}
\begin{tcolorbox}[enhanced, colback=white, colframe=black, title=System A: Gemini 1.5 Pro Full-Context, sharp corners]
The protagonist's encounter with the huntress, Marian, in the mountain ravine reveals a surprising connection to his pursuit of Lilian as she is revealed to be Lilian's sister. Marian's story of being deceived and nearly abducted by a Mormon named Stebbins confirms the protagonist's suspicions about Lilian's situation. This connection not only deepens the protagonist's determination to rescue Lilian but also adds a layer of personal urgency and emotional complexity to the narrative. Marian's presence as a skilled huntress and knowledgeable ally further complicates the rescue mission, as her involvement carries significant risks due to her past relationship with Stebbins.
\end{tcolorbox}
\end{minipage}
\begin{minipage}{0.9\linewidth}
\begin{tcolorbox}[enhanced, colback=white, colframe=black, title=System B: Claude 3 No-Context, sharp corners]
In the mountain ravine, the protagonist unexpectedly encounters the mysterious huntress they have been pursuing. This meeting reveals that the huntress is actually the protagonist's long-lost sister, adding a surprising familial connection to their chase. The discovery introduces a new layer of complexity to the narrative, as the protagonist must now navigate both the external challenges of the hunt and the internal emotional turmoil of this revelation. This unexpected twist deepens the story's themes of identity, family, and the blurring of lines between hunter and hunted.
\end{tcolorbox}
\end{minipage}
\begin{minipage}{0.9\linewidth}
\begin{tcolorbox}[enhanced, colback=white, colframe=black, title=Model Analysis, sharp corners]
\begin{Verbatim}[breaklines=true, breakanywhere=true, breaksymbol=, breakanywheresymbolpre=]
{
'system is better': 'A is better',
'evidence': [
'Answer A accurately identifies the huntress as Marian Holt and correctly states that she is Lilian's sister, as revealed in Chapter 71: ""I think it high time they looked to their own lives, without being so particular about that of their prisoner. Santissima Virgen! As I live, there's a woman among them!"" ""Yes--I see her--I know her. Her presence explains why they are taking him alive."" ""You know her?"" ""And him too. Poor fellow! I hope she will befriend him; but--""',
'Answer A correctly points out that Marian's story confirms the protagonist's suspicions about Lilian's situation, as detailed in Chapter 76: ""Virgen Santa! for the harem of the Mormon prophet!""',
'Answer A accurately mentions the risks associated with Marian's involvement due to her past relationship with Stebbins, as highlighted in Chapter 89: ""As for Marian Holt, hers would be a position of positive danger--even equalling that in which her sister was now placed. Stebbins could claim her--if not by a true husband's right, at least by the laws of Mormon matrimony; and of course by those laws would the case be judged in a Mormon camp--the apostle himself being their interpreter!""',
'Answer B provides a more generic and less accurate account of the encounter. It fails to identify the huntress as Marian and doesn't mention the crucial connection to Stebbins and the Mormon plot. While it touches upon the introduction of romantic elements and moral dilemmas, it lacks the specific details and accuracy present in Answer A.'
]
}
\end{Verbatim}
\end{tcolorbox}
\end{minipage}

\subsection{The Wild Huntress Example 4:\modelname{} Full-Context vs Claude 3 No-Context}
\begin{minipage}{0.9\linewidth}
\begin{tcolorbox}[enhanced, colback=white, colframe=black, title=Question, sharp corners]
How does the elder sister's transformation from a victim of betrayal to a skilled huntress symbolize resilience and the power of self-reliance in overcoming adversity?
\end{tcolorbox}
\end{minipage}
\begin{minipage}{0.9\linewidth}
\begin{tcolorbox}[enhanced, colback=white, colframe=black, title=System A: Gemini 1.5 Pro Full-Context, sharp corners]
Marian Holt's transformation into a skilled huntress symbolizes her resilience and strength in the face of betrayal and adversity. After being deceived by her lover and forced into a sham marriage, she escapes into the wilderness and embraces a life of self-reliance. Through her mastery of hunting and survival skills, she reclaims her agency and finds empowerment in the face of hardship. This transformation highlights the potential for personal growth and the ability to overcome challenges through self-determination. Marian's journey embodies the idea that even in the face of betrayal and loss, one can find strength and forge a new path.
\end{tcolorbox}
\end{minipage}
\begin{minipage}{0.9\linewidth}
\begin{tcolorbox}[enhanced, colback=white, colframe=black, title=System B: Claude 3 No-Context, sharp corners]
In The Wild Huntress, the elder sister's transformation symbolizes resilience and self-reliance as she overcomes the betrayal and abandonment by her lover. She channels her pain and anger into mastering the skills of hunting, becoming fiercely independent and self-sufficient in the wilderness. This transformation showcases her inner strength and ability to adapt and thrive in the face of adversity. Through her journey, she emerges as a powerful and resilient figure, no longer defined by her past victimhood but by her newfound identity as a skilled huntress.
\end{tcolorbox}
\end{minipage}
\begin{minipage}{0.9\linewidth}
\begin{tcolorbox}[enhanced, colback=white, colframe=black, title=Model Analysis, sharp corners]
\begin{Verbatim}[breaklines=true, breakanywhere=true, breaksymbol=, breakanywheresymbolpre=]
{
'system is better': 'A is better',
'evidence': [
'Both answers accurately address the question and provide relevant information about Marian's transformation into a skilled huntress. They both highlight her resilience and self-reliance in overcoming adversity.',
'However, Answer A provides a slightly more detailed and nuanced explanation. It mentions specific details from the story, such as Marian being ""forced into a sham marriage"" and ""escaping into the wilderness,"" which helps to better illustrate her journey of resilience. It also emphasizes the theme of ""reclaiming her agency"" and ""finding empowerment,"" which adds depth to the analysis.',
'Answer B, while accurate, is slightly more generic in its language and doesn’t provide as much specific evidence from the text to support its claims.'
]
}
\end{Verbatim}
\end{tcolorbox}
\end{minipage}

\subsection{The Wild Huntress Example 5: \modelname{} Full-Context vs \modelname{} 4k Context}
\begin{minipage}{0.9\linewidth}
\begin{tcolorbox}[enhanced, colback=white, colframe=black, title=Question, sharp corners]
How does the protagonist's encounter with the huntress in the mountain ravine reveal a surprising connection to their own pursuit and introduce a new layer of complexity to the narrative?
\end{tcolorbox}
\end{minipage}
\begin{minipage}{0.9\linewidth}
\begin{tcolorbox}[enhanced, colback=white, colframe=black, title=System A: Gemini 1.5 Pro Full-Context, sharp corners]
The protagonist's encounter with the huntress, Marian, in the mountain ravine reveals a surprising connection to his pursuit of Lilian as she is revealed to be Lilian's sister. Marian's story of being deceived and nearly abducted by a Mormon named Stebbins confirms the protagonist's suspicions about Lilian's situation. This connection not only deepens the protagonist's determination to rescue Lilian but also adds a layer of personal urgency and emotional complexity to the narrative. Marian's presence as a skilled huntress and knowledgeable ally further complicates the rescue mission, as her involvement carries significant risks due to her past relationship with Stebbins.
\end{tcolorbox}
\end{minipage}
\begin{minipage}{0.9\linewidth}
\begin{tcolorbox}[enhanced, colback=white, colframe=black, title=System B: Gemini 1.5 Pro 4k Context, sharp corners]
The protagonist's encounter with the huntress in the ravine reveals a surprising connection when she mentions a "sister" who is being held captive by the same group pursuing the protagonist. This revelation suggests that the huntress's pursuit intersects with the protagonist's, as they both seek to rescue their loved ones. The huntress's personal stake in the conflict adds a new layer of complexity, as her knowledge of the captors and the terrain becomes crucial to the protagonist's mission. This encounter transforms the narrative from a solitary pursuit into a shared endeavor, uniting the protagonist and the huntress in a common goal.
\end{tcolorbox}
\end{minipage}
\begin{minipage}{0.9\linewidth}
\begin{tcolorbox}[enhanced, colback=white, colframe=black, title=Model Analysis, sharp corners]
\begin{Verbatim}[breaklines=true, breakanywhere=true, breaksymbol=, breakanywheresymbolpre=]
{
'system is better': 'A is better',
'evidence': [
'Answer A is more accurate and detailed in identifying the key elements of the encounter and their significance to the narrative. It correctly names the huntress as Marian, identifies her as Lilian's sister, and highlights the crucial information about Stebbins's deception and Marian's intended fate. This information is explicitly stated in the text, particularly in chapters 70-71 and 93-94.',
'For example, chapter 71 reveals: ""It was a hellish purpose; but you shall hear it.  These Mormons have at their head a great chief priest--una propheta, as they call him.  He is a polygamist--a perfect Turco--and keeps a harem of beautiful ninas, who pass under the name of `spiritual wives.'  It was only after the young Americana had got far out upon the plains--indeed, to the Big Timbers, where she escaped from him--that she found out the terrible fate for which her false husband had designed her.""',
'Answer B, while touching upon the connection and complexity, lacks the specific details that make Answer A a more accurate and insightful evaluation. It fails to name the huntress, mention Stebbins, or elaborate on the nature of the shared pursuit. This lack of detail makes Answer B less informative and less grounded in the context.'
]
}
\end{Verbatim}
\end{tcolorbox}
\end{minipage}

\subsection{The Wild Huntress Example 6: \modelname{} Full-Context vs \modelname{} 4k Context}
\begin{minipage}{0.9\linewidth}
\begin{tcolorbox}[enhanced, colback=white, colframe=black, title=Question, sharp corners]
How does the elder sister's transformation from a victim of betrayal to a skilled huntress symbolize resilience and the power of self-reliance in overcoming adversity?
\end{tcolorbox}
\end{minipage}
\begin{minipage}{0.9\linewidth}
\begin{tcolorbox}[enhanced, colback=white, colframe=black, title=System A: Gemini 1.5 Pro Full-Context, sharp corners]
Marian Holt's transformation into a skilled huntress symbolizes her resilience and strength in the face of betrayal and adversity. After being deceived by her lover and forced into a sham marriage, she escapes into the wilderness and embraces a life of self-reliance. Through her mastery of hunting and survival skills, she reclaims her agency and finds empowerment in the face of hardship. This transformation highlights the potential for personal growth and the ability to overcome challenges through self-determination. Marian's journey embodies the idea that even in the face of betrayal and loss, one can find strength and forge a new path.
\end{tcolorbox}
\end{minipage}
\begin{minipage}{0.9\linewidth}
\begin{tcolorbox}[enhanced, colback=white, colframe=black, title=System B: Gemini 1.5 Pro 4k Context, sharp corners]
Marian Holt's journey from a vulnerable young woman betrayed by a manipulative figure to a skilled and independent huntress embodies the transformative power of resilience. Faced with adversity, she did not succumb to despair but instead embraced a life of self-reliance in the wilderness. Her mastery of survival skills and hunting prowess symbolizes her inner strength and determination to overcome past trauma. Marian's transformation serves as an inspiring example of how individuals can rise above challenging circumstances and forge a new path defined by strength and independence.
\end{tcolorbox}
\end{minipage}
\begin{minipage}{0.9\linewidth}
\begin{tcolorbox}[enhanced, colback=white, colframe=black, title=Model Analysis, sharp corners]
\begin{Verbatim}[breaklines=true, breakanywhere=true, breaksymbol=, breakanywheresymbolpre=]
{
'system is better': 'None',
'evidence': [
'Both answers provide unsupported claims about Marian's motivations for becoming a skilled huntress. While the text describes her as a skilled huntress and notes her resilience, it does not explicitly link her hunting skills to overcoming her betrayal. Answer A states, ""After being deceived by her lover and forced into a sham marriage, she escapes into the wilderness and embraces a life of self-reliance. Through her mastery of hunting and survival skills, she reclaims her agency and finds empowerment in the face of hardship.""  However, the text does not directly state that she developed these skills after the betrayal. It is mentioned that she was ""Indianised"" by her upbringing, suggesting a familiarity with hunting from an earlier age. Answer B similarly claims, ""Faced with adversity, she did not succumb to despair but instead embraced a life of self-reliance in the wilderness. Her mastery of survival skills and hunting prowess symbolizes her inner strength and determination to overcome past trauma."" Again, the text does not explicitly connect her hunting skills to overcoming the trauma of betrayal.'
]
}
\end{Verbatim}
\end{tcolorbox}
\end{minipage}

%% file: main.bbl
\begin{thebibliography}{40}
\providecommand{\natexlab}[1]{#1}
\providecommand{\url}[1]{\texttt{#1}}
\expandafter\ifx\csname urlstyle\endcsname\relax
  \providecommand{\doi}[1]{doi: #1}\else
  \providecommand{\doi}{doi: \begingroup \urlstyle{rm}\Url}\fi

\bibitem[Kočiský et~al.(2017)Kočiský, Schwarz, Blunsom, Dyer, Hermann,
  Melis, and Grefenstette]{narrativeqa}
Tomáš Kočiský, Jonathan Schwarz, Phil Blunsom, Chris Dyer, Karl~Moritz
  Hermann, Gábor Melis, and Edward Grefenstette.
\newblock The narrativeqa reading comprehension challenge, 2017.

\bibitem[Team et~al.(2024{\natexlab{a}})Team, Reid, Savinov, Teplyashin,
  Dmitry, Lepikhin, Lillicrap, baptiste Alayrac, Soricut, Lazaridou, Firat,
  Schrittwieser, Antonoglou, Anil, Borgeaud, Dai, Millican, Dyer, Glaese,
  Sottiaux, Lee, Viola, Reynolds, Xu, Molloy, Chen, Isard, Barham, Hennigan,
  McIlroy, Johnson, Schalkwyk, Collins, Rutherford, Moreira, Ayoub, Goel,
  Meyer, Thornton, Yang, Michalewski, Abbas, Schucher, Anand, Ives, Keeling,
  Lenc, Haykal, Shakeri, Shyam, Chowdhery, Ring, Spencer, Sezener, Vilnis,
  Chang, Morioka, Tucker, Zheng, Woodman, Attaluri, Kocisky, Eltyshev, Chen,
  Chung, Selo, Brahma, Georgiev, Slone, Zhu, Lottes, Qiao, Caine, Riedel,
  Tomala, Chadwick, Love, Choy, Mittal, Houlsby, Tang, Lamm, Bai, Zhang, He,
  Cheng, Humphreys, Li, Brin, Cassirer, Miao, Zilka, Tobin, Xu, Proleev, Sohn,
  Magni, Hendricks, Gao, Ontanon, Bunyan, Byrd, Sharma, Zhang, Pinto, Sinha,
  Mehta, Jia, Caelles, Webson, Morris, Roelofs, Ding, Strudel, Xiong, Ritter,
  Dehghani, Chaabouni, Karmarkar, Lai, Mentzer, Xu, Li, Zhang, Paine, Goldin,
  Neyshabur, Baumli, Levskaya, Laskin, Jia, Rae, Xiao, He, Giordano, Yagati,
  Lespiau, Natsev, Ganapathy, Liu, Martins, Chen, Xu, Barnes, May, Vezer, Oh,
  Franko, Bridgers, Zhao, Wu, Mustafa, Sechrist, Parisotto, Pillai, Larkin, Gu,
  Sorokin, Krikun, Guseynov, Landon, Datta, Pritzel, Thacker, Yang, Hui, Hauth,
  Yeh, Barker, Mao-Jones, Austin, Sheahan, Schuh, Svensson, Jain, Ramasesh,
  Briukhov, Chung, von Glehn, Butterfield, Jhakra, Wiethoff, Frye, Grimstad,
  Changpinyo, Lan, Bortsova, Wu, Voigtlaender, Sainath, Gu, Smith, Hawkins,
  Cao, Besley, Srinivasan, Omernick, Gaffney, Surita, Burnell, Damoc, Ahn,
  Brock, Pajarskas, Petrushkina, Noury, Blanco, Swersky, Ahuja, Avrahami,
  Misra, de~Liedekerke, Iinuma, Polozov, York, van~den Driessche, Michel, Chiu,
  Blevins, Gleicher, Recasens, Rrustemi, Gribovskaya, Roy, Gworek, Arnold, Lee,
  Lee-Thorp, Maggioni, Piqueras, Badola, Vikram, Gonzalez, Baddepudi, Senter,
  Devlin, Qin, Azzam, Trebacz, Polacek, Krishnakumar, yiin Chang, Tung,
  Penchev, Joshi, Olszewska, Muir, Wirth, Hartman, Newlan, Kashem, Bolina,
  Dabir, van Amersfoort, Ahmed, Cobon-Kerr, Kamath, Hrafnkelsson, Hou,
  Mackinnon, Frechette, Noland, Si, Taropa, Li, Crone, Gulati, Cevey, Adler,
  Ma, Silver, Tokumine, Powell, Lee, Vodrahalli, Hassan, Mincu, Yang, Levine,
  Brennan, Wang, Hodkinson, Zhao, Lipschultz, Pope, Chang, Li, Shafey,
  Paganini, Douglas, Bohnet, Pardo, Odoom, Rosca, dos Santos, Soparkar, Guez,
  Hudson, Hansen, Asawaroengchai, Addanki, Yu, Stokowiec, Khan, Gilmer, Lee,
  Bostock, Rong, Caton, Pejman, Pavetic, Brown, Sharma, Lučić, Samuel,
  Djolonga, Mandhane, Sjösund, Buchatskaya, White, Clay, Jiang, Lim, Hemsley,
  Cankara, Labanowski, Cao, Steiner, Hashemi, Austin, Gergely, Blyth, Stanton,
  Shivakumar, Siddhant, Andreassen, Araya, Sethi, Shivanna, Hand, Bapna,
  Khodaei, Miech, Tanzer, Swing, Thakoor, Aroyo, Pan, Nado, Sygnowski, Winkler,
  Yu, Saleh, Maggiore, Bansal, Garcia, Kazemi, Patil, Dasgupta, Barr, Giang,
  Kagohara, Danihelka, Marathe, Feinberg, Elhawaty, Ghelani, Horgan, Miller,
  Walker, Tanburn, Tariq, Shrivastava, Xia, Wang, Chiu, Ashwood, Baatarsukh,
  Samangooei, Kaufman, Alcober, Stjerngren, Komarek, Tsihlas, Boral, Comanescu,
  Chen, Liu, Welty, Bloxwich, Chen, Sun, Feng, Mauger, Dotiwalla, Hellendoorn,
  Sharman, Zheng, Haridasan, Barth-Maron, Swanson, Rogozińska, Andreev,
  Rubenstein, Sang, Hurt, Elsayed, Wang, Lacey, Ilić, Zhao, Iwanicki, Lince,
  Chen, Lyu, Lebsack, Griffith, Gaba, Sandhu, Chen, Koop, Rajwar, Yeganeh,
  Chang, Zhu, Radpour, Davoodi, Lei, Xu, Toyama, Segal, Wicke, Lin, Bulanova,
  Badia, Rakićević, Sprechmann, Filos, Hou, Campos, Kassner, Sachan,
  Fortunato, Iwuanyanwu, Nikolaev, Lakshminarayanan, Jazayeri, Varadarajan,
  Tekur, Fritz, Khalman, Reitter, Dasgupta, Sarcar, Ornduff, Snaider, Huot,
  Jia, Kemp, Trdin, Vijayakumar, Kim, Angermueller, Lao, Liu, Zhang, Engel,
  Greene, White, Austin, Taylor, Ashraf, Liu, Georgaki, Cai, Kulizhskaya,
  Goenka, Saeta, Xu, Frank, de~Cesare, Robenek, Richardson, Alnahlawi, Yew,
  Ponnapalli, Tagliasacchi, Korchemniy, Kim, Li, Rosgen, Levin, Wiesner,
  Banzal, Srinivasan, Yu, Çağlar Ünlü, Reid, Tung, Finchelstein, Kumar,
  Elisseeff, Huang, Zhang, Aguilar, Giménez, Xia, Dousse, Gierke, Yates,
  Jalan, Li, Latorre-Chimoto, Nguyen, Durden, Kallakuri, Liu, Johnson, Tsai,
  Talbert, Liu, Neitz, Elkind, Selvi, Jasarevic, Soares, Cui, Wang, Wang, Ye,
  Kallarackal, Loher, Lam, Broder, Holtmann-Rice, Martin, Ramadhana, Shukla,
  Basu, Mohan, Fernando, Fiedel, Paterson, Li, Garg, Park, Choi, Wu, Singh,
  Zhang, Globerson, Yu, Carpenter, de~Chaumont~Quitry, Radebaugh, Lin, Tudor,
  Shroff, Garmon, Du, Vats, Lu, Iqbal, Yakubovich, Tripuraneni, Manyika,
  Qureshi, Hua, Ngani, Raad, Forbes, Stanway, Sundararajan, Ungureanu, Bishop,
  Li, Venkatraman, Li, Thornton, Scellato, Gupta, Wang, Tenney, Wu, Shenoy,
  Carvajal, Wright, Bariach, Xiao, Hawkins, Dalmia, Farabet, Valenzuela, Yuan,
  Agarwal, Chen, Kim, Hulse, Dukkipati, Paszke, Bolt, Choo, Beattie, Prendki,
  Vashisht, Santamaria-Fernandez, Cobo, Wilkiewicz, Madras, Elqursh, Uy,
  Ramirez, Harvey, Liechty, Zen, Seibert, Hu, Khorlin, Le, Aharoni, Li, Wang,
  Kumar, Casagrande, Hoover, Badawy, Soergel, Vnukov, Miecnikowski, Simsa,
  Kumar, Sellam, Vlasic, Daruki, Shabat, Zhang, Su, Zhang, Liu, Sun, Palmer,
  Ghaffarkhah, Xiong, Cotruta, Fink, Dixon, Sreevatsa, Goedeckemeyer,
  Dimitriev, Jafari, Crocker, FitzGerald, Kumar, Ghemawat, Philips, Liu, Liang,
  Sterneck, Repina, Wu, Knight, Georgiev, Lee, Askham, Chakladar, Louis, Crous,
  Cate, Petrova, Quinn, Owusu-Afriyie, Singhal, Wei, Kim, Vincent, Nasr,
  Choquette-Choo, Tojo, Lu, de~Las~Casas, Cheng, Bolukbasi, Lee, Fatehi,
  Ananthanarayanan, Patel, Kaed, Li, Belle, Chen, Konzelmann, Põder, Garg,
  Koverkathu, Brown, Dyer, Liu, Nova, Xu, Walton, Parrish, Epstein, McCarthy,
  Petrov, Hassabis, Kavukcuoglu, Dean, and Vinyals]{gemini1-5}
Gemini Team, Machel Reid, Nikolay Savinov, Denis Teplyashin, Dmitry, Lepikhin,
  Timothy Lillicrap, Jean baptiste Alayrac, Radu Soricut, Angeliki Lazaridou,
  Orhan Firat, Julian Schrittwieser, Ioannis Antonoglou, Rohan Anil, Sebastian
  Borgeaud, Andrew Dai, Katie Millican, Ethan Dyer, Mia Glaese, Thibault
  Sottiaux, Benjamin Lee, Fabio Viola, Malcolm Reynolds, Yuanzhong Xu, James
  Molloy, Jilin Chen, Michael Isard, Paul Barham, Tom Hennigan, Ross McIlroy,
  Melvin Johnson, Johan Schalkwyk, Eli Collins, Eliza Rutherford, Erica
  Moreira, Kareem Ayoub, Megha Goel, Clemens Meyer, Gregory Thornton, Zhen
  Yang, Henryk Michalewski, Zaheer Abbas, Nathan Schucher, Ankesh Anand,
  Richard Ives, James Keeling, Karel Lenc, Salem Haykal, Siamak Shakeri, Pranav
  Shyam, Aakanksha Chowdhery, Roman Ring, Stephen Spencer, Eren Sezener, Luke
  Vilnis, Oscar Chang, Nobuyuki Morioka, George Tucker, Ce~Zheng, Oliver
  Woodman, Nithya Attaluri, Tomas Kocisky, Evgenii Eltyshev, Xi~Chen, Timothy
  Chung, Vittorio Selo, Siddhartha Brahma, Petko Georgiev, Ambrose Slone,
  Zhenkai Zhu, James Lottes, Siyuan Qiao, Ben Caine, Sebastian Riedel, Alex
  Tomala, Martin Chadwick, Juliette Love, Peter Choy, Sid Mittal, Neil Houlsby,
  Yunhao Tang, Matthew Lamm, Libin Bai, Qiao Zhang, Luheng He, Yong Cheng,
  Peter Humphreys, Yujia Li, Sergey Brin, Albin Cassirer, Yingjie Miao, Lukas
  Zilka, Taylor Tobin, Kelvin Xu, Lev Proleev, Daniel Sohn, Alberto Magni,
  Lisa~Anne Hendricks, Isabel Gao, Santiago Ontanon, Oskar Bunyan, Nathan Byrd,
  Abhanshu Sharma, Biao Zhang, Mario Pinto, Rishika Sinha, Harsh Mehta, Dawei
  Jia, Sergi Caelles, Albert Webson, Alex Morris, Becca Roelofs, Yifan Ding,
  Robin Strudel, Xuehan Xiong, Marvin Ritter, Mostafa Dehghani, Rahma
  Chaabouni, Abhijit Karmarkar, Guangda Lai, Fabian Mentzer, Bibo Xu, YaGuang
  Li, Yujing Zhang, Tom~Le Paine, Alex Goldin, Behnam Neyshabur, Kate Baumli,
  Anselm Levskaya, Michael Laskin, Wenhao Jia, Jack~W. Rae, Kefan Xiao, Antoine
  He, Skye Giordano, Lakshman Yagati, Jean-Baptiste Lespiau, Paul Natsev,
  Sanjay Ganapathy, Fangyu Liu, Danilo Martins, Nanxin Chen, Yunhan Xu, Megan
  Barnes, Rhys May, Arpi Vezer, Junhyuk Oh, Ken Franko, Sophie Bridgers, Ruizhe
  Zhao, Boxi Wu, Basil Mustafa, Sean Sechrist, Emilio Parisotto,
  Thanumalayan~Sankaranarayana Pillai, Chris Larkin, Chenjie Gu, Christina
  Sorokin, Maxim Krikun, Alexey Guseynov, Jessica Landon, Romina Datta,
  Alexander Pritzel, Phoebe Thacker, Fan Yang, Kevin Hui, Anja Hauth, Chih-Kuan
  Yeh, David Barker, Justin Mao-Jones, Sophia Austin, Hannah Sheahan, Parker
  Schuh, James Svensson, Rohan Jain, Vinay Ramasesh, Anton Briukhov, Da-Woon
  Chung, Tamara von Glehn, Christina Butterfield, Priya Jhakra, Matthew
  Wiethoff, Justin Frye, Jordan Grimstad, Beer Changpinyo, Charline~Le Lan,
  Anna Bortsova, Yonghui Wu, Paul Voigtlaender, Tara Sainath, Shane Gu,
  Charlotte Smith, Will Hawkins, Kris Cao, James Besley, Srivatsan Srinivasan,
  Mark Omernick, Colin Gaffney, Gabriela Surita, Ryan Burnell, Bogdan Damoc,
  Junwhan Ahn, Andrew Brock, Mantas Pajarskas, Anastasia Petrushkina, Seb
  Noury, Lorenzo Blanco, Kevin Swersky, Arun Ahuja, Thi Avrahami, Vedant Misra,
  Raoul de~Liedekerke, Mariko Iinuma, Alex Polozov, Sarah York, George van~den
  Driessche, Paul Michel, Justin Chiu, Rory Blevins, Zach Gleicher, Adrià
  Recasens, Alban Rrustemi, Elena Gribovskaya, Aurko Roy, Wiktor Gworek,
  Sébastien M.~R. Arnold, Lisa Lee, James Lee-Thorp, Marcello Maggioni,
  Enrique Piqueras, Kartikeya Badola, Sharad Vikram, Lucas Gonzalez, Anirudh
  Baddepudi, Evan Senter, Jacob Devlin, James Qin, Michael Azzam, Maja Trebacz,
  Martin Polacek, Kashyap Krishnakumar, Shuo yiin Chang, Matthew Tung, Ivo
  Penchev, Rishabh Joshi, Kate Olszewska, Carrie Muir, Mateo Wirth, Ale~Jakse
  Hartman, Josh Newlan, Sheleem Kashem, Vijay Bolina, Elahe Dabir, Joost van
  Amersfoort, Zafarali Ahmed, James Cobon-Kerr, Aishwarya Kamath, Arnar~Mar
  Hrafnkelsson, Le~Hou, Ian Mackinnon, Alexandre Frechette, Eric Noland, Xiance
  Si, Emanuel Taropa, Dong Li, Phil Crone, Anmol Gulati, Sébastien Cevey,
  Jonas Adler, Ada Ma, David Silver, Simon Tokumine, Richard Powell, Stephan
  Lee, Kiran Vodrahalli, Samer Hassan, Diana Mincu, Antoine Yang, Nir Levine,
  Jenny Brennan, Mingqiu Wang, Sarah Hodkinson, Jeffrey Zhao, Josh Lipschultz,
  Aedan Pope, Michael~B. Chang, Cheng Li, Laurent~El Shafey, Michela Paganini,
  Sholto Douglas, Bernd Bohnet, Fabio Pardo, Seth Odoom, Mihaela Rosca,
  Cicero~Nogueira dos Santos, Kedar Soparkar, Arthur Guez, Tom Hudson, Steven
  Hansen, Chulayuth Asawaroengchai, Ravi Addanki, Tianhe Yu, Wojciech
  Stokowiec, Mina Khan, Justin Gilmer, Jaehoon Lee, Carrie~Grimes Bostock,
  Keran Rong, Jonathan Caton, Pedram Pejman, Filip Pavetic, Geoff Brown, Vivek
  Sharma, Mario Lučić, Rajkumar Samuel, Josip Djolonga, Amol Mandhane,
  Lars~Lowe Sjösund, Elena Buchatskaya, Elspeth White, Natalie Clay, Jiepu
  Jiang, Hyeontaek Lim, Ross Hemsley, Zeyncep Cankara, Jane Labanowski,
  Nicola~De Cao, David Steiner, Sayed~Hadi Hashemi, Jacob Austin, Anita
  Gergely, Tim Blyth, Joe Stanton, Kaushik Shivakumar, Aditya Siddhant, Anders
  Andreassen, Carlos Araya, Nikhil Sethi, Rakesh Shivanna, Steven Hand, Ankur
  Bapna, Ali Khodaei, Antoine Miech, Garrett Tanzer, Andy Swing, Shantanu
  Thakoor, Lora Aroyo, Zhufeng Pan, Zachary Nado, Jakub Sygnowski, Stephanie
  Winkler, Dian Yu, Mohammad Saleh, Loren Maggiore, Yamini Bansal, Xavier
  Garcia, Mehran Kazemi, Piyush Patil, Ishita Dasgupta, Iain Barr, Minh Giang,
  Thais Kagohara, Ivo Danihelka, Amit Marathe, Vladimir Feinberg, Mohamed
  Elhawaty, Nimesh Ghelani, Dan Horgan, Helen Miller, Lexi Walker, Richard
  Tanburn, Mukarram Tariq, Disha Shrivastava, Fei Xia, Qingze Wang, Chung-Cheng
  Chiu, Zoe Ashwood, Khuslen Baatarsukh, Sina Samangooei, Raphaël~Lopez
  Kaufman, Fred Alcober, Axel Stjerngren, Paul Komarek, Katerina Tsihlas,
  Anudhyan Boral, Ramona Comanescu, Jeremy Chen, Ruibo Liu, Chris Welty, Dawn
  Bloxwich, Charlie Chen, Yanhua Sun, Fangxiaoyu Feng, Matthew Mauger, Xerxes
  Dotiwalla, Vincent Hellendoorn, Michael Sharman, Ivy Zheng, Krishna
  Haridasan, Gabe Barth-Maron, Craig Swanson, Dominika Rogozińska, Alek
  Andreev, Paul~Kishan Rubenstein, Ruoxin Sang, Dan Hurt, Gamaleldin Elsayed,
  Renshen Wang, Dave Lacey, Anastasija Ilić, Yao Zhao, Adam Iwanicki,
  Alejandro Lince, Alexander Chen, Christina Lyu, Carl Lebsack, Jordan
  Griffith, Meenu Gaba, Paramjit Sandhu, Phil Chen, Anna Koop, Ravi Rajwar,
  Soheil~Hassas Yeganeh, Solomon Chang, Rui Zhu, Soroush Radpour, Elnaz
  Davoodi, Ving~Ian Lei, Yang Xu, Daniel Toyama, Constant Segal, Martin Wicke,
  Hanzhao Lin, Anna Bulanova, Adrià~Puigdomènech Badia, Nemanja Rakićević,
  Pablo Sprechmann, Angelos Filos, Shaobo Hou, Víctor Campos, Nora Kassner,
  Devendra Sachan, Meire Fortunato, Chimezie Iwuanyanwu, Vitaly Nikolaev,
  Balaji Lakshminarayanan, Sadegh Jazayeri, Mani Varadarajan, Chetan Tekur,
  Doug Fritz, Misha Khalman, David Reitter, Kingshuk Dasgupta, Shourya Sarcar,
  Tina Ornduff, Javier Snaider, Fantine Huot, Johnson Jia, Rupert Kemp, Nejc
  Trdin, Anitha Vijayakumar, Lucy Kim, Christof Angermueller, Li~Lao, Tianqi
  Liu, Haibin Zhang, David Engel, Somer Greene, Anaïs White, Jessica Austin,
  Lilly Taylor, Shereen Ashraf, Dangyi Liu, Maria Georgaki, Irene Cai, Yana
  Kulizhskaya, Sonam Goenka, Brennan Saeta, Ying Xu, Christian Frank, Dario
  de~Cesare, Brona Robenek, Harry Richardson, Mahmoud Alnahlawi, Christopher
  Yew, Priya Ponnapalli, Marco Tagliasacchi, Alex Korchemniy, Yelin Kim,
  Dinghua Li, Bill Rosgen, Kyle Levin, Jeremy Wiesner, Praseem Banzal, Praveen
  Srinivasan, Hongkun Yu, Çağlar Ünlü, David Reid, Zora Tung, Daniel
  Finchelstein, Ravin Kumar, Andre Elisseeff, Jin Huang, Ming Zhang, Ricardo
  Aguilar, Mai Giménez, Jiawei Xia, Olivier Dousse, Willi Gierke, Damion
  Yates, Komal Jalan, Lu~Li, Eri Latorre-Chimoto, Duc~Dung Nguyen, Ken Durden,
  Praveen Kallakuri, Yaxin Liu, Matthew Johnson, Tomy Tsai, Alice Talbert,
  Jasmine Liu, Alexander Neitz, Chen Elkind, Marco Selvi, Mimi Jasarevic,
  Livio~Baldini Soares, Albert Cui, Pidong Wang, Alek~Wenjiao Wang, Xinyu Ye,
  Krystal Kallarackal, Lucia Loher, Hoi Lam, Josef Broder, Dan Holtmann-Rice,
  Nina Martin, Bramandia Ramadhana, Mrinal Shukla, Sujoy Basu, Abhi Mohan, Nick
  Fernando, Noah Fiedel, Kim Paterson, Hui Li, Ankush Garg, Jane Park, DongHyun
  Choi, Diane Wu, Sankalp Singh, Zhishuai Zhang, Amir Globerson, Lily Yu, John
  Carpenter, Félix de~Chaumont~Quitry, Carey Radebaugh, Chu-Cheng Lin, Alex
  Tudor, Prakash Shroff, Drew Garmon, Dayou Du, Neera Vats, Han Lu, Shariq
  Iqbal, Alex Yakubovich, Nilesh Tripuraneni, James Manyika, Haroon Qureshi,
  Nan Hua, Christel Ngani, Maria~Abi Raad, Hannah Forbes, Jeff Stanway, Mukund
  Sundararajan, Victor Ungureanu, Colton Bishop, Yunjie Li, Balaji Venkatraman,
  Bo~Li, Chloe Thornton, Salvatore Scellato, Nishesh Gupta, Yicheng Wang, Ian
  Tenney, Xihui Wu, Ashish Shenoy, Gabriel Carvajal, Diana~Gage Wright, Ben
  Bariach, Zhuyun Xiao, Peter Hawkins, Sid Dalmia, Clement Farabet, Pedro
  Valenzuela, Quan Yuan, Ananth Agarwal, Mia Chen, Wooyeol Kim, Brice Hulse,
  Nandita Dukkipati, Adam Paszke, Andrew Bolt, Kiam Choo, Jennifer Beattie,
  Jennifer Prendki, Harsha Vashisht, Rebeca Santamaria-Fernandez, Luis~C. Cobo,
  Jarek Wilkiewicz, David Madras, Ali Elqursh, Grant Uy, Kevin Ramirez, Matt
  Harvey, Tyler Liechty, Heiga Zen, Jeff Seibert, Clara~Huiyi Hu, Andrey
  Khorlin, Maigo Le, Asaf Aharoni, Megan Li, Lily Wang, Sandeep Kumar, Norman
  Casagrande, Jay Hoover, Dalia~El Badawy, David Soergel, Denis Vnukov, Matt
  Miecnikowski, Jiri Simsa, Praveen Kumar, Thibault Sellam, Daniel Vlasic,
  Samira Daruki, Nir Shabat, John Zhang, Guolong Su, Jiageng Zhang, Jeremiah
  Liu, Yi~Sun, Evan Palmer, Alireza Ghaffarkhah, Xi~Xiong, Victor Cotruta,
  Michael Fink, Lucas Dixon, Ashwin Sreevatsa, Adrian Goedeckemeyer, Alek
  Dimitriev, Mohsen Jafari, Remi Crocker, Nicholas FitzGerald, Aviral Kumar,
  Sanjay Ghemawat, Ivan Philips, Frederick Liu, Yannie Liang, Rachel Sterneck,
  Alena Repina, Marcus Wu, Laura Knight, Marin Georgiev, Hyo Lee, Harry Askham,
  Abhishek Chakladar, Annie Louis, Carl Crous, Hardie Cate, Dessie Petrova,
  Michael Quinn, Denese Owusu-Afriyie, Achintya Singhal, Nan Wei, Solomon Kim,
  Damien Vincent, Milad Nasr, Christopher~A. Choquette-Choo, Reiko Tojo, Shawn
  Lu, Diego de~Las~Casas, Yuchung Cheng, Tolga Bolukbasi, Katherine Lee, Saaber
  Fatehi, Rajagopal Ananthanarayanan, Miteyan Patel, Charbel Kaed, Jing Li,
  Shreyas~Rammohan Belle, Zhe Chen, Jaclyn Konzelmann, Siim Põder, Roopal
  Garg, Vinod Koverkathu, Adam Brown, Chris Dyer, Rosanne Liu, Azade Nova, Jun
  Xu, Alanna Walton, Alicia Parrish, Mark Epstein, Sara McCarthy, Slav Petrov,
  Demis Hassabis, Koray Kavukcuoglu, Jeffrey Dean, and Oriol Vinyals.
\newblock Gemini 1.5: Unlocking multimodal understanding across millions of
  tokens of context, 2024{\natexlab{a}}.

\bibitem[Team et~al.(2024{\natexlab{b}})Team, Anil, Borgeaud, Alayrac, Yu,
  Soricut, Schalkwyk, Dai, Hauth, Millican, Silver, Johnson, Antonoglou,
  Schrittwieser, Glaese, Chen, Pitler, Lillicrap, Lazaridou, Firat, Molloy,
  Isard, Barham, Hennigan, Lee, Viola, Reynolds, Xu, Doherty, Collins, Meyer,
  Rutherford, Moreira, Ayoub, Goel, Krawczyk, Du, Chi, Cheng, Ni, Shah, Kane,
  Chan, Faruqui, Severyn, Lin, Li, Cheng, Ittycheriah, Mahdieh, Chen, Sun,
  Tran, Bagri, Lakshminarayanan, Liu, Orban, Güra, Zhou, Song, Boffy,
  Ganapathy, Zheng, Choe, Ágoston Weisz, Zhu, Lu, Gopal, Kahn, Kula, Pitman,
  Shah, Taropa, Merey, Baeuml, Chen, Shafey, Zhang, Sercinoglu, Tucker,
  Piqueras, Krikun, Barr, Savinov, Danihelka, Roelofs, White, Andreassen, von
  Glehn, Yagati, Kazemi, Gonzalez, Khalman, Sygnowski, Frechette, Smith, Culp,
  Proleev, Luan, Chen, Lottes, Schucher, Lebron, Rrustemi, Clay, Crone,
  Kocisky, Zhao, Perz, Yu, Howard, Bloniarz, Rae, Lu, Sifre, Maggioni, Alcober,
  Garrette, Barnes, Thakoor, Austin, Barth-Maron, Wong, Joshi, Chaabouni,
  Fatiha, Ahuja, Tomar, Senter, Chadwick, Kornakov, Attaluri, Iturrate, Liu,
  Li, Cogan, Chen, Jia, Gu, Zhang, Grimstad, Hartman, Garcia, Pillai, Devlin,
  Laskin, de~Las~Casas, Valter, Tao, Blanco, Badia, Reitter, Chen, Brennan,
  Rivera, Brin, Iqbal, Surita, Labanowski, Rao, Winkler, Parisotto, Gu,
  Olszewska, Addanki, Miech, Louis, Teplyashin, Brown, Catt, Balaguer, Xiang,
  Wang, Ashwood, Briukhov, Webson, Ganapathy, Sanghavi, Kannan, Chang,
  Stjerngren, Djolonga, Sun, Bapna, Aitchison, Pejman, Michalewski, Yu, Wang,
  Love, Ahn, Bloxwich, Han, Humphreys, Sellam, Bradbury, Godbole, Samangooei,
  Damoc, Kaskasoli, Arnold, Vasudevan, Agrawal, Riesa, Lepikhin, Tanburn,
  Srinivasan, Lim, Hodkinson, Shyam, Ferret, Hand, Garg, Paine, Li, Li, Giang,
  Neitz, Abbas, York, Reid, Cole, Chowdhery, Das, Rogozińska, Nikolaev,
  Sprechmann, Nado, Zilka, Prost, He, Monteiro, Mishra, Welty, Newlan, Jia,
  Allamanis, Hu, de~Liedekerke, Gilmer, Saroufim, Rijhwani, Hou, Shrivastava,
  Baddepudi, Goldin, Ozturel, Cassirer, Xu, Sohn, Sachan, Amplayo, Swanson,
  Petrova, Narayan, Guez, Brahma, Landon, Patel, Zhao, Villela, Wang, Jia,
  Rahtz, Giménez, Yeung, Keeling, Georgiev, Mincu, Wu, Haykal, Saputro,
  Vodrahalli, Qin, Cankara, Sharma, Fernando, Hawkins, Neyshabur, Kim, Hutter,
  Agrawal, Castro-Ros, van~den Driessche, Wang, Yang, yiin Chang, Komarek,
  McIlroy, Lučić, Zhang, Farhan, Sharman, Natsev, Michel, Bansal, Qiao, Cao,
  Shakeri, Butterfield, Chung, Rubenstein, Agrawal, Mensch, Soparkar, Lenc,
  Chung, Pope, Maggiore, Kay, Jhakra, Wang, Maynez, Phuong, Tobin, Tacchetti,
  Trebacz, Robinson, Katariya, Riedel, Bailey, Xiao, Ghelani, Aroyo, Slone,
  Houlsby, Xiong, Yang, Gribovskaya, Adler, Wirth, Lee, Li, Kagohara,
  Pavagadhi, Bridgers, Bortsova, Ghemawat, Ahmed, Liu, Powell, Bolina, Iinuma,
  Zablotskaia, Besley, Chung, Dozat, Comanescu, Si, Greer, Su, Polacek,
  Kaufman, Tokumine, Hu, Buchatskaya, Miao, Elhawaty, Siddhant, Tomasev, Xing,
  Greer, Miller, Ashraf, Roy, Zhang, Ma, Filos, Besta, Blevins, Klimenko, Yeh,
  Changpinyo, Mu, Chang, Pajarskas, Muir, Cohen, Lan, Haridasan, Marathe,
  Hansen, Douglas, Samuel, Wang, Austin, Lan, Jiang, Chiu, Lorenzo, Sjösund,
  Cevey, Gleicher, Avrahami, Boral, Srinivasan, Selo, May, Aisopos, Hussenot,
  Soares, Baumli, Chang, Recasens, Caine, Pritzel, Pavetic, Pardo, Gergely,
  Frye, Ramasesh, Horgan, Badola, Kassner, Roy, Dyer, Campos, Tomala, Tang,
  Badawy, White, Mustafa, Lang, Jindal, Vikram, Gong, Caelles, Hemsley,
  Thornton, Feng, Stokowiec, Zheng, Thacker, Çağlar Ünlü, Zhang, Saleh,
  Svensson, Bileschi, Patil, Anand, Ring, Tsihlas, Vezer, Selvi, Shevlane,
  Rodriguez, Kwiatkowski, Daruki, Rong, Dafoe, FitzGerald, Gu-Lemberg, Khan,
  Hendricks, Pellat, Feinberg, Cobon-Kerr, Sainath, Rauh, Hashemi, Ives,
  Hasson, Noland, Cao, Byrd, Hou, Wang, Sottiaux, Paganini, Lespiau, Moufarek,
  Hassan, Shivakumar, van Amersfoort, Mandhane, Joshi, Goyal, Tung, Brock,
  Sheahan, Misra, Li, Rakićević, Dehghani, Liu, Mittal, Oh, Noury, Sezener,
  Huot, Lamm, Cao, Chen, Mudgal, Stella, Brooks, Vasudevan, Liu, Chain,
  Melinkeri, Cohen, Wang, Seymore, Zubkov, Goel, Yue, Krishnakumaran, Albert,
  Hurley, Sano, Mohananey, Joughin, Filonov, Kępa, Eldawy, Lim, Rishi,
  Badiezadegan, Bos, Chang, Jain, Padmanabhan, Puttagunta, Krishna, Baker,
  Kalb, Bedapudi, Kurzrok, Lei, Yu, Litvin, Zhou, Wu, Sobell, Siciliano, Papir,
  Neale, Bragagnolo, Toor, Chen, Anklin, Wang, Feng, Gholami, Ling, Liu,
  Walter, Moghaddam, Kishore, Adamek, Mercado, Mallinson, Wandekar, Cagle,
  Ofek, Garrido, Lombriser, Mukha, Sun, Mohammad, Matak, Qian, Peswani, Janus,
  Yuan, Schelin, David, Garg, He, Duzhyi, Älgmyr, Lottaz, Li, Yadav, Xu,
  Chinien, Shivanna, Chuklin, Li, Spadine, Wolfe, Mohamed, Das, Dai, He, von
  Dincklage, Upadhyay, Maurya, Chi, Krause, Salama, Rabinovitch, M, Selvan,
  Dektiarev, Ghiasi, Guven, Gupta, Liu, Sharma, Shtacher, Paul, Akerlund,
  Aubet, Huang, Zhu, Zhu, Teixeira, Fritze, Bertolini, Marinescu, Bölle,
  Paulus, Gupta, Latkar, Chang, Sanders, Wilson, Wu, Tan, Thiet, Doshi, Lall,
  Mishra, Chen, Luong, Benjamin, Lee, Andrejczuk, Rabiej, Ranjan, Styrc, Yin,
  Simon, Harriott, Bansal, Robsky, Bacon, Greene, Mirylenka, Zhou, Sarvana,
  Goyal, Andermatt, Siegler, Horn, Israel, Pongetti, Chen, Selvatici, Silva,
  Wang, Tolins, Guu, Yogev, Cai, Agostini, Shah, Nguyen, Donnaile, Pereira,
  Friso, Stambler, Kurzrok, Kuang, Romanikhin, Geller, Yan, Jang, Lee, Fica,
  Malmi, Tan, Banica, Balle, Pham, Huang, Avram, Shi, Singh, Hidey, Ahuja,
  Saxena, Dooley, Potharaju, O'Neill, Gokulchandran, Foley, Zhao, Dusenberry,
  Liu, Mehta, Kotikalapudi, Safranek-Shrader, Goodman, Kessinger, Globen,
  Kolhar, Gorgolewski, Ibrahim, Song, Eichenbaum, Brovelli, Potluri, Lahoti,
  Baetu, Ghorbani, Chen, Crawford, Pal, Sridhar, Gurita, Mujika, Petrovski,
  Cedoz, Li, Chen, Santo, Goyal, Punjabi, Kappaganthu, Kwak, LV, Velury,
  Choudhury, Hall, Shah, Figueira, Thomas, Lu, Zhou, Kumar, Jurdi, Chikkerur,
  Ma, Yu, Kwak, Ähdel, Rajayogam, Choma, Liu, Barua, Ji, Park, Hellendoorn,
  Bailey, Bilal, Zhou, Khatir, Sutton, Rzadkowski, Macintosh, Shagin, Medina,
  Liang, Zhou, Shah, Bi, Dankovics, Banga, Lehmann, Bredesen, Lin, Hoffmann,
  Lai, Chung, Yang, Balani, Bražinskas, Sozanschi, Hayes, Alcalde, Makarov,
  Chen, Stella, Snijders, Mandl, Kärrman, Nowak, Wu, Dyck, Vaidyanathan, R,
  Mallet, Rudominer, Johnston, Mittal, Udathu, Christensen, Verma, Irving,
  Santucci, Elsayed, Davoodi, Georgiev, Tenney, Hua, Cideron, Leurent,
  Alnahlawi, Georgescu, Wei, Zheng, Scandinaro, Jiang, Snoek, Sundararajan,
  Wang, Ontiveros, Karo, Cole, Rajashekhar, Tumeh, Ben-David, Jain, Uesato,
  Datta, Bunyan, Wu, Zhang, Stanczyk, Zhang, Steiner, Naskar, Azzam, Johnson,
  Paszke, Chiu, Elias, Mohiuddin, Muhammad, Miao, Lee, Vieillard, Park, Zhang,
  Stanway, Garmon, Karmarkar, Dong, Lee, Kumar, Zhou, Evens, Isaac, Irving,
  Loper, Fink, Arkatkar, Chen, Shafran, Petrychenko, Chen, Jia, Levskaya, Zhu,
  Grabowski, Mao, Magni, Yao, Snaider, Casagrande, Palmer, Suganthan, Castaño,
  Giannoumis, Kim, Rybiński, Sreevatsa, Prendki, Soergel, Goedeckemeyer,
  Gierke, Jafari, Gaba, Wiesner, Wright, Wei, Vashisht, Kulizhskaya, Hoover,
  Le, Li, Iwuanyanwu, Liu, Ramirez, Khorlin, Cui, LIN, Wu, Aguilar, Pallo,
  Chakladar, Perng, Abellan, Zhang, Dasgupta, Kushman, Penchev, Repina, Wu,
  van~der Weide, Ponnapalli, Kaplan, Simsa, Li, Dousse, Yang, Piper, Ie,
  Pasumarthi, Lintz, Vijayakumar, Andor, Valenzuela, Lui, Paduraru, Peng, Lee,
  Zhang, Greene, Nguyen, Kurylowicz, Hardin, Dixon, Janzer, Choo, Feng, Zhang,
  Singhal, Du, McKinnon, Antropova, Bolukbasi, Keller, Reid, Finchelstein,
  Raad, Crocker, Hawkins, Dadashi, Gaffney, Franko, Bulanova, Leblond, Chung,
  Askham, Cobo, Xu, Fischer, Xu, Sorokin, Alberti, Lin, Evans, Dimitriev,
  Forbes, Banarse, Tung, Omernick, Bishop, Sterneck, Jain, Xia, Amid, Piccinno,
  Wang, Banzal, Mankowitz, Polozov, Krakovna, Brown, Bateni, Duan, Firoiu,
  Thotakuri, Natan, Geist, tan Girgin, Li, Ye, Roval, Tojo, Kwong, Lee-Thorp,
  Yew, Sinopalnikov, Ramos, Mellor, Sharma, Wu, Miller, Sonnerat, Vnukov,
  Greig, Beattie, Caveness, Bai, Eisenschlos, Korchemniy, Tsai, Jasarevic,
  Kong, Dao, Zheng, Liu, Yang, Zhu, Teh, Sanmiya, Gladchenko, Trdin, Toyama,
  Rosen, Tavakkol, Xue, Elkind, Woodman, Carpenter, Papamakarios, Kemp, Kafle,
  Grunina, Sinha, Talbert, Wu, Owusu-Afriyie, Du, Thornton, Pont-Tuset,
  Narayana, Li, Fatehi, Wieting, Ajmeri, Uria, Ko, Knight, Héliou, Niu, Gu,
  Pang, Li, Levine, Stolovich, Santamaria-Fernandez, Goenka, Yustalim, Strudel,
  Elqursh, Deck, Lee, Li, Levin, Hoffmann, Holtmann-Rice, Bachem, Arora, Koh,
  Yeganeh, Põder, Tariq, Sun, Ionita, Seyedhosseini, Tafti, Liu, Gulati, Liu,
  Ye, Chrzaszcz, Wang, Sethi, Li, Brown, Singh, Fan, Parisi, Stanton,
  Koverkathu, Choquette-Choo, Li, Lu, Ittycheriah, Shroff, Varadarajan,
  Bahargam, Willoughby, Gaddy, Desjardins, Cornero, Robenek, Mittal, Albrecht,
  Shenoy, Moiseev, Jacobsson, Ghaffarkhah, Rivière, Walton, Crepy, Parrish,
  Zhou, Farabet, Radebaugh, Srinivasan, van~der Salm, Fidjeland, Scellato,
  Latorre-Chimoto, Klimczak-Plucińska, Bridson, de~Cesare, Hudson,
  Mendolicchio, Walker, Morris, Mauger, Guseynov, Reid, Odoom, Loher, Cotruta,
  Yenugula, Grewe, Petrushkina, Duerig, Sanchez, Yadlowsky, Shen, Globerson,
  Webb, Dua, Li, Bhupatiraju, Hurt, Qureshi, Agarwal, Shani, Eyal, Khare,
  Belle, Wang, Tekur, Kale, Wei, Sang, Saeta, Liechty, Sun, Zhao, Lee, Nayak,
  Fritz, Vuyyuru, Aslanides, Vyas, Wicke, Ma, Eltyshev, Martin, Cate, Manyika,
  Amiri, Kim, Xiong, Kang, Luisier, Tripuraneni, Madras, Guo, Waters, Wang,
  Ainslie, Baldridge, Zhang, Pruthi, Bauer, Yang, Mansour, Gelman, Xu,
  Polovets, Liu, Cai, Chen, Sheng, Xue, Ozair, Angermueller, Li, Sinha, Wang,
  Wiesinger, Koukoumidis, Tian, Iyer, Gurumurthy, Goldenson, Shah, Blake, Yu,
  Urbanowicz, Palomaki, Fernando, Durden, Mehta, Momchev, Rahimtoroghi,
  Georgaki, Raul, Ruder, Redshaw, Lee, Zhou, Jalan, Li, Hechtman, Schuh, Nasr,
  Milan, Mikulik, Franco, Green, Nguyen, Kelley, Mahendru, Hu, Howland, Vargas,
  Hui, Bansal, Rao, Ghiya, Wang, Ye, Sarr, Preston, Elish, Li, Kaku, Gupta,
  Pasupat, Juan, Someswar, M., Chen, Amini, Fabrikant, Chu, Dong, Muthal,
  Buthpitiya, Jauhari, Hua, Khandelwal, Hitron, Ren, Rinaldi, Drath, Dabush,
  Jiang, Godhia, Sachs, Chen, Fan, Taitelbaum, Noga, Dai, Wang, Liang, Hamer,
  Ferng, Elkind, Atias, Lee, Listík, Carlen, van~de Kerkhof, Pikus, Zaher,
  Müller, Zykova, Stefanec, Gatsko, Hirnschall, Sethi, Xu, Ahuja, Tsai,
  Stefanoiu, Feng, Dhandhania, Katyal, Gupta, Parulekar, Pitta, Zhao, Bhatia,
  Bhavnani, Alhadlaq, Li, Danenberg, Tu, Pine, Filippova, Ghosh, Limonchik,
  Urala, Lanka, Clive, Sun, Li, Wu, Hongtongsak, Li, Thakkar, Omarov,
  Majmundar, Alverson, Kucharski, Patel, Jain, Zabelin, Pelagatti, Kohli,
  Kumar, Kim, Sankar, Shah, Ramachandruni, Zeng, Bariach, Weidinger, Vu,
  Subramanya, Hsiao, Hassabis, Kavukcuoglu, Sadovsky, Le, Strohman, Wu, Petrov,
  Dean, and Vinyals]{geminiteam2024gemini}
Gemini Team, Rohan Anil, Sebastian Borgeaud, Jean-Baptiste Alayrac, Jiahui Yu,
  Radu Soricut, Johan Schalkwyk, Andrew~M. Dai, Anja Hauth, Katie Millican,
  David Silver, Melvin Johnson, Ioannis Antonoglou, Julian Schrittwieser,
  Amelia Glaese, Jilin Chen, Emily Pitler, Timothy Lillicrap, Angeliki
  Lazaridou, Orhan Firat, James Molloy, Michael Isard, Paul~R. Barham, Tom
  Hennigan, Benjamin Lee, Fabio Viola, Malcolm Reynolds, Yuanzhong Xu, Ryan
  Doherty, Eli Collins, Clemens Meyer, Eliza Rutherford, Erica Moreira, Kareem
  Ayoub, Megha Goel, Jack Krawczyk, Cosmo Du, Ed~Chi, Heng-Tze Cheng, Eric Ni,
  Purvi Shah, Patrick Kane, Betty Chan, Manaal Faruqui, Aliaksei Severyn,
  Hanzhao Lin, YaGuang Li, Yong Cheng, Abe Ittycheriah, Mahdis Mahdieh, Mia
  Chen, Pei Sun, Dustin Tran, Sumit Bagri, Balaji Lakshminarayanan, Jeremiah
  Liu, Andras Orban, Fabian Güra, Hao Zhou, Xinying Song, Aurelien Boffy,
  Harish Ganapathy, Steven Zheng, HyunJeong Choe, Ágoston Weisz, Tao Zhu,
  Yifeng Lu, Siddharth Gopal, Jarrod Kahn, Maciej Kula, Jeff Pitman, Rushin
  Shah, Emanuel Taropa, Majd~Al Merey, Martin Baeuml, Zhifeng Chen, Laurent~El
  Shafey, Yujing Zhang, Olcan Sercinoglu, George Tucker, Enrique Piqueras,
  Maxim Krikun, Iain Barr, Nikolay Savinov, Ivo Danihelka, Becca Roelofs,
  Anaïs White, Anders Andreassen, Tamara von Glehn, Lakshman Yagati, Mehran
  Kazemi, Lucas Gonzalez, Misha Khalman, Jakub Sygnowski, Alexandre Frechette,
  Charlotte Smith, Laura Culp, Lev Proleev, Yi~Luan, Xi~Chen, James Lottes,
  Nathan Schucher, Federico Lebron, Alban Rrustemi, Natalie Clay, Phil Crone,
  Tomas Kocisky, Jeffrey Zhao, Bartek Perz, Dian Yu, Heidi Howard, Adam
  Bloniarz, Jack~W. Rae, Han Lu, Laurent Sifre, Marcello Maggioni, Fred
  Alcober, Dan Garrette, Megan Barnes, Shantanu Thakoor, Jacob Austin, Gabriel
  Barth-Maron, William Wong, Rishabh Joshi, Rahma Chaabouni, Deeni Fatiha, Arun
  Ahuja, Gaurav~Singh Tomar, Evan Senter, Martin Chadwick, Ilya Kornakov,
  Nithya Attaluri, Iñaki Iturrate, Ruibo Liu, Yunxuan Li, Sarah Cogan, Jeremy
  Chen, Chao Jia, Chenjie Gu, Qiao Zhang, Jordan Grimstad, Ale~Jakse Hartman,
  Xavier Garcia, Thanumalayan~Sankaranarayana Pillai, Jacob Devlin, Michael
  Laskin, Diego de~Las~Casas, Dasha Valter, Connie Tao, Lorenzo Blanco,
  Adrià~Puigdomènech Badia, David Reitter, Mianna Chen, Jenny Brennan, Clara
  Rivera, Sergey Brin, Shariq Iqbal, Gabriela Surita, Jane Labanowski, Abhi
  Rao, Stephanie Winkler, Emilio Parisotto, Yiming Gu, Kate Olszewska, Ravi
  Addanki, Antoine Miech, Annie Louis, Denis Teplyashin, Geoff Brown, Elliot
  Catt, Jan Balaguer, Jackie Xiang, Pidong Wang, Zoe Ashwood, Anton Briukhov,
  Albert Webson, Sanjay Ganapathy, Smit Sanghavi, Ajay Kannan, Ming-Wei Chang,
  Axel Stjerngren, Josip Djolonga, Yuting Sun, Ankur Bapna, Matthew Aitchison,
  Pedram Pejman, Henryk Michalewski, Tianhe Yu, Cindy Wang, Juliette Love,
  Junwhan Ahn, Dawn Bloxwich, Kehang Han, Peter Humphreys, Thibault Sellam,
  James Bradbury, Varun Godbole, Sina Samangooei, Bogdan Damoc, Alex Kaskasoli,
  Sébastien M.~R. Arnold, Vijay Vasudevan, Shubham Agrawal, Jason Riesa,
  Dmitry Lepikhin, Richard Tanburn, Srivatsan Srinivasan, Hyeontaek Lim, Sarah
  Hodkinson, Pranav Shyam, Johan Ferret, Steven Hand, Ankush Garg, Tom~Le
  Paine, Jian Li, Yujia Li, Minh Giang, Alexander Neitz, Zaheer Abbas, Sarah
  York, Machel Reid, Elizabeth Cole, Aakanksha Chowdhery, Dipanjan Das,
  Dominika Rogozińska, Vitaliy Nikolaev, Pablo Sprechmann, Zachary Nado, Lukas
  Zilka, Flavien Prost, Luheng He, Marianne Monteiro, Gaurav Mishra, Chris
  Welty, Josh Newlan, Dawei Jia, Miltiadis Allamanis, Clara~Huiyi Hu, Raoul
  de~Liedekerke, Justin Gilmer, Carl Saroufim, Shruti Rijhwani, Shaobo Hou,
  Disha Shrivastava, Anirudh Baddepudi, Alex Goldin, Adnan Ozturel, Albin
  Cassirer, Yunhan Xu, Daniel Sohn, Devendra Sachan, Reinald~Kim Amplayo, Craig
  Swanson, Dessie Petrova, Shashi Narayan, Arthur Guez, Siddhartha Brahma,
  Jessica Landon, Miteyan Patel, Ruizhe Zhao, Kevin Villela, Luyu Wang, Wenhao
  Jia, Matthew Rahtz, Mai Giménez, Legg Yeung, James Keeling, Petko Georgiev,
  Diana Mincu, Boxi Wu, Salem Haykal, Rachel Saputro, Kiran Vodrahalli, James
  Qin, Zeynep Cankara, Abhanshu Sharma, Nick Fernando, Will Hawkins, Behnam
  Neyshabur, Solomon Kim, Adrian Hutter, Priyanka Agrawal, Alex Castro-Ros,
  George van~den Driessche, Tao Wang, Fan Yang, Shuo yiin Chang, Paul Komarek,
  Ross McIlroy, Mario Lučić, Guodong Zhang, Wael Farhan, Michael Sharman,
  Paul Natsev, Paul Michel, Yamini Bansal, Siyuan Qiao, Kris Cao, Siamak
  Shakeri, Christina Butterfield, Justin Chung, Paul~Kishan Rubenstein, Shivani
  Agrawal, Arthur Mensch, Kedar Soparkar, Karel Lenc, Timothy Chung, Aedan
  Pope, Loren Maggiore, Jackie Kay, Priya Jhakra, Shibo Wang, Joshua Maynez,
  Mary Phuong, Taylor Tobin, Andrea Tacchetti, Maja Trebacz, Kevin Robinson,
  Yash Katariya, Sebastian Riedel, Paige Bailey, Kefan Xiao, Nimesh Ghelani,
  Lora Aroyo, Ambrose Slone, Neil Houlsby, Xuehan Xiong, Zhen Yang, Elena
  Gribovskaya, Jonas Adler, Mateo Wirth, Lisa Lee, Music Li, Thais Kagohara,
  Jay Pavagadhi, Sophie Bridgers, Anna Bortsova, Sanjay Ghemawat, Zafarali
  Ahmed, Tianqi Liu, Richard Powell, Vijay Bolina, Mariko Iinuma, Polina
  Zablotskaia, James Besley, Da-Woon Chung, Timothy Dozat, Ramona Comanescu,
  Xiance Si, Jeremy Greer, Guolong Su, Martin Polacek, Raphaël~Lopez Kaufman,
  Simon Tokumine, Hexiang Hu, Elena Buchatskaya, Yingjie Miao, Mohamed
  Elhawaty, Aditya Siddhant, Nenad Tomasev, Jinwei Xing, Christina Greer, Helen
  Miller, Shereen Ashraf, Aurko Roy, Zizhao Zhang, Ada Ma, Angelos Filos, Milos
  Besta, Rory Blevins, Ted Klimenko, Chih-Kuan Yeh, Soravit Changpinyo, Jiaqi
  Mu, Oscar Chang, Mantas Pajarskas, Carrie Muir, Vered Cohen, Charline~Le Lan,
  Krishna Haridasan, Amit Marathe, Steven Hansen, Sholto Douglas, Rajkumar
  Samuel, Mingqiu Wang, Sophia Austin, Chang Lan, Jiepu Jiang, Justin Chiu,
  Jaime~Alonso Lorenzo, Lars~Lowe Sjösund, Sébastien Cevey, Zach Gleicher,
  Thi Avrahami, Anudhyan Boral, Hansa Srinivasan, Vittorio Selo, Rhys May,
  Konstantinos Aisopos, Léonard Hussenot, Livio~Baldini Soares, Kate Baumli,
  Michael~B. Chang, Adrià Recasens, Ben Caine, Alexander Pritzel, Filip
  Pavetic, Fabio Pardo, Anita Gergely, Justin Frye, Vinay Ramasesh, Dan Horgan,
  Kartikeya Badola, Nora Kassner, Subhrajit Roy, Ethan Dyer, Víctor~Campos
  Campos, Alex Tomala, Yunhao Tang, Dalia~El Badawy, Elspeth White, Basil
  Mustafa, Oran Lang, Abhishek Jindal, Sharad Vikram, Zhitao Gong, Sergi
  Caelles, Ross Hemsley, Gregory Thornton, Fangxiaoyu Feng, Wojciech Stokowiec,
  Ce~Zheng, Phoebe Thacker, Çağlar Ünlü, Zhishuai Zhang, Mohammad Saleh,
  James Svensson, Max Bileschi, Piyush Patil, Ankesh Anand, Roman Ring,
  Katerina Tsihlas, Arpi Vezer, Marco Selvi, Toby Shevlane, Mikel Rodriguez,
  Tom Kwiatkowski, Samira Daruki, Keran Rong, Allan Dafoe, Nicholas FitzGerald,
  Keren Gu-Lemberg, Mina Khan, Lisa~Anne Hendricks, Marie Pellat, Vladimir
  Feinberg, James Cobon-Kerr, Tara Sainath, Maribeth Rauh, Sayed~Hadi Hashemi,
  Richard Ives, Yana Hasson, Eric Noland, Yuan Cao, Nathan Byrd, Le~Hou, Qingze
  Wang, Thibault Sottiaux, Michela Paganini, Jean-Baptiste Lespiau, Alexandre
  Moufarek, Samer Hassan, Kaushik Shivakumar, Joost van Amersfoort, Amol
  Mandhane, Pratik Joshi, Anirudh Goyal, Matthew Tung, Andrew Brock, Hannah
  Sheahan, Vedant Misra, Cheng Li, Nemanja Rakićević, Mostafa Dehghani,
  Fangyu Liu, Sid Mittal, Junhyuk Oh, Seb Noury, Eren Sezener, Fantine Huot,
  Matthew Lamm, Nicola~De Cao, Charlie Chen, Sidharth Mudgal, Romina Stella,
  Kevin Brooks, Gautam Vasudevan, Chenxi Liu, Mainak Chain, Nivedita Melinkeri,
  Aaron Cohen, Venus Wang, Kristie Seymore, Sergey Zubkov, Rahul Goel, Summer
  Yue, Sai Krishnakumaran, Brian Albert, Nate Hurley, Motoki Sano, Anhad
  Mohananey, Jonah Joughin, Egor Filonov, Tomasz Kępa, Yomna Eldawy, Jiawern
  Lim, Rahul Rishi, Shirin Badiezadegan, Taylor Bos, Jerry Chang, Sanil Jain,
  Sri Gayatri~Sundara Padmanabhan, Subha Puttagunta, Kalpesh Krishna, Leslie
  Baker, Norbert Kalb, Vamsi Bedapudi, Adam Kurzrok, Shuntong Lei, Anthony Yu,
  Oren Litvin, Xiang Zhou, Zhichun Wu, Sam Sobell, Andrea Siciliano, Alan
  Papir, Robby Neale, Jonas Bragagnolo, Tej Toor, Tina Chen, Valentin Anklin,
  Feiran Wang, Richie Feng, Milad Gholami, Kevin Ling, Lijuan Liu, Jules
  Walter, Hamid Moghaddam, Arun Kishore, Jakub Adamek, Tyler Mercado, Jonathan
  Mallinson, Siddhinita Wandekar, Stephen Cagle, Eran Ofek, Guillermo Garrido,
  Clemens Lombriser, Maksim Mukha, Botu Sun, Hafeezul~Rahman Mohammad, Josip
  Matak, Yadi Qian, Vikas Peswani, Pawel Janus, Quan Yuan, Leif Schelin, Oana
  David, Ankur Garg, Yifan He, Oleksii Duzhyi, Anton Älgmyr, Timothée Lottaz,
  Qi~Li, Vikas Yadav, Luyao Xu, Alex Chinien, Rakesh Shivanna, Aleksandr
  Chuklin, Josie Li, Carrie Spadine, Travis Wolfe, Kareem Mohamed, Subhabrata
  Das, Zihang Dai, Kyle He, Daniel von Dincklage, Shyam Upadhyay, Akanksha
  Maurya, Luyan Chi, Sebastian Krause, Khalid Salama, Pam~G Rabinovitch, Pavan
  Kumar~Reddy M, Aarush Selvan, Mikhail Dektiarev, Golnaz Ghiasi, Erdem Guven,
  Himanshu Gupta, Boyi Liu, Deepak Sharma, Idan~Heimlich Shtacher, Shachi Paul,
  Oscar Akerlund, François-Xavier Aubet, Terry Huang, Chen Zhu, Eric Zhu,
  Elico Teixeira, Matthew Fritze, Francesco Bertolini, Liana-Eleonora
  Marinescu, Martin Bölle, Dominik Paulus, Khyatti Gupta, Tejasi Latkar, Max
  Chang, Jason Sanders, Roopa Wilson, Xuewei Wu, Yi-Xuan Tan, Lam~Nguyen Thiet,
  Tulsee Doshi, Sid Lall, Swaroop Mishra, Wanming Chen, Thang Luong, Seth
  Benjamin, Jasmine Lee, Ewa Andrejczuk, Dominik Rabiej, Vipul Ranjan,
  Krzysztof Styrc, Pengcheng Yin, Jon Simon, Malcolm~Rose Harriott, Mudit
  Bansal, Alexei Robsky, Geoff Bacon, David Greene, Daniil Mirylenka, Chen
  Zhou, Obaid Sarvana, Abhimanyu Goyal, Samuel Andermatt, Patrick Siegler, Ben
  Horn, Assaf Israel, Francesco Pongetti, Chih-Wei~"Louis" Chen, Marco
  Selvatici, Pedro Silva, Kathie Wang, Jackson Tolins, Kelvin Guu, Roey Yogev,
  Xiaochen Cai, Alessandro Agostini, Maulik Shah, Hung Nguyen, Noah~Ó
  Donnaile, Sébastien Pereira, Linda Friso, Adam Stambler, Adam Kurzrok,
  Chenkai Kuang, Yan Romanikhin, Mark Geller, ZJ~Yan, Kane Jang, Cheng-Chun
  Lee, Wojciech Fica, Eric Malmi, Qijun Tan, Dan Banica, Daniel Balle, Ryan
  Pham, Yanping Huang, Diana Avram, Hongzhi Shi, Jasjot Singh, Chris Hidey,
  Niharika Ahuja, Pranab Saxena, Dan Dooley, Srividya~Pranavi Potharaju, Eileen
  O'Neill, Anand Gokulchandran, Ryan Foley, Kai Zhao, Mike Dusenberry, Yuan
  Liu, Pulkit Mehta, Ragha Kotikalapudi, Chalence Safranek-Shrader, Andrew
  Goodman, Joshua Kessinger, Eran Globen, Prateek Kolhar, Chris Gorgolewski,
  Ali Ibrahim, Yang Song, Ali Eichenbaum, Thomas Brovelli, Sahitya Potluri,
  Preethi Lahoti, Cip Baetu, Ali Ghorbani, Charles Chen, Andy Crawford, Shalini
  Pal, Mukund Sridhar, Petru Gurita, Asier Mujika, Igor Petrovski, Pierre-Louis
  Cedoz, Chenmei Li, Shiyuan Chen, Niccolò~Dal Santo, Siddharth Goyal, Jitesh
  Punjabi, Karthik Kappaganthu, Chester Kwak, Pallavi LV, Sarmishta Velury,
  Himadri Choudhury, Jamie Hall, Premal Shah, Ricardo Figueira, Matt Thomas,
  Minjie Lu, Ting Zhou, Chintu Kumar, Thomas Jurdi, Sharat Chikkerur, Yenai Ma,
  Adams Yu, Soo Kwak, Victor Ähdel, Sujeevan Rajayogam, Travis Choma, Fei Liu,
  Aditya Barua, Colin Ji, Ji~Ho Park, Vincent Hellendoorn, Alex Bailey, Taylan
  Bilal, Huanjie Zhou, Mehrdad Khatir, Charles Sutton, Wojciech Rzadkowski,
  Fiona Macintosh, Konstantin Shagin, Paul Medina, Chen Liang, Jinjing Zhou,
  Pararth Shah, Yingying Bi, Attila Dankovics, Shipra Banga, Sabine Lehmann,
  Marissa Bredesen, Zifan Lin, John~Eric Hoffmann, Jonathan Lai, Raynald Chung,
  Kai Yang, Nihal Balani, Arthur Bražinskas, Andrei Sozanschi, Matthew Hayes,
  Héctor~Fernández Alcalde, Peter Makarov, Will Chen, Antonio Stella,
  Liselotte Snijders, Michael Mandl, Ante Kärrman, Paweł Nowak, Xinyi Wu,
  Alex Dyck, Krishnan Vaidyanathan, Raghavender R, Jessica Mallet, Mitch
  Rudominer, Eric Johnston, Sushil Mittal, Akhil Udathu, Janara Christensen,
  Vishal Verma, Zach Irving, Andreas Santucci, Gamaleldin Elsayed, Elnaz
  Davoodi, Marin Georgiev, Ian Tenney, Nan Hua, Geoffrey Cideron, Edouard
  Leurent, Mahmoud Alnahlawi, Ionut Georgescu, Nan Wei, Ivy Zheng, Dylan
  Scandinaro, Heinrich Jiang, Jasper Snoek, Mukund Sundararajan, Xuezhi Wang,
  Zack Ontiveros, Itay Karo, Jeremy Cole, Vinu Rajashekhar, Lara Tumeh, Eyal
  Ben-David, Rishub Jain, Jonathan Uesato, Romina Datta, Oskar Bunyan, Shimu
  Wu, John Zhang, Piotr Stanczyk, Ye~Zhang, David Steiner, Subhajit Naskar,
  Michael Azzam, Matthew Johnson, Adam Paszke, Chung-Cheng Chiu, Jaume~Sanchez
  Elias, Afroz Mohiuddin, Faizan Muhammad, Jin Miao, Andrew Lee, Nino
  Vieillard, Jane Park, Jiageng Zhang, Jeff Stanway, Drew Garmon, Abhijit
  Karmarkar, Zhe Dong, Jong Lee, Aviral Kumar, Luowei Zhou, Jonathan Evens,
  William Isaac, Geoffrey Irving, Edward Loper, Michael Fink, Isha Arkatkar,
  Nanxin Chen, Izhak Shafran, Ivan Petrychenko, Zhe Chen, Johnson Jia, Anselm
  Levskaya, Zhenkai Zhu, Peter Grabowski, Yu~Mao, Alberto Magni, Kaisheng Yao,
  Javier Snaider, Norman Casagrande, Evan Palmer, Paul Suganthan, Alfonso
  Castaño, Irene Giannoumis, Wooyeol Kim, Mikołaj Rybiński, Ashwin
  Sreevatsa, Jennifer Prendki, David Soergel, Adrian Goedeckemeyer, Willi
  Gierke, Mohsen Jafari, Meenu Gaba, Jeremy Wiesner, Diana~Gage Wright, Yawen
  Wei, Harsha Vashisht, Yana Kulizhskaya, Jay Hoover, Maigo Le, Lu~Li, Chimezie
  Iwuanyanwu, Lu~Liu, Kevin Ramirez, Andrey Khorlin, Albert Cui, Tian LIN,
  Marcus Wu, Ricardo Aguilar, Keith Pallo, Abhishek Chakladar, Ginger Perng,
  Elena~Allica Abellan, Mingyang Zhang, Ishita Dasgupta, Nate Kushman, Ivo
  Penchev, Alena Repina, Xihui Wu, Tom van~der Weide, Priya Ponnapalli,
  Caroline Kaplan, Jiri Simsa, Shuangfeng Li, Olivier Dousse, Fan Yang, Jeff
  Piper, Nathan Ie, Rama Pasumarthi, Nathan Lintz, Anitha Vijayakumar, Daniel
  Andor, Pedro Valenzuela, Minnie Lui, Cosmin Paduraru, Daiyi Peng, Katherine
  Lee, Shuyuan Zhang, Somer Greene, Duc~Dung Nguyen, Paula Kurylowicz, Cassidy
  Hardin, Lucas Dixon, Lili Janzer, Kiam Choo, Ziqiang Feng, Biao Zhang,
  Achintya Singhal, Dayou Du, Dan McKinnon, Natasha Antropova, Tolga Bolukbasi,
  Orgad Keller, David Reid, Daniel Finchelstein, Maria~Abi Raad, Remi Crocker,
  Peter Hawkins, Robert Dadashi, Colin Gaffney, Ken Franko, Anna Bulanova,
  Rémi Leblond, Shirley Chung, Harry Askham, Luis~C. Cobo, Kelvin Xu, Felix
  Fischer, Jun Xu, Christina Sorokin, Chris Alberti, Chu-Cheng Lin, Colin
  Evans, Alek Dimitriev, Hannah Forbes, Dylan Banarse, Zora Tung, Mark
  Omernick, Colton Bishop, Rachel Sterneck, Rohan Jain, Jiawei Xia, Ehsan Amid,
  Francesco Piccinno, Xingyu Wang, Praseem Banzal, Daniel~J. Mankowitz, Alex
  Polozov, Victoria Krakovna, Sasha Brown, MohammadHossein Bateni, Dennis Duan,
  Vlad Firoiu, Meghana Thotakuri, Tom Natan, Matthieu Geist, Ser tan Girgin,
  Hui Li, Jiayu Ye, Ofir Roval, Reiko Tojo, Michael Kwong, James Lee-Thorp,
  Christopher Yew, Danila Sinopalnikov, Sabela Ramos, John Mellor, Abhishek
  Sharma, Kathy Wu, David Miller, Nicolas Sonnerat, Denis Vnukov, Rory Greig,
  Jennifer Beattie, Emily Caveness, Libin Bai, Julian Eisenschlos, Alex
  Korchemniy, Tomy Tsai, Mimi Jasarevic, Weize Kong, Phuong Dao, Zeyu Zheng,
  Frederick Liu, Fan Yang, Rui Zhu, Tian~Huey Teh, Jason Sanmiya, Evgeny
  Gladchenko, Nejc Trdin, Daniel Toyama, Evan Rosen, Sasan Tavakkol, Linting
  Xue, Chen Elkind, Oliver Woodman, John Carpenter, George Papamakarios, Rupert
  Kemp, Sushant Kafle, Tanya Grunina, Rishika Sinha, Alice Talbert, Diane Wu,
  Denese Owusu-Afriyie, Cosmo Du, Chloe Thornton, Jordi Pont-Tuset, Pradyumna
  Narayana, Jing Li, Saaber Fatehi, John Wieting, Omar Ajmeri, Benigno Uria,
  Yeongil Ko, Laura Knight, Amélie Héliou, Ning Niu, Shane Gu, Chenxi Pang,
  Yeqing Li, Nir Levine, Ariel Stolovich, Rebeca Santamaria-Fernandez, Sonam
  Goenka, Wenny Yustalim, Robin Strudel, Ali Elqursh, Charlie Deck, Hyo Lee,
  Zonglin Li, Kyle Levin, Raphael Hoffmann, Dan Holtmann-Rice, Olivier Bachem,
  Sho Arora, Christy Koh, Soheil~Hassas Yeganeh, Siim Põder, Mukarram Tariq,
  Yanhua Sun, Lucian Ionita, Mojtaba Seyedhosseini, Pouya Tafti, Zhiyu Liu,
  Anmol Gulati, Jasmine Liu, Xinyu Ye, Bart Chrzaszcz, Lily Wang, Nikhil Sethi,
  Tianrun Li, Ben Brown, Shreya Singh, Wei Fan, Aaron Parisi, Joe Stanton,
  Vinod Koverkathu, Christopher~A. Choquette-Choo, Yunjie Li, TJ~Lu, Abe
  Ittycheriah, Prakash Shroff, Mani Varadarajan, Sanaz Bahargam, Rob
  Willoughby, David Gaddy, Guillaume Desjardins, Marco Cornero, Brona Robenek,
  Bhavishya Mittal, Ben Albrecht, Ashish Shenoy, Fedor Moiseev, Henrik
  Jacobsson, Alireza Ghaffarkhah, Morgane Rivière, Alanna Walton, Clément
  Crepy, Alicia Parrish, Zongwei Zhou, Clement Farabet, Carey Radebaugh,
  Praveen Srinivasan, Claudia van~der Salm, Andreas Fidjeland, Salvatore
  Scellato, Eri Latorre-Chimoto, Hanna Klimczak-Plucińska, David Bridson,
  Dario de~Cesare, Tom Hudson, Piermaria Mendolicchio, Lexi Walker, Alex
  Morris, Matthew Mauger, Alexey Guseynov, Alison Reid, Seth Odoom, Lucia
  Loher, Victor Cotruta, Madhavi Yenugula, Dominik Grewe, Anastasia
  Petrushkina, Tom Duerig, Antonio Sanchez, Steve Yadlowsky, Amy Shen, Amir
  Globerson, Lynette Webb, Sahil Dua, Dong Li, Surya Bhupatiraju, Dan Hurt,
  Haroon Qureshi, Ananth Agarwal, Tomer Shani, Matan Eyal, Anuj Khare,
  Shreyas~Rammohan Belle, Lei Wang, Chetan Tekur, Mihir~Sanjay Kale, Jinliang
  Wei, Ruoxin Sang, Brennan Saeta, Tyler Liechty, Yi~Sun, Yao Zhao, Stephan
  Lee, Pandu Nayak, Doug Fritz, Manish~Reddy Vuyyuru, John Aslanides, Nidhi
  Vyas, Martin Wicke, Xiao Ma, Evgenii Eltyshev, Nina Martin, Hardie Cate,
  James Manyika, Keyvan Amiri, Yelin Kim, Xi~Xiong, Kai Kang, Florian Luisier,
  Nilesh Tripuraneni, David Madras, Mandy Guo, Austin Waters, Oliver Wang,
  Joshua Ainslie, Jason Baldridge, Han Zhang, Garima Pruthi, Jakob Bauer, Feng
  Yang, Riham Mansour, Jason Gelman, Yang Xu, George Polovets, Ji~Liu, Honglong
  Cai, Warren Chen, XiangHai Sheng, Emily Xue, Sherjil Ozair, Christof
  Angermueller, Xiaowei Li, Anoop Sinha, Weiren Wang, Julia Wiesinger,
  Emmanouil Koukoumidis, Yuan Tian, Anand Iyer, Madhu Gurumurthy, Mark
  Goldenson, Parashar Shah, MK~Blake, Hongkun Yu, Anthony Urbanowicz,
  Jennimaria Palomaki, Chrisantha Fernando, Ken Durden, Harsh Mehta, Nikola
  Momchev, Elahe Rahimtoroghi, Maria Georgaki, Amit Raul, Sebastian Ruder,
  Morgan Redshaw, Jinhyuk Lee, Denny Zhou, Komal Jalan, Dinghua Li, Blake
  Hechtman, Parker Schuh, Milad Nasr, Kieran Milan, Vladimir Mikulik, Juliana
  Franco, Tim Green, Nam Nguyen, Joe Kelley, Aroma Mahendru, Andrea Hu, Joshua
  Howland, Ben Vargas, Jeffrey Hui, Kshitij Bansal, Vikram Rao, Rakesh Ghiya,
  Emma Wang, Ke~Ye, Jean~Michel Sarr, Melanie~Moranski Preston, Madeleine
  Elish, Steve Li, Aakash Kaku, Jigar Gupta, Ice Pasupat, Da-Cheng Juan, Milan
  Someswar, Tejvi M., Xinyun Chen, Aida Amini, Alex Fabrikant, Eric Chu, Xuanyi
  Dong, Amruta Muthal, Senaka Buthpitiya, Sarthak Jauhari, Nan Hua, Urvashi
  Khandelwal, Ayal Hitron, Jie Ren, Larissa Rinaldi, Shahar Drath, Avigail
  Dabush, Nan-Jiang Jiang, Harshal Godhia, Uli Sachs, Anthony Chen, Yicheng
  Fan, Hagai Taitelbaum, Hila Noga, Zhuyun Dai, James Wang, Chen Liang, Jenny
  Hamer, Chun-Sung Ferng, Chenel Elkind, Aviel Atias, Paulina Lee, Vít
  Listík, Mathias Carlen, Jan van~de Kerkhof, Marcin Pikus, Krunoslav Zaher,
  Paul Müller, Sasha Zykova, Richard Stefanec, Vitaly Gatsko, Christoph
  Hirnschall, Ashwin Sethi, Xingyu~Federico Xu, Chetan Ahuja, Beth Tsai, Anca
  Stefanoiu, Bo~Feng, Keshav Dhandhania, Manish Katyal, Akshay Gupta, Atharva
  Parulekar, Divya Pitta, Jing Zhao, Vivaan Bhatia, Yashodha Bhavnani, Omar
  Alhadlaq, Xiaolin Li, Peter Danenberg, Dennis Tu, Alex Pine, Vera Filippova,
  Abhipso Ghosh, Ben Limonchik, Bhargava Urala, Chaitanya~Krishna Lanka, Derik
  Clive, Yi~Sun, Edward Li, Hao Wu, Kevin Hongtongsak, Ianna Li, Kalind
  Thakkar, Kuanysh Omarov, Kushal Majmundar, Michael Alverson, Michael
  Kucharski, Mohak Patel, Mudit Jain, Maksim Zabelin, Paolo Pelagatti, Rohan
  Kohli, Saurabh Kumar, Joseph Kim, Swetha Sankar, Vineet Shah, Lakshmi
  Ramachandruni, Xiangkai Zeng, Ben Bariach, Laura Weidinger, Tu~Vu, Amar
  Subramanya, Sissie Hsiao, Demis Hassabis, Koray Kavukcuoglu, Adam Sadovsky,
  Quoc Le, Trevor Strohman, Y.
\newblock Gemini: A family of highly capable multimodal models,
  2024{\natexlab{b}}.

\bibitem[OpenAI et~al.(2024)OpenAI, Achiam, Adler, Agarwal, Ahmad, Akkaya,
  Aleman, Almeida, Altenschmidt, Altman, Anadkat, Avila, Babuschkin, Balaji,
  Balcom, Baltescu, Bao, Bavarian, Belgum, Bello, Berdine, Bernadett-Shapiro,
  Berner, Bogdonoff, Boiko, Boyd, Brakman, Brockman, Brooks, Brundage, Button,
  Cai, Campbell, Cann, Carey, Carlson, Carmichael, Chan, Chang, Chantzis, Chen,
  Chen, Chen, Chen, Chen, Chess, Cho, Chu, Chung, Cummings, Currier, Dai,
  Decareaux, Degry, Deutsch, Deville, Dhar, Dohan, Dowling, Dunning, Ecoffet,
  Eleti, Eloundou, Farhi, Fedus, Felix, Fishman, Forte, Fulford, Gao, Georges,
  Gibson, Goel, Gogineni, Goh, Gontijo-Lopes, Gordon, Grafstein, Gray, Greene,
  Gross, Gu, Guo, Hallacy, Han, Harris, He, Heaton, Heidecke, Hesse, Hickey,
  Hickey, Hoeschele, Houghton, Hsu, Hu, Hu, Huizinga, Jain, Jain, Jang, Jiang,
  Jiang, Jin, Jin, Jomoto, Jonn, Jun, Kaftan, Łukasz Kaiser, Kamali,
  Kanitscheider, Keskar, Khan, Kilpatrick, Kim, Kim, Kim, Kirchner, Kiros,
  Knight, Kokotajlo, Łukasz Kondraciuk, Kondrich, Konstantinidis, Kosic,
  Krueger, Kuo, Lampe, Lan, Lee, Leike, Leung, Levy, Li, Lim, Lin, Lin, Litwin,
  Lopez, Lowe, Lue, Makanju, Malfacini, Manning, Markov, Markovski, Martin,
  Mayer, Mayne, McGrew, McKinney, McLeavey, McMillan, McNeil, Medina, Mehta,
  Menick, Metz, Mishchenko, Mishkin, Monaco, Morikawa, Mossing, Mu, Murati,
  Murk, Mély, Nair, Nakano, Nayak, Neelakantan, Ngo, Noh, Ouyang, O'Keefe,
  Pachocki, Paino, Palermo, Pantuliano, Parascandolo, Parish, Parparita,
  Passos, Pavlov, Peng, Perelman, de~Avila Belbute~Peres, Petrov,
  de~Oliveira~Pinto, Michael, Pokorny, Pokrass, Pong, Powell, Power, Power,
  Proehl, Puri, Radford, Rae, Ramesh, Raymond, Real, Rimbach, Ross, Rotsted,
  Roussez, Ryder, Saltarelli, Sanders, Santurkar, Sastry, Schmidt, Schnurr,
  Schulman, Selsam, Sheppard, Sherbakov, Shieh, Shoker, Shyam, Sidor, Sigler,
  Simens, Sitkin, Slama, Sohl, Sokolowsky, Song, Staudacher, Such, Summers,
  Sutskever, Tang, Tezak, Thompson, Tillet, Tootoonchian, Tseng, Tuggle,
  Turley, Tworek, Uribe, Vallone, Vijayvergiya, Voss, Wainwright, Wang, Wang,
  Wang, Ward, Wei, Weinmann, Welihinda, Welinder, Weng, Weng, Wiethoff,
  Willner, Winter, Wolrich, Wong, Workman, Wu, Wu, Wu, Xiao, Xu, Yoo, Yu, Yuan,
  Zaremba, Zellers, Zhang, Zhang, Zhao, Zheng, Zhuang, Zhuk, and
  Zoph]{openai2024gpt4}
OpenAI, Josh Achiam, Steven Adler, Sandhini Agarwal, Lama Ahmad, Ilge Akkaya,
  Florencia~Leoni Aleman, Diogo Almeida, Janko Altenschmidt, Sam Altman,
  Shyamal Anadkat, Red Avila, Igor Babuschkin, Suchir Balaji, Valerie Balcom,
  Paul Baltescu, Haiming Bao, Mohammad Bavarian, Jeff Belgum, Irwan Bello, Jake
  Berdine, Gabriel Bernadett-Shapiro, Christopher Berner, Lenny Bogdonoff, Oleg
  Boiko, Madelaine Boyd, Anna-Luisa Brakman, Greg Brockman, Tim Brooks, Miles
  Brundage, Kevin Button, Trevor Cai, Rosie Campbell, Andrew Cann, Brittany
  Carey, Chelsea Carlson, Rory Carmichael, Brooke Chan, Che Chang, Fotis
  Chantzis, Derek Chen, Sully Chen, Ruby Chen, Jason Chen, Mark Chen, Ben
  Chess, Chester Cho, Casey Chu, Hyung~Won Chung, Dave Cummings, Jeremiah
  Currier, Yunxing Dai, Cory Decareaux, Thomas Degry, Noah Deutsch, Damien
  Deville, Arka Dhar, David Dohan, Steve Dowling, Sheila Dunning, Adrien
  Ecoffet, Atty Eleti, Tyna Eloundou, David Farhi, Liam Fedus, Niko Felix,
  Simón~Posada Fishman, Juston Forte, Isabella Fulford, Leo Gao, Elie Georges,
  Christian Gibson, Vik Goel, Tarun Gogineni, Gabriel Goh, Rapha Gontijo-Lopes,
  Jonathan Gordon, Morgan Grafstein, Scott Gray, Ryan Greene, Joshua Gross,
  Shixiang~Shane Gu, Yufei Guo, Chris Hallacy, Jesse Han, Jeff Harris, Yuchen
  He, Mike Heaton, Johannes Heidecke, Chris Hesse, Alan Hickey, Wade Hickey,
  Peter Hoeschele, Brandon Houghton, Kenny Hsu, Shengli Hu, Xin Hu, Joost
  Huizinga, Shantanu Jain, Shawn Jain, Joanne Jang, Angela Jiang, Roger Jiang,
  Haozhun Jin, Denny Jin, Shino Jomoto, Billie Jonn, Heewoo Jun, Tomer Kaftan,
  Łukasz Kaiser, Ali Kamali, Ingmar Kanitscheider, Nitish~Shirish Keskar,
  Tabarak Khan, Logan Kilpatrick, Jong~Wook Kim, Christina Kim, Yongjik Kim,
  Jan~Hendrik Kirchner, Jamie Kiros, Matt Knight, Daniel Kokotajlo, Łukasz
  Kondraciuk, Andrew Kondrich, Aris Konstantinidis, Kyle Kosic, Gretchen
  Krueger, Vishal Kuo, Michael Lampe, Ikai Lan, Teddy Lee, Jan Leike, Jade
  Leung, Daniel Levy, Chak~Ming Li, Rachel Lim, Molly Lin, Stephanie Lin,
  Mateusz Litwin, Theresa Lopez, Ryan Lowe, Patricia Lue, Anna Makanju, Kim
  Malfacini, Sam Manning, Todor Markov, Yaniv Markovski, Bianca Martin, Katie
  Mayer, Andrew Mayne, Bob McGrew, Scott~Mayer McKinney, Christine McLeavey,
  Paul McMillan, Jake McNeil, David Medina, Aalok Mehta, Jacob Menick, Luke
  Metz, Andrey Mishchenko, Pamela Mishkin, Vinnie Monaco, Evan Morikawa, Daniel
  Mossing, Tong Mu, Mira Murati, Oleg Murk, David Mély, Ashvin Nair, Reiichiro
  Nakano, Rajeev Nayak, Arvind Neelakantan, Richard Ngo, Hyeonwoo Noh, Long
  Ouyang, Cullen O'Keefe, Jakub Pachocki, Alex Paino, Joe Palermo, Ashley
  Pantuliano, Giambattista Parascandolo, Joel Parish, Emy Parparita, Alex
  Passos, Mikhail Pavlov, Andrew Peng, Adam Perelman, Filipe de~Avila
  Belbute~Peres, Michael Petrov, Henrique~Ponde de~Oliveira~Pinto, Michael,
  Pokorny, Michelle Pokrass, Vitchyr~H. Pong, Tolly Powell, Alethea Power,
  Boris Power, Elizabeth Proehl, Raul Puri, Alec Radford, Jack Rae, Aditya
  Ramesh, Cameron Raymond, Francis Real, Kendra Rimbach, Carl Ross, Bob
  Rotsted, Henri Roussez, Nick Ryder, Mario Saltarelli, Ted Sanders, Shibani
  Santurkar, Girish Sastry, Heather Schmidt, David Schnurr, John Schulman,
  Daniel Selsam, Kyla Sheppard, Toki Sherbakov, Jessica Shieh, Sarah Shoker,
  Pranav Shyam, Szymon Sidor, Eric Sigler, Maddie Simens, Jordan Sitkin,
  Katarina Slama, Ian Sohl, Benjamin Sokolowsky, Yang Song, Natalie Staudacher,
  Felipe~Petroski Such, Natalie Summers, Ilya Sutskever, Jie Tang, Nikolas
  Tezak, Madeleine~B. Thompson, Phil Tillet, Amin Tootoonchian, Elizabeth
  Tseng, Preston Tuggle, Nick Turley, Jerry Tworek, Juan Felipe~Cerón Uribe,
  Andrea Vallone, Arun Vijayvergiya, Chelsea Voss, Carroll Wainwright,
  Justin~Jay Wang, Alvin Wang, Ben Wang, Jonathan Ward, Jason Wei, CJ~Weinmann,
  Akila Welihinda, Peter Welinder, Jiayi Weng, Lilian Weng, Matt Wiethoff, Dave
  Willner, Clemens Winter, Samuel Wolrich, Hannah Wong, Lauren Workman, Sherwin
  Wu, Jeff Wu, Michael Wu, Kai Xiao, Tao Xu, Sarah Yoo, Kevin Yu, Qiming Yuan,
  Wojciech Zaremba, Rowan Zellers, Chong Zhang, Marvin Zhang, Shengjia Zhao,
  Tianhao Zheng, Juntang Zhuang, William Zhuk, and Barret Zoph.
\newblock Gpt-4 technical report, 2024.

\bibitem[Anthropic(2023)]{claude}
Anthropic.
\newblock {Model Card and Evaluations for Claude Models}, 2023.

\bibitem[DeepMind()]{pg-19}
DeepMind.
\newblock The pg-19 language modeling benchmark.
\newblock URL \url{https://github.com/deepmind/pg19}.

\bibitem[Bohnet et~al.(2023{\natexlab{a}})Bohnet, Alberti, and
  Collins]{bohnet-etal-2023-coreference}
Bernd Bohnet, Chris Alberti, and Michael Collins.
\newblock Coreference resolution through a seq2seq transition-based system.
\newblock \emph{Transactions of the Association for Computational Linguistics},
  11:\penalty0 212--226, 2023{\natexlab{a}}.
\newblock \doi{10.1162/tacl_a_00543}.
\newblock URL \url{https://aclanthology.org/2023.tacl-1.13}.

\bibitem[Dasigi et~al.(2019)Dasigi, Liu, Marasovi{\'c}, Smith, and
  Gardner]{dasigi-etal-2019-quoref}
Pradeep Dasigi, Nelson~F. Liu, Ana Marasovi{\'c}, Noah~A. Smith, and Matt
  Gardner.
\newblock {Q}uoref: A reading comprehension dataset with questions requiring
  coreferential reasoning.
\newblock In Kentaro Inui, Jing Jiang, Vincent Ng, and Xiaojun Wan, editors,
  \emph{Proceedings of the 2019 Conference on Empirical Methods in Natural
  Language Processing and the 9th International Joint Conference on Natural
  Language Processing (EMNLP-IJCNLP)}, pages 5925--5932, Hong Kong, China,
  November 2019. Association for Computational Linguistics.
\newblock \doi{10.18653/v1/D19-1606}.
\newblock URL \url{https://aclanthology.org/D19-1606}.

\bibitem[Rashkin et~al.(2023)Rashkin, Nikolaev, Lamm, Aroyo, Collins, Das,
  Petrov, Tomar, Turc, and Reitter]{rashkin_ais_2023}
Hannah Rashkin, Vitaly Nikolaev, Matthew Lamm, Lora Aroyo, Michael Collins,
  Dipanjan Das, Slav Petrov, Gaurav~Singh Tomar, Iulia Turc, and David Reitter.
\newblock {Measuring Attribution in Natural Language Generation Models}.
\newblock \emph{Computational Linguistics}, 49\penalty0 (4):\penalty0 777--840,
  12 2023.
\newblock ISSN 0891-2017.
\newblock \doi{10.1162/coli_a_00486}.
\newblock URL \url{https://doi.org/10.1162/coli\_a\_00486}.

\bibitem[Gao et~al.(2023)Gao, Dai, Pasupat, Chen, Chaganty, Fan, Zhao, Lao,
  Lee, Juan, and Guu]{gao-etal-2023-rarr}
Luyu Gao, Zhuyun Dai, Panupong Pasupat, Anthony Chen, Arun~Tejasvi Chaganty,
  Yicheng Fan, Vincent Zhao, Ni~Lao, Hongrae Lee, Da-Cheng Juan, and Kelvin
  Guu.
\newblock {RARR}: Researching and revising what language models say, using
  language models.
\newblock In Anna Rogers, Jordan Boyd-Graber, and Naoaki Okazaki, editors,
  \emph{Proceedings of the 61st Annual Meeting of the Association for
  Computational Linguistics (Volume 1: Long Papers)}, Toronto, Canada, July
  2023. Association for Computational Linguistics.
\newblock URL \url{https://aclanthology.org/2023.acl-long.910}.

\bibitem[Bohnet et~al.(2023{\natexlab{b}})Bohnet, Tran, Verga, Aharoni, Andor,
  Soares, Ciaramita, Eisenstein, Ganchev, Herzig, Hui, Kwiatkowski, Ma, Ni,
  Saralegui, Schuster, Cohen, Collins, Das, Metzler, Petrov, and
  Webster]{bohnet2023attributed}
Bernd Bohnet, Vinh~Q. Tran, Pat Verga, Roee Aharoni, Daniel Andor,
  Livio~Baldini Soares, Massimiliano Ciaramita, Jacob Eisenstein, Kuzman
  Ganchev, Jonathan Herzig, Kai Hui, Tom Kwiatkowski, Ji~Ma, Jianmo Ni,
  Lierni~Sestorain Saralegui, Tal Schuster, William~W. Cohen, Michael Collins,
  Dipanjan Das, Donald Metzler, Slav Petrov, and Kellie Webster.
\newblock Attributed question answering: Evaluation and modeling for attributed
  large language models, 2023{\natexlab{b}}.

\bibitem[Chiang et~al.(2024)Chiang, Zheng, Sheng, Angelopoulos, Li, Li, Zhang,
  Zhu, Jordan, Gonzalez, et~al.]{lmsys}
Wei-Lin Chiang, Lianmin Zheng, Ying Sheng, Anastasios~Nikolas Angelopoulos,
  Tianle Li, Dacheng Li, Hao Zhang, Banghua Zhu, Michael Jordan, Joseph~E
  Gonzalez, et~al.
\newblock Chatbot arena: An open platform for evaluating llms by human
  preference.
\newblock \emph{arXiv preprint}, 2024.
\newblock arXiv:2403.04132.

\bibitem[Rafailov et~al.(2023)Rafailov, Sharma, Mitchell, Ermon, Manning, and
  Finn]{directpreferenceoptimization}
Rafael Rafailov, Archit Sharma, Eric Mitchell, Stefano Ermon, Christopher~D.
  Manning, and Chelsea Finn.
\newblock Direct preference optimization: Your language model is secretly a
  reward model, 2023.

\bibitem[Saharia et~al.(2022)Saharia, Chan, Saxena, Li, Whang, Denton,
  Ghasemipour, Ayan, Mahdavi, Lopes, Salimans, Ho, Fleet, and Norouzi]{imagen}
Chitwan Saharia, William Chan, Saurabh Saxena, Lala Li, Jay Whang, Emily
  Denton, Seyed Kamyar~Seyed Ghasemipour, Burcu~Karagol Ayan, S.~Sara Mahdavi,
  Rapha~Gontijo Lopes, Tim Salimans, Jonathan Ho, David~J Fleet, and Mohammad
  Norouzi.
\newblock Photorealistic text-to-image diffusion models with deep language
  understanding, 2022.

\bibitem[Esser et~al.(2024)Esser, Kulal, Blattmann, Entezari, Müller, Saini,
  Levi, Lorenz, Sauer, Boesel, Podell, Dockhorn, English, Lacey, Goodwin,
  Marek, and Rombach]{stablediffusion3}
Patrick Esser, Sumith Kulal, Andreas Blattmann, Rahim Entezari, Jonas Müller,
  Harry Saini, Yam Levi, Dominik Lorenz, Axel Sauer, Frederic Boesel, Dustin
  Podell, Tim Dockhorn, Zion English, Kyle Lacey, Alex Goodwin, Yannik Marek,
  and Robin Rombach.
\newblock Scaling rectified flow transformers for high-resolution image
  synthesis, 2024.

\bibitem[Bradley and Terry(1952)]{BradleyTerry}
Ralph~A. Bradley and Milton~E. Terry.
\newblock The rank analysis of incomplete block designs --- {I. The} method of
  paired comparisons.
\newblock \emph{Biometrika}, 39:\penalty0 324--345, 1952.

\bibitem[Honovich et~al.(2021)Honovich, Choshen, Aharoni, Neeman, Szpektor, and
  Abend]{honovich-etal-2021-q2}
Or~Honovich, Leshem Choshen, Roee Aharoni, Ella Neeman, Idan Szpektor, and Omri
  Abend.
\newblock $q^{2}$: {E}valuating factual consistency in knowledge-grounded
  dialogues via question generation and question answering.
\newblock In Marie-Francine Moens, Xuanjing Huang, Lucia Specia, and Scott
  Wen-tau Yih, editors, \emph{Proceedings of the 2021 Conference on Empirical
  Methods in Natural Language Processing}, pages 7856--7870, Online and Punta
  Cana, Dominican Republic, November 2021. Association for Computational
  Linguistics.
\newblock \doi{10.18653/v1/2021.emnlp-main.619}.
\newblock URL \url{https://aclanthology.org/2021.emnlp-main.619}.

\bibitem[Kwiatkowski et~al.(2019)Kwiatkowski, Palomaki, Redfield, Collins,
  Parikh, Alberti, Epstein, Polosukhin, Devlin, Lee,
  et~al.]{kwiatkowski2019natural}
Tom Kwiatkowski, Jennimaria Palomaki, Olivia Redfield, Michael Collins, Ankur
  Parikh, Chris Alberti, Danielle Epstein, Illia Polosukhin, Jacob Devlin,
  Kenton Lee, et~al.
\newblock Natural questions: a benchmark for question answering research.
\newblock \emph{Transactions of the Association for Computational Linguistics},
  7:\penalty0 453--466, 2019.

\bibitem[Hunter(2004)]{hunter_2004_mm}
David~R. Hunter.
\newblock {MM Algorithms for Generalized Bradley-Terry Models}.
\newblock \emph{The Annals of Statistics}, 32\penalty0 (1):\penalty0 384--406,
  2004.
\newblock ISSN 00905364.
\newblock \doi{10.2307/3448514}.
\newblock URL \url{http://dx.doi.org/10.2307/3448514}.

\bibitem[Zermelo(1929)]{Zermelo1929}
Ernst Zermelo.
\newblock Die berechnung der turnier-ergebnisse als ein maximumproblem der
  wahrscheinlichkeitsrechnung.
\newblock \emph{Mathematische Zeitschrift}, 29\penalty0 (1):\penalty0 436--460,
  1929.
\newblock \doi{10.1007/BF01180541}.

\bibitem[Robertson(2009)]{robertson2009probabilistic}
S.~Robertson.
\newblock {The Probabilistic Relevance Framework: BM25 and Beyond}.
\newblock \emph{Foundations and Trends{\textregistered} in Information
  Retrieval}, 3\penalty0 (4):\penalty0 333--389, 2009.

\bibitem[Honovich et~al.(2022)Honovich, Aharoni, Herzig, Taitelbaum, Kukliansy,
  Cohen, Scialom, Szpektor, Hassidim, and Matias]{honovich-etal-2022-true}
Or~Honovich, Roee Aharoni, Jonathan Herzig, Hagai Taitelbaum, Doron Kukliansy,
  Vered Cohen, Thomas Scialom, Idan Szpektor, Avinatan Hassidim, and Yossi
  Matias.
\newblock {TRUE}: Re-evaluating factual consistency evaluation.
\newblock In Song Feng, Hui Wan, Caixia Yuan, and Han Yu, editors,
  \emph{Proceedings of the Second DialDoc Workshop on Document-grounded
  Dialogue and Conversational Question Answering}, pages 161--175, Dublin,
  Ireland, May 2022. Association for Computational Linguistics.
\newblock \doi{10.18653/v1/2022.dialdoc-1.19}.
\newblock URL \url{https://aclanthology.org/2022.dialdoc-1.19}.

\bibitem[Yang et~al.(2015)Yang, Yih, and Meek]{yang-etal-2015-wikiqa}
Yi~Yang, Wen-tau Yih, and Christopher Meek.
\newblock {W}iki{QA}: A challenge dataset for open-domain question answering.
\newblock In Llu{\'\i}s M{\`a}rquez, Chris Callison-Burch, and Jian Su,
  editors, \emph{Proceedings of the 2015 Conference on Empirical Methods in
  Natural Language Processing}, pages 2013--2018, Lisbon, Portugal, September
  2015. Association for Computational Linguistics.
\newblock \doi{10.18653/v1/D15-1237}.
\newblock URL \url{https://aclanthology.org/D15-1237}.

\bibitem[Yang et~al.(2018)Yang, Qi, Zhang, Bengio, Cohen, Salakhutdinov, and
  Manning]{yang2018hotpotqa}
Zhilin Yang, Peng Qi, Saizheng Zhang, Yoshua Bengio, William~W Cohen, Ruslan
  Salakhutdinov, and Christopher~D Manning.
\newblock Hotpotqa: A dataset for diverse, explainable multi-hop question
  answering.
\newblock \emph{arXiv preprint arXiv:1809.09600}, 2018.

\bibitem[Rajpurkar et~al.(2016)Rajpurkar, Zhang, Lopyrev, and
  Liang]{rajpurkar2016squad}
Pranav Rajpurkar, Jian Zhang, Konstantin Lopyrev, and Percy Liang.
\newblock Squad: 100,000+ questions for machine comprehension of text.
\newblock \emph{arXiv preprint arXiv:1606.05250}, 2016.

\bibitem[Reddy et~al.(2019)Reddy, Chen, and Manning]{reddy2019coqa}
Siva Reddy, Danqi Chen, and Christopher~D Manning.
\newblock Coqa: A conversational question answering challenge.
\newblock \emph{Transactions of the Association for Computational Linguistics},
  7:\penalty0 249--266, 2019.

\bibitem[Chen et~al.(2017)Chen, Fisch, Weston, and Bordes]{chen2017reading}
Danqi Chen, Adam Fisch, Jason Weston, and Antoine Bordes.
\newblock Reading wikipedia to answer open-domain questions.
\newblock 2017.

\bibitem[Joshi et~al.(2017)Joshi, Choi, Weld, and
  Zettlemoyer]{joshi2017triviaqa}
Mandar Joshi, Eunsol Choi, Daniel~S. Weld, and Luke Zettlemoyer.
\newblock Triviaqa: A large scale distantly supervised challenge dataset for
  reading comprehension.
\newblock 2017.

\bibitem[Talmor et~al.(2018)Talmor, Herzig, Lourie, and
  Berant]{talmor2018commonsenseqa}
Alon Talmor, Jonathan Herzig, Nicholas Lourie, and Jonathan Berant.
\newblock Commonsenseqa: A question answering challenge targeting commonsense
  knowledge.
\newblock \emph{arXiv preprint arXiv:1811.00937}, 2018.

\bibitem[Sakaguchi et~al.(2021)Sakaguchi, Bras, Bhagavatula, and
  Choi]{sakaguchi2021winogrande}
Keisuke Sakaguchi, Ronan~Le Bras, Chandra Bhagavatula, and Yejin Choi.
\newblock Winogrande: An adversarial winograd schema challenge at scale.
\newblock \emph{Communications of the ACM}, 64\penalty0 (9):\penalty0 99--106,
  2021.

\bibitem[Zhang and Choi(2021)]{zhang2021situatedqa}
Michael~JQ Zhang and Eunsol Choi.
\newblock Situatedqa: Incorporating extra-linguistic contexts into qa.
\newblock \emph{arXiv preprint arXiv:2109.06157}, 2021.

\bibitem[Liska et~al.(2022)Liska, Kocisky, Gribovskaya, Terzi, Sezener,
  Agrawal, Cyprien De~Masson, Scholtes, Zaheer, Young,
  et~al.]{liska2022streamingqa}
Adam Liska, Tomas Kocisky, Elena Gribovskaya, Tayfun Terzi, Eren Sezener,
  Devang Agrawal, D’Autume Cyprien De~Masson, Tim Scholtes, Manzil Zaheer,
  Susannah Young, et~al.
\newblock Streamingqa: A benchmark for adaptation to new knowledge over time in
  question answering models.
\newblock In \emph{International Conference on Machine Learning}, pages
  13604--13622. PMLR, 2022.

\bibitem[Kasai et~al.(2024)Kasai, Sakaguchi, Le~Bras, Asai, Yu, Radev, Smith,
  Choi, Inui, et~al.]{kasai2024realtime}
Jungo Kasai, Keisuke Sakaguchi, Ronan Le~Bras, Akari Asai, Xinyan Yu, Dragomir
  Radev, Noah~A Smith, Yejin Choi, Kentaro Inui, et~al.
\newblock Realtime qa: What's the answer right now?
\newblock \emph{Advances in Neural Information Processing Systems}, 36, 2024.

\bibitem[Wang et~al.(2018)Wang, Singh, Michael, Hill, Levy, and
  Bowman]{wang2018glue}
Alex Wang, Amanpreet Singh, Julian Michael, Felix Hill, Omer Levy, and Samuel~R
  Bowman.
\newblock Glue: A multi-task benchmark and analysis platform for natural
  language understanding.
\newblock \emph{arXiv preprint arXiv:1804.07461}, 2018.

\bibitem[Wang et~al.(2019)Wang, Pruksachatkun, Nangia, Singh, Michael, Hill,
  Levy, and Bowman]{wang2019superglue}
Alex Wang, Yada Pruksachatkun, Nikita Nangia, Amanpreet Singh, Julian Michael,
  Felix Hill, Omer Levy, and Samuel Bowman.
\newblock Superglue: A stickier benchmark for general-purpose language
  understanding systems.
\newblock \emph{Advances in neural information processing systems}, 32, 2019.

\bibitem[Hendrycks et~al.(2020)Hendrycks, Burns, Basart, Zou, Mazeika, Song,
  and Steinhardt]{hendrycks2020measuring}
Dan Hendrycks, Collin Burns, Steven Basart, Andy Zou, Mantas Mazeika, Dawn
  Song, and Jacob Steinhardt.
\newblock Measuring massive multitask language understanding.
\newblock \emph{arXiv preprint arXiv:2009.03300}, 2020.

\bibitem[Srivastava et~al.(2022)Srivastava, Rastogi, Rao, Shoeb, Abid, Fisch,
  Brown, Santoro, Gupta, Garriga-Alonso, et~al.]{srivastava2022beyond}
Aarohi Srivastava, Abhinav Rastogi, Abhishek Rao, Abu Awal~Md Shoeb, Abubakar
  Abid, Adam Fisch, Adam~R Brown, Adam Santoro, Aditya Gupta, Adri{\`a}
  Garriga-Alonso, et~al.
\newblock Beyond the imitation game: Quantifying and extrapolating the
  capabilities of language models.
\newblock \emph{arXiv preprint arXiv:2206.04615}, 2022.

\bibitem[Fan et~al.(2019)Fan, Jernite, Perez, Grangier, Weston, and
  Auli]{fan2019eli5}
Angela Fan, Yacine Jernite, Ethan Perez, David Grangier, Jason Weston, and
  Michael Auli.
\newblock Eli5: Long form question answering.
\newblock \emph{arXiv preprint arXiv:1907.09190}, 2019.

\bibitem[Cohan et~al.(2018)Cohan, Dernoncourt, Kim, Bui, Kim, Chang, and
  Goharian]{cohan-etal-2018-discourse}
Arman Cohan, Franck Dernoncourt, Doo~Soon Kim, Trung Bui, Seokhwan Kim, Walter
  Chang, and Nazli Goharian.
\newblock A discourse-aware attention model for abstractive summarization of
  long documents.
\newblock In Marilyn Walker, Heng Ji, and Amanda Stent, editors,
  \emph{Proceedings of the 2018 Conference of the North {A}merican Chapter of
  the Association for Computational Linguistics: Human Language Technologies,
  Volume 2 (Short Papers)}, pages 615--621, New Orleans, Louisiana, June 2018.
  Association for Computational Linguistics.
\newblock \doi{10.18653/v1/N18-2097}.
\newblock URL \url{https://aclanthology.org/N18-2097}.

\bibitem[Budzianowski et~al.(2018)Budzianowski, Wen, Tseng, Casanueva, Ultes,
  Ramadan, and Ga{\v{s}}i{\'c}]{budzianowski2018multiwoz}
Pawe{\l} Budzianowski, Tsung-Hsien Wen, Bo-Hsiang Tseng, Inigo Casanueva,
  Stefan Ultes, Osman Ramadan, and Milica Ga{\v{s}}i{\'c}.
\newblock Multiwoz--a large-scale multi-domain wizard-of-oz dataset for
  task-oriented dialogue modelling.
\newblock \emph{arXiv preprint arXiv:1810.00278}, 2018.

\end{thebibliography}
